\documentclass{article}

\PassOptionsToPackage{numbers, compress}{natbib}

\usepackage[preprint]{neurips_2023}
\usepackage{microtype}
\usepackage{subfig}
\usepackage{graphicx}
\usepackage{booktabs} 
\usepackage{caption} 
\usepackage{array} 
\usepackage{float}

\usepackage{algorithm}
\usepackage{algpseudocode}

\usepackage{booktabs} 

\usepackage[utf8]{inputenc} 
\usepackage[T1]{fontenc}    
\usepackage{hyperref}       
\usepackage{url}            
\usepackage{booktabs}       
\usepackage{amsfonts}       
\usepackage{nicefrac}       
\usepackage{microtype}      
\usepackage{xcolor}         
\usepackage{graphicx}

\usepackage{amsmath,amsfonts,bm}
\usepackage{amssymb}
\usepackage{mathtools}
\usepackage{amsthm}
\newcommand{\R}{\mathbb{R}}
\newcommand{\N}{\mathbb{N}}
\newcommand{\Z}{\mathbb{Z}}

\newcommand{\C}{\mathbb{C}}

\newcommand{\defeq}{\mathrel{\mathop:}=}
\newcommand{\inc}{\mathrm{in}}
\newcommand{\sca}{\mathrm{sc}}

\newcommand{\sfF}{\mathsf{F}}

\newcommand{\data}{\text{data}}

\DeclareMathOperator*{\argmax}{arg\,max}
\DeclareMathOperator*{\argmin}{arg\,min}

\DeclarePairedDelimiter\inner{\langle}{\rangle}

\newtheorem{thm}{Theorem}[section]
\newtheorem{lem}[thm]{Lemma}

\newtheorem{remark}[thm]{Remark}

\newtheorem{prop}[thm]{Proposition}
\newtheorem{definition}[thm]{Definition}

\graphicspath{ {./images/} }

\usepackage{hyperref}
\usepackage[capitalize,noabbrev]{cleveref}

\title{Back-Projection Diffusion: Solving the Wideband Inverse Scattering Problem with Diffusion Models}

\author{%
  Borong Zhang \\
  Department of Mathematics\\
  University of Wisconsin-Madison\\
  Madison, WI 53706 \\
  \texttt{bzhang388@wisc.edu} \\
  \And
  Martin Guerra \\
  Department of Mathematics\\
  University of Wisconsin-Madison\\
  Madison, WI 53706 \\
  \texttt{mguerra4@wisc.edu} \\
  \AND
  Qin Li \\
  Department of Mathematics \\
  University of Wisconsin-Madison\\
  Madison, WI 53706 \\
  \texttt{qinli@math.wisc.edu} \\
  \And
  Leonardo Zepeda-N\'u\~nez\thanks{Corresponding author.} \\
  Google Research \\
  Mountain View, CA 94043 \\
  \texttt{lzepedanunez@google.com} \\
}

\begin{document}

\maketitle

\begin{abstract} 
We present \textit{Wideband Back-Projection Diffusion}, an end-to-end probabilistic framework for approximating the posterior distribution induced by the inverse scattering map from wideband scattering data. 
This framework produces highly accurate reconstructions, leveraging conditional diffusion models to draw samples, and also honors the symmetries of the underlying physics of wave-propagation. The procedure is factored into two steps: the first step, inspired by the filtered back-propagation formula, transforms data into a physics-based latent representation, while the second step learns a conditional score function conditioned on this latent representation.
These two steps individually obey their associated symmetries and are amenable to compression by imposing the rank structure found in the filtered back-projection formula.
Empirically, our framework has both low sample and computational complexity, with its number of parameters scaling only sub-linearly with the target resolution, and has stable training dynamics. It provides sharp reconstructions effortlessly and is capable of recovering even sub-Nyquist features in the multiple-scattering regime.
\end{abstract}

\section{Introduction}
In this paper, we study the problem of high-resolution reconstruction of scatterers arising from wave-based inverse
problems~\cite{Beylkin_Burridge:1990,Tarantola:Inversion_of_seismic_reflection_data_in_the_acoustic_approximation,Virieux_Operto:An_overview_of_full-waveform_inversion_in_exploration_geophysics}. 
Wave-based inverse problems aim to reconstruct the properties (typically the refractive index) of an unknown medium by probing it with impinging waves and measuring the medium impulse response, in the form of scattered waves, at the boundary.
This task naturally arises in many scientific applications: for example, biomedical imaging~\cite{Schomberg:1978}, synthetic aperture radar~\cite{Cheney_SAR:2001}, nondestructive testing~\cite{Pettit:2015}, and
geophysics~\cite{Rawlinson:2010}.

Historically, the development of algorithmic pipelines for wave-based inverse problems has been hampered by three main issues. First, the diffraction limit~\cite{abbe1873beitrage,garnier2016passive} caps the maximum
resolution that a reconstruction can have. Following the Rayleigh criterion~\cite{rayleigh1879xxxi}, one typically increases the resolution by increasing the frequency of the probing wave. However, this can be
infeasible in practice, especially when data at high frequency is not readily available. Second, the numerical problem lacks the stability of the reconstruction~\cite{Hahner_Hohage2001_inverse_problem_estimates}. Even
though the problem is well-posed and stable at the continuum level, it becomes increasingly ill-posed in the finite-data regime. For classical methods based on PDE-constrained optimization, this translates into a myriad of
spurious local minima~\cite{Leeuwen_Herrmann:2013}, while in the case of ML-based methods, it translates to highly unstable training stages~\cite{MLZ}. Third, the algorithmic pipeline for inversion incurs a high
computational cost. As the resolution increases to capture fine-grained details, the computational complexity of state-of-the-art methods typically increases super-linearly with respect to the number of degrees of
freedom~\cite{ZepedaDemanet:the_method_of_polarized_traces,zepeda2019method}.

To bypass these issues, many methods have been proposed throughout the years, which we review in  \ref{sec:classical}. Such techniques can be broadly categorized in two main groups: analytical techniques, which rely on
asymptotic expansions coupled with {a painstaking} analysis of the mathematical properties of the involved operators~\cite{cakoni2005qualitative}, and optimization-based techniques, in which a data misfit loss is minimized
using {gradient-based} methods with either geometrical~\cite{chung2011adaptive} or PDE
constraints~\cite{Pratt:Seismic_waveform_inversion_in_the_frequency_domain;_Part_1_Theory_and_verification_in_a_physical_scale_model,chen1997inverse}. In general, there is a trade-off between computational cost and the
quality of the reconstruction, and depending on the computational and time constraints of the downstream applications, the PDE-constrained optimization techniques are often the preferred methodology.
Even though recent advances have been strikingly successful at accelerating the solution to the associated
PDE~\cite{appelo2020waveholtz,ZepedaDemanet:the_method_of_polarized_traces,taus2020sweeps,zepeda2016fast,gander2019class}, the overall algorithmic pipelines remain prohibitive.

In this context, one alluring alternative is to reconstruct the quantities of interest \textit{directly} from the scattered data, which amounts to parametrizing and finding the underlying non-linear \textit{inverse map}. The advent of modern ML tools has spurred the development of several ML-based models seeking to approximate such a map. Such approaches, which we review in Section~\ref{sec:ml_approaches}, usually rely on wideband data~\cite{MLZ,melia2024multi}, which has proven crucial to obtain high-resolution reconstructions, and on bespoke architectures~\cite{Butterfly-Net2,Khoo_YingSwitchNet:2019,YL} to avoid the pitfalls~\cite{Fprinciple_NeuroIPS} of dealing with highly oscillatory data.

Unfortunately, approximating this map prototypically exhibits three challenges commonly encountered in scientific ML (SciML). First: obtaining the training data in this setting -- whether synthetically or experimentally -- comes at considerable expense, which bottlenecks the size of the models that can be trained to satisfy the stringent accuracy requirements.
This necessitates the use of highly tailored architectures. Second: wave scattering involves \emph{non-smooth data} whose recordings are of highly oscillatory, broadband, scattered waveforms. These highly oscillatory (i.e.~high-frequency) signals are known to greatly hamper the training dynamics of many machine learning algorithms~\cite{Fprinciple_NeuroIPS} and thus require tailored strategies to mitigate their effect.
Third, current downstream applications often require quantification of the uncertainty on the reconstruction, which necessitates {learning} the distribution of all possible reconstructions for a given input. This usually involves stochastic methods that require the repeated application of the reconstruction, and thus rapidly increases the overall cost.

While many recently proposed ML-based methodologies~\cite{melia2024multi,YL,Khoo_YingSwitchNet:2019,MLZ,equivariant,li2021accurate} have been able to bypass the first two challenges, they are usually deterministic {and} do
not {\emph{natively}} provide any quantification of uncertainty.

Quantifying uncertainty in the reconstruction has a long story dating back to the Bayesian formulation of the inverse problem championed by Tarantola in the
80's~\cite{Tarantola:Inversion_of_seismic_reflection_data_in_the_acoustic_approximation,tarantola1982inverse}. In a nutshell, we seek to obtain the distribution of possible reconstructions conditioned on the input data,
instead of one particular reconstruction. Unfortunately, computing {this} distribution becomes computationally intractable as the dimension of the problem grows.

However, recent advances in generative models have shown that it is possible to approximate high-dimensional distributions efficiently from its samples~\cite{Ho_DDPM2020}. In particular, diffusion models have enjoyed great
empirical success, and more notably, they rely heavily on stochastic differential equations (SDEs), such as Langevin-type equations, which is remarkably close to the original formulation of
Tarantola.\footnote{We redirect the interest reader to~\cite{tarantola2005inverse,Stuart_2010:inverse_problems} for excellent reviews.} This has spurred a renewed interest {in} inverse problems from a
probabilistic standpoint~\cite{implicit_neural_vlasic_2022,Kawar_2022,khorashadizadeh2023deep,chung2022come,song2021solving,chung2022improving,pmlr-v202-finzi23a}. In this setting, the problem is recast as sampling the
{to-be-reconstructed} media from a learned distribution conditioned on the input data.
Even though such methods provide excellent reconstruction, they mostly focus on linear problems, as they merge off-the-shelf diffusion model architectures, such as
transformers~\cite{Vaswani_2017:attention_is_all_you_need}, for learning a prior, with a data misfit term encapsulating the physics. Computing the derivative for this last term, which is required for Langevin-type
formulations, becomes prohibitive as the dimension {increases,} as it requires {repeatedly simulating/solving} the system in the case of inverse scattering. However, as shown
in~\cite{dasgupta2024conditional,baldassari2024conditional}, it is possible to target the conditional probability \emph{directly} with standard architectures for diffusion models by \emph{learning from input--output
pairs} (far-field data and scatterers in the case of inverse scattering), thus bypassing the need for expensive simulations.
 Nonetheless, as we will show below, the behavior of such methods is suboptimal when applied to inverse scattering, as {the highly-oscillatory data requires specifically tailored architectures.}

Thus, considering the strengths of {recently introduced} deterministic architectures for inverse scattering and the empirically powerful frameworks of generative {AI,} the question arises: 

\textit{How can we incorporate physical information into a generative AI model that leverages the pairs of scatterers and far-field data directly?}

In this paper, we provide an answer to this question by introducing the Wideband Back-Projection Diffusion framework, which leverages diffusion models with architectures inspired {by} the analytical properties of the filtered
back-projection formula~\cite{fbp}, a centerpiece of many imaging technologies~\cite{Virieux_FWI:2017,Cheney_SAR:2001}, while exploiting symmetries in the formulation.

{The inverse map is factorized in two steps. The first step generates a latent space by aggregating information from the input and processing it in a hierarchical fashion, following the physics of wave propagation that preserves rotational equivariance. The second step performs a conditional sampling using a conditional diffusion model instantiated with a tailored conditional score function that preserves translational equivariance.} 

We showcase the properties of this framework on different distributions of perturbations, including standard biomedical imaging examples such as Shepp--Logan phantoms and brain data coming from MRI (NYU
fastMRI~\cite{fastmri1,fastmri2}), and more challenging examples with overlapping scatterers with sub-Nyquist features that exhibit {a} large amount of multiple scattering, which occurs when an impinging wave bounces between many objects before being captured by the receiver. 

\subsection{Contributions}
\begin{figure}[h!]
  \centering
  \includegraphics[width=0.9\textwidth]{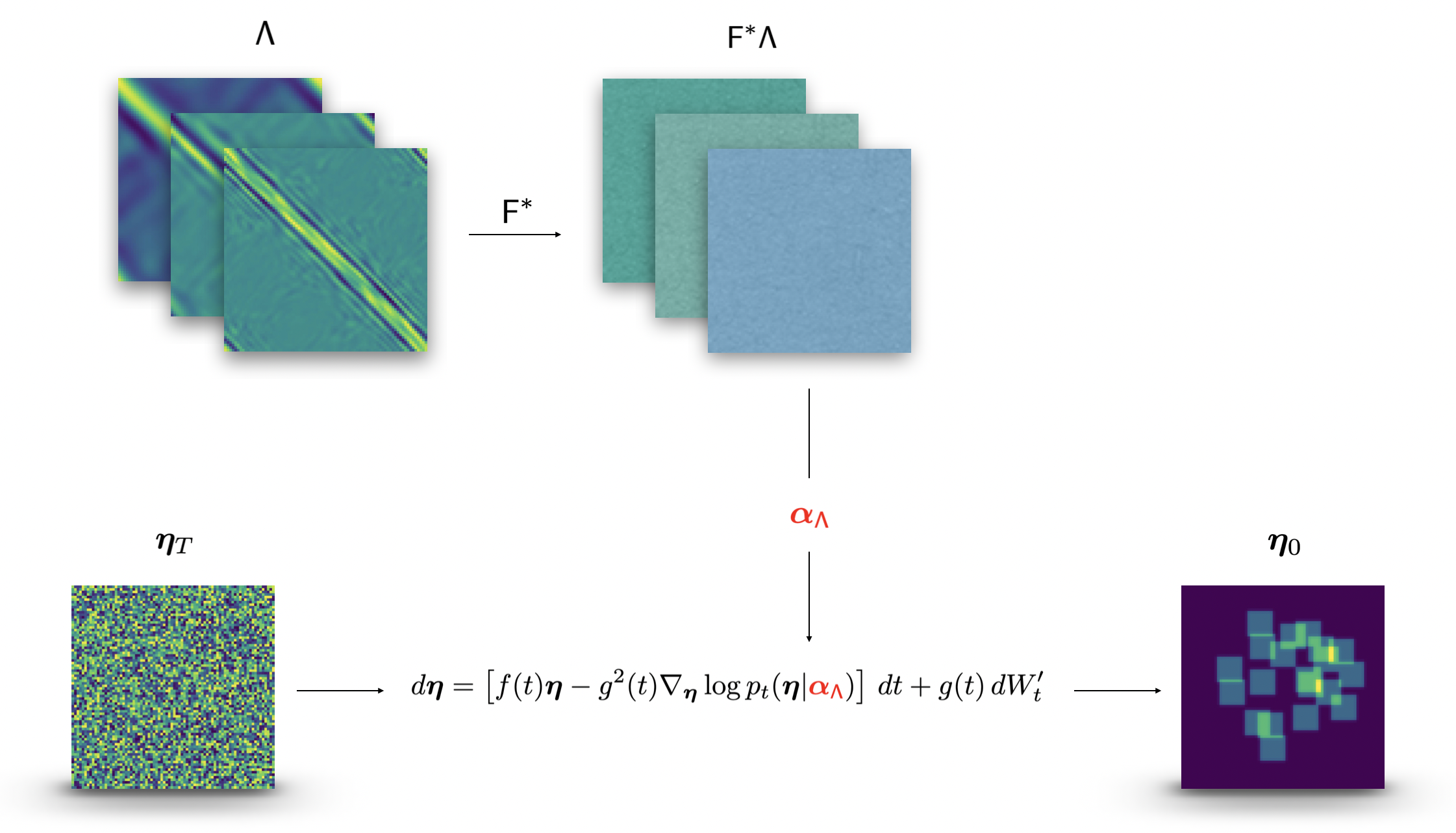}
  \caption{Sketch of the approach. The input data denoted by $\sf{\Lambda}$, a collection of scattered data at different frequency, is mapped to the reconstruction $\bm{\eta}_0$. The network $\sf{F}^{*}$ is used to map the input data to a latent representation, $\bm{\alpha}_{\sf{\Lambda}} = \sf{F}^{\star}\sf{\Lambda}$, which is then fed into the conditional score function, $\nabla_{\bm{\eta}} \log p_t(\bm{\eta} | \bm{\alpha}_{\sf{\Lambda}} )$ of a diffusion model. To reconstruct the medium, we solve the Langevin equation backwards, using Gaussian noise as terminal condition.}\label{fig:diagram_approach}
\end{figure}

We leverage generative models to sample from the posterior distribution of the scatterers conditioned on the input data.
The main novelty of our approach relies on the factorization of the conditional score function to incorporate the physics of wave propagation inspired {by} the filtered back-projection formula, while leveraging symmetries in the problem formulation, which we rigorously justify.
The factorization decomposes the score function approximation {into} two parts; the first processes the input data by exploiting the rotational equivariance of the problem and following a Butterfly-like architecture that mimics a
Fourier Integral Operator~\cite{Hormander:FIO}. This step creates a latent representation of the input data. The second part is instantiated by a conditional score function conditioned {on} the latent representation, which
preserves the {translational} equivariance of the operator.  \ref{fig:diagram_approach} shows {a} sketch of the approach.

This strategy has the following highly desirable properties: 

\textbf{Highly Accurate Reconstruction}: We show that our framework is able to reconstruct the underlying medium accurately, producing very sharp images even of objects with features below the diffraction limit, with around 1-2\% relative error, which is virtually indistinguishable to the naked eye {(see Figures~\ref{fig:reconstruction_shepplogan}, \ref{fig:reconstruction_3510triangles}, and \ref{fig:reconstruction_10hsquares})} . It outperforms other state-of-the-art models {(see Tables~\ref{tab:rrmse_melr_results} and~\ref{tab:ablation_study})} in the challenging cases involving multiple overlapping scatterers with strong multiple-scattering, and when the scatterers have features below the diffraction limit, such as diffraction corners  (see Figures~\ref{fig:comparison_3510triangles} and \ref{fig:comparison_10hsquares}).  

{\textbf{Training and Parameter Efficiency}: We demonstrate that high-accuracy reconstruction can be achieved with relatively low sample complexity (see Table~\ref{tab:sample_complexity} and Figure~\ref{fig:sample_complexity}), while requiring only a modest number of trainable parameters (see Table~\ref{tab:equinet_cnn_mri_brain}). This is accomplished by incorporating rotational equivariance into the latent representation and translational equivariance into the generative sampling. Additionally, leveraging rank-structured neural networks, such as Butterfly Networks~\cite{Butterfly-Net2}, to process the wideband data efficiently further reduces the scaling of the number of parameters relative to the target reconstruction resolution (see Table~\ref{table:complexity_comparison}), albeit with a slight trade-off in accuracy (see Table~\ref{tab:rrmse_melr_results}).}

\textbf{Training Stability}: The training stage is remarkably robust, particularly when compared to other inverse problem {algorithms}, which we inherit from the generative AI training pipelines~\cite{karras2022elucidating}. {We showcase these properties in Section~\ref{sec:cycle_skipping}.} 

\textbf{Resilience to Noise}: We show that our methodology is resilient to moderate measurement noise, {it} is able to learn different distributions of datasets, and {it} is able to handle scattered data from different {discretizations}, with only a minor reduction {in} the accuracy {(see Tables~\ref{tab:model_stability},~\ref{tab:noise_vs_rrmse}, and~\ref{tab:rrmse_comparison_datasets}; and Figures~\ref{fig:noise_comparison} and~\ref{fig:model_comparison_noise}).}

\subsection{Outline}

In Section \ref{sec:preliminary}, {we introduce the inverse scattering problem along with two classical approaches}. This includes the formulation of the inverse scattering problem and the associated filtered back-projection formula, and the Bayesian interpretation of PDE-constrained optimization. Section \ref{sec:diffusion_model_preliminary} briefly reviews some basics of score-based diffusion models, and the extension to sampling from conditional distributions. Section \ref{sec:methodology} is dedicated to the presentation of our proposed method that we term ``Wideband Back-Projection Diffusion model.'' This includes a specific factorization inspired by the filtered back-projection formula, with an examination of the mathematical properties of each component and their integration into the neural network design. This section also presents the major theoretical results of our work, demonstrating the required properties of the score function to ensure a certain equivariance structure. Finally, in Section \ref{sec:numerical_examples}, we provide ample numerical evidence showcasing the properties of the methodology.

\section{Inverse Scattering Problem}\label{sec:preliminary}

In this section, we provide a brief overview of the problem and {we discuss two classical approaches} for solving it. We present the filtered back-projection formula and highlight its properties that we will leverage in later sections. We introduce a Bayesian approach to solve this problem through posterior sampling, the numerical framework {that we will also leverage in later sections}.

\subsection{Problem Setup}
We focus on time-harmonic constant-density acoustic scattering in two dimensions, whose underlying physical model is given by the Helmholtz equation. Despite its simplicity, this model encapsulates the core challenges found in more complex models. In this case, the Helmholtz equation, the Fourier transform in time of the constant-density acoustic wave equation, is written as
\begin{equation}\label{eqn:helmholtz}
\Delta u(\bm{x}) + \omega^2 n(\bm{x}) u(\bm{x}) = 0\,, \quad \bm{x} \in \Omega \subset \mathbb{R}^2\,,
\end{equation}
where $ u $ is the total wave field, $ \omega $ is the frequency, and $ n $ is the refractive index. The domain of interest is $\Omega \subset \mathbb{R}^2$, {and the} homogeneous background is set to be $ n(\bm{x}) = 1 $ for $\bm{x} \notin \Omega$. {Defining} the perturbation $\eta(\bm{x}) = n(\bm{x}) - 1$, we have $\text{supp}(\eta(\bm{x})) \subset \Omega$.

\paragraph{Forward Problem}
For a given $n$ (or $\eta$), the forward problem involves solving for the scattered wave field resulting from the impulse response of the medium as it is impinged by a monochromatic plane wave,
\begin{equation} \label{eq:impinging_wave}
u^\inc = e^{i\omega \bm{s} \cdot \bm{x}}\,,
\end{equation}
where $\bm{s} \in \mathbb{S}^1$ is the direction of the incoming wave, and the scattered wave field $ u^\sca $ is defined as
\begin{equation} \label{eq:wavefield_decomposition}
u^\sca = u - u^\inc\,.
\end{equation}
Given that $u^\inc$ solves the Helmholtz equation in the background medium ($ n(\bm{x}) = 1 $), $ u^\sca(\bm{x}; \bm{s}) $ satisfies
\begin{equation}\label{eq:scattereqn}
\begin{cases}
\Delta u^\sca(\bm{x}) + \omega^2 (1 + \eta(\bm{x})) u^\sca(\bm{x}) = -\omega^2 \eta(\bm{x}) u^\inc\,, \\
\frac{\partial u^\sca}{\partial |\bm{x}|} - i\omega u^\sca = \mathcal{O}(|\bm{x}|^{-3/2}) \text{ uniformly as } |\bm{x}| \to \infty\,.
\end{cases}
\end{equation}
The second equation is the Sommerfeld radiation condition, ensuring the uniqueness of the solution.

We select the detector manifold~$D$ to be a circle of radius~$R$ that encloses the domain of interest~$\Omega$, i.e., $R > \text{radius}(\Omega)$.
For each incoming direction $\bm{s} \in \mathbb{S}^1$, the data is given by sampling the scattered field with receivers located on $D$ and indexed by $\bm{r} \in \mathbb{S}^1$.
This process yields the scattering data for each frequency~$\omega$ as a function~$\Lambda^\omega:[0,2\pi]^2\to\C$ such that
\begin{equation} \label{eq:far_field_pattern_def}
    \Lambda^{\omega}(r,s) = u^{\sca}(R\bm{r};\bm{s})\,,
\end{equation}
where $\bm{s} = (\cos(s), \sin(s))$, $\bm{r} = (\cos(r), \sin(r))$. We omit the dependence on $\omega$ on the right-hand side when the context is clear.

\begin{figure}[h!]
    \centering
        \centering
        \includegraphics[width=0.9\textwidth]{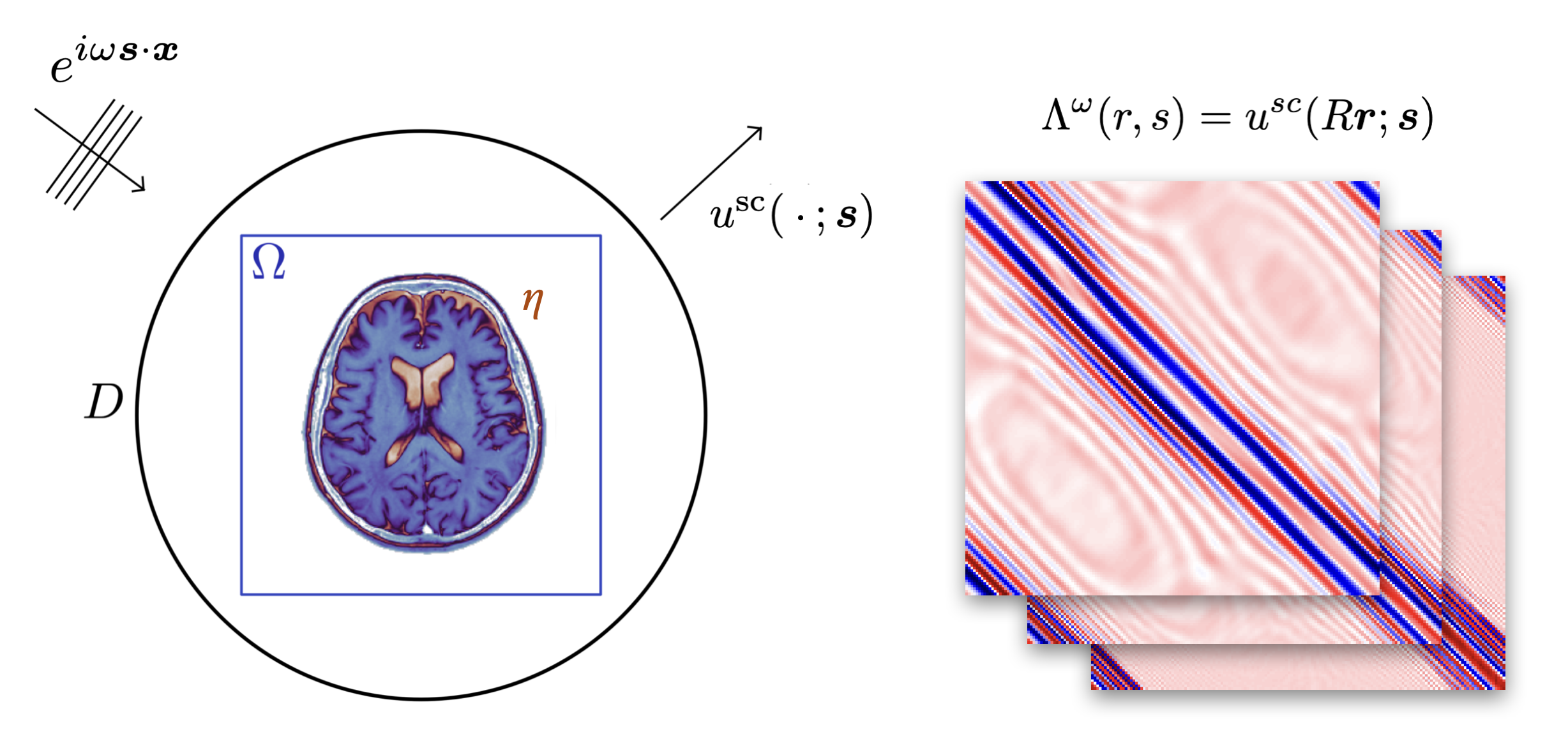}
    \caption{The plot on the left shows the setup for the inverse scattering problem: The perturbation $\eta$ is supported on the domain of interest $\Omega$. The media is impinged by a probing plane wave of frequency $\omega$ in the direction $\bm{s}$ and the scattered wave $u^\sca$ is collected. On the right, we show a typical scattering data $\Lambda^\omega(r,s)=u^\sca(R\bm{r};\bm{s})$.}\label{fig:inverse_scattering_setup}
\end{figure}

Each refractive index field $\eta$ can be mapped to a set of scattering data $\Lambda^\omega$. This map is denoted the {\it forward map}: $\Lambda^\omega=\mathcal{F}^{\omega}[\eta]$\footnote{We point out that even though the equation is linear, the map is nonlinear, since $u^\sca$ nonlinearly depends on $\eta$.}. Figure~\ref{fig:inverse_scattering_setup} illustrates the setup of the forward problem and the data acquisition. In practice, one can obtain the scattering data produced by multiple impinging wave frequencies, and we denote $\bar{\Omega}$ the discrete set of chosen frequencies.

\paragraph{Inverse Problem}

The inverse problem is to revert the process and to reconstruct $\eta$ from $\Lambda^\omega$. This amounts to {finding}:
\begin{equation} \label{eq:inv_formulation}
 \eta^{\ast} = \mathcal{F}^{-1}(\{\Lambda^{\omega}\}_{\omega \in \bar{\Omega}})\,.
\end{equation}
We can recast the inverse problem as a PDE-constrained optimization problem that seeks to minimize the \text{data misfit}, i.e., 
\begin{equation}\label{eq:inv_problem}
    \eta^{\ast} = \argmin_{\eta}\sum_{\omega\in\bar{\Omega}}\|\Lambda^{\omega} - \mathcal{F}^\omega[\eta]\|^{2}\,,
\end{equation}
where we consider the $L^{2}([0,2\pi]^2)$ norm for the data misfit, namely:
\begin{equation} \label{eq:data_misfit}
 \|\Lambda^\omega - \mathcal{F}^\omega \eta\|^2 = \int_{[0,2\pi]^2} |\Lambda^\omega(r,s) - \left(\mathcal{F}^\omega \eta\right)(r,s)|^2 \, dr\, ds\,.
\end{equation}
This formulation seeks the configuration of $\eta$ that minimizes the misfit between the synthetic data generated by $\eta$ (solving PDE in~\eqref{eq:scattereqn}) and the observed scattering data $\Lambda^\omega$. When a single frequency is used, the objective function is highly non-convex, with a standard gradient-based optimization approach converging to spurious local minima, a phenomenon termed cycle-skipping. Setting $|\Omega|\neq 1$ to utilize wideband data is a strategy to stabilize optimization~\cite{Pratt:Seismic_waveform_inversion_in_the_frequency_domain;_Part_1_Theory_and_verification_in_a_physical_scale_model,chen1997inverse}.

The optimization problem~\eqref{eq:inv_problem} is typically solved using tailored gradient-based optimization techniques 
whose gradients are computed via adjoint-state methods~\cite{Plessix_2006:ajoint_state}. Such optimization techniques either incorporate an explicit regularization term~\cite{SYMES1991147}, or leverage the sensitivity of \eqref{eq:data_misfit}
at different frequencies to solve \eqref{eq:inv_problem} in a hierarchical fashion~\cite{Pratt:Seismic_waveform_inversion_in_the_frequency_domain;_Part_1_Theory_and_verification_in_a_physical_scale_model,Borges_Gillman_Greengard:2017}.

\subsection{Filtered Back-Projection}~\label{sec:FBP}

Linearizing the forward operator, $\mathcal{F}^{\omega},$ is instructive as it sheds light on the essential difficulties of this problem and naturally leads to the filtered back-projection formula. This formula has inspired many of the recent machine learning-based algorithms~\cite{equivariant,MLZ,Khoo_YingSwitchNet:2019}. This formula also serves as an inspiration for our factorization to be presented in Section~\ref{sec:methodology}.

Using the classical Born approximation~\cite{Kirsch}, in \eqref{eq:scattereqn}, we obtain that 
\begin{equation}
    u^\sca(\bm{x}) = \omega^2 \int_{\mathbb{R}^2} \Phi^{\omega}(\bm{x}, \bm{y})  \eta(\bm{y}) e^{i \omega ( \bm{s} \cdot \bm{y})} d\bm{y}\,,
\end{equation}
where  $\Phi^{\omega}$ is the Green's function of the two-dimensional Helmholtz equation in a homogeneous medium, i.e.,  $\Phi^{\omega}$ solves
\begin{equation} \label{eq:GreenFunction}
    \left \{   \begin{array}{ll} \displaystyle
                \left(\Delta+\omega^{2}\right) \Phi^{\omega}(\bm{x}, \bm{y}) = -\delta(\bm{x}, \bm{y}) & \text{ for } \bm{x} \in \mathbb{R}^2\,,  \\ \displaystyle
                \lim _{|\bm{x}| \rightarrow \infty}|\bm{x}|^{1/2}\left(\frac{\partial}{\partial|\bm{x}|}-\mathrm{i} \omega\right) \Phi^{\omega}(\bm{x}, \bm{y}) =0\,.
                \end{array}
    \right .
\end{equation} 
Furthermore, we can use the classical far-field asymptotics of the Green's function to express
\begin{equation}
    u^\sca(R \bm{r}) = -\frac{e^{i\pi/4}}{\sqrt{8\pi\omega}}\omega^2 \frac{e^{i \omega R}}{\sqrt{R}} \int_{\mathbb{R}^2}  \eta(\bm{y}) e^{i \omega ( \bm{s} - \bm{r}) \cdot \bm{y}} d\bm{y} + \mathcal{O}(R^{-3/2})\,.
\end{equation}
Thus, up to a re-scaling factor and a phase change, the far-field pattern defined in \eqref{eq:far_field_pattern_def} can be approximately written as a Fourier transform of the perturbation, viz.:
\begin{equation} \label{eq:far_field_pattern}
    \Lambda^\omega(r,s) \approx F^{\omega} \eta = \int_{\mathbb{R}^2}  e^{i \omega ( \bm{s} - \bm{r}) \cdot \bm{y}} \eta(\bm{y}) d\bm{y}\,. 
\end{equation}
In this notation, $F^\omega$ is the linearized forward operator acting on the perturbation. In this linearized setting, solving the inverse problem in~\eqref{eq:data_misfit} using a single frequency $\omega$ with the linearized operator $F^\omega$ leads to the explicit solution
\begin{equation}
    \eta^* = \left ( F^{\omega} \right )^{\dagger} \Lambda^\omega \qquad \text{where} \quad \left(F^\omega\right)^\dagger = \left((F^\omega)^\ast F^\omega\right)^{-1} (F^\omega)^\ast\,.
\end{equation}
However, $F^{\omega}$ is usually ill-conditioned\footnote{One can show that this operator is compact~\cite{Kirsch}.},  one routinely leverages Tikhonov-regularization with regularization parameter~$\epsilon$, which results in the formula
\begin{equation}\label{eqn:FBP}
    \eta^* =  \left ( \left(F^{\omega} \right )^* F^{\omega} + \epsilon I \right )^{-1} \left(F^{\omega} \right )^*  \Lambda^\omega.
\end{equation}
This formula is referred to as \underline{filtered back-projection}~\cite{Colton_Kress:Integral_Equation_Methods_in_Scattering_Theory}, and is optimal with respect to the $L^2$-objective.  {The filtering step given by $\left(\left(F^{\omega} \right )^* F^{\omega} + \epsilon I \right )^{-1}$ induces a low-pass filter, particularly for large $\epsilon$.}
In practice,~$\epsilon$ is chosen to be sufficiently large so as to remedy the ill-conditioning of the normal operator $\left(F^{\omega} \right)^* F^{\omega}$, but small enough not to dampen the high-frequency content of the reconstruction.

This formula {decomposes the reconstruction into two stages}. The first stage applies the back-scattering operator $(F^\omega)^\ast$ to produce $\alpha^\omega$, the intermediate field:
\begin{equation}\label{eqn:def_alpha}
\alpha^\omega\defeq(F^\omega)^\ast \Lambda^\omega\,.
\end{equation}
This intermediate field can be computed explicitly following \eqref{eq:far_field_pattern}, up to a re-scaling factor, as:
\begin{equation}\label{eqn:F_ast}
\alpha^\omega(\bm{y}) = (F^\omega)^* \Lambda^\omega(\bm{y})=\int_{[0,2\pi]^2} e^{i\omega(\bm{r}-\bm{s})\cdot \bm{y}}\Lambda^\omega(r,s)\,dr\,ds\,.
\end{equation}
Then the second stage maps $\alpha^\omega$ through the filtering operator to the final reconstruction of $\eta^*$.

{Furthermore, in our setup,} the back-scattering operator and the filtering operator enjoy several mathematical properties and  symmetries, as outlined in the following propositions.
\begin{prop}\label{prop:Fstar_invertible}
    The back-scattering operator $(F^\omega)^\ast:L^2([0,2\pi]^2)\rightarrow L^2(\R^2)$ is injective.
\end{prop}
\begin{prop}\label{prop2}
    The back-scattering operator $(F^\omega)^\ast:L^2([0,2\pi]^2)\rightarrow L^2(\R^2)$ is  {rotationally equivariant.}
\end{prop}
\begin{prop}\label{prop3}
    The filtering operator $\left((F^\omega)^\ast F^\omega + \epsilon I\right)^{-1}$ that maps $\alpha^\omega$ to $\eta^\ast$ is {translationally} {equivariant}.
\end{prop}
The proof of Proposition~\ref{prop:Fstar_invertible} is included in Appendix~\ref{sec:appendixA}. The precise definitions of rotational and translational equivariance are provided in~\cite{equivariant}, where the justifications for Propositions~\ref{prop2} and~\ref{prop3} are also detailed.

\textbf{Remark:} So far, the method has been presented using single-frequency data, and the reconstruction is usually ill-posed in this regime. {Within many algorithmic pipelines}, data at additional frequencies is collectively used to stabilize the reconstruction~\cite{Hahner_Hohage2001_inverse_problem_estimates}. In particular, a time-domain formulation commonly known as the \emph{imaging condition} yields a more stable reconstruction using the full frequency bandwidth formally resulting in
\begin{equation}
\eta^* = \int_{\mathbb{R}} \left ( \left(F^{\omega} \right )^* F^{\omega} + \epsilon I \right )^{-1} \left(F^{\omega} \right )^*  \Lambda^\omega d \rho (\omega)\,,
\end{equation}
where $d \rho(\omega)$ is a density related to the frequency content of the probing wavelet. When the density is closely approximated by a discrete measure then 
\begin{equation} \label{eq:imaging_cond}
\eta^* \approx \sum_{\omega \in \bar{\Omega}} \left ( \left(F^{\omega} \right )^* F^{\omega} + \epsilon I  \right )^{-1} \left(F^{\omega} \right )^*  \Lambda^{\omega} \rho(\omega)\,,
\end{equation}
over a discrete set of frequencies $\bar{\Omega}$ and weights $\rho(\omega)$. We note that the selection of these frequencies, in addition to the optimal ordering in which the summation is computed under an iterative regime, remains an open question and an area of active research~\cite{Borges_Gillman_Greengard:2017}.

\subsection{Discretization}\label{sec:discretization}
We translate the discussion from the previous sections to the discrete setting. To streamline the notation, quantities in {calligraphic} fonts, such as $\mathcal{F}^\omega$, are used to denote nonlinear maps, while those in regular fonts, such as $F^\omega$ and $\Lambda^\omega$, are used to denote the linearized version. The quantities written in serif font, such as $\sfF^\omega$ and ${\sf\Lambda}^\omega$, are used to present the discretized version of the associated linear operators.

Since $\bm{s},\bm{r}\in\mathbb{S}^1$, we associate them with angles
\begin{equation}
\bm{s} = (\cos(s), \sin(s))\text{\quad and\quad}\bm{r} = (\cos(r), \sin(r))\,.
\end{equation}
Numerically, the directions of sources and detectors are represented by the same uniform grid in $\mathbb{S}^1$ with $n_\sca$ grid points given by
\begin{equation}
s_j, r_j = \frac{2\pi j}{n_\sca},\ j=0,\dots,n_\sca-1\,.
\end{equation}
Using this setting, the discrete scattering data ${\sf\Lambda}^\omega$ takes its values on the tensor of both grids with complex values, which are decomposed {into} their real and imaginary parts
\begin{equation}\label{eqn:def_dis_d}
{\sf\Lambda}^\omega={\sf\Lambda}^\omega_R+i{\sf\Lambda}^\omega_I\in \C^{n_\sca\times n_\sca}\,.
\end{equation}

We set the physical domain to be $\Omega=[-0.5,0.5]^2$ and use a Cartesian mesh of $n_\eta\times n_\eta$ grids. As a consequence, $\eta(\bm{x})$ is represented as a matrix: $\bm{\eta}\in\R^{n_\eta\times n_\eta}$ indexed by $i,j\in\N_\eta$ where $\N_\eta=\{0,1,\dots,n_\eta-1\}$ is the collection of grid points. In this form, $\bm{\eta}_{(i,j)}$ represents $\eta(\bm{x})$ evaluated on the Cartesian mesh. The intermediate field $\alpha^\omega(\bm{x})$ also lies in the physical domain, so it is discretized as a matrix: ${\bm{\alpha}^\omega\in\R^{n_\eta\times n_\eta}}$ {in the} same way as $\bm{\eta}$.

Upon this discretization, all operators, $\mathcal{F}^\omega$, $F^\omega$ and $(F^\omega)^\ast$ have their discrete counterparts. More specifically, we denote $\mathcal{F}_d^\omega:\R^{n_\eta\times n_\eta}\rightarrow\C^{n_\sca\times n_\sca}$, $\sfF^\omega:\R^{n_\eta\times n_\eta}\rightarrow\C^{n_\sca\times n_\sca}$ and $(\sfF^\omega)^\ast:\C^{n_\sca\times n_\sca}\rightarrow\R^{n_\eta\times n_\eta}$ 
the discretized forward map, linearized forward map, and back-scattering operator, respectively.

\subsection{Bayesian Sampling}\label{sec:bayesian_sampling}

Even though the reconstruction is unique, it has been shown to be unstable, particularly as the frequency increases~\cite{Hahner_Hohage2001_inverse_problem_estimates}. This poses a conundrum: reconstructions usually require higher frequency data to capture small-grained features, which are of great interest for downstream applications, but the reconstruction itself becomes increasingly unstable. This issue is further compounded in realistic scenarios where data always contains measurement errors and models present epistemic uncertainties. Thus, one alternative is to treat this problem under the Bayesian umbrella. Namely, one wants to compute, or have access to, the posterior distribution drawn from Bayes' rule~\cite{degroot2012probability}, i.e.,
\begin{equation}\label{eqn:post}
p(\bm{\eta} | {\sf\Lambda}^{\omega})\propto p(\bm{\eta})p({\sf\Lambda}^{\omega}|\bm{\eta})\,.
\end{equation}
with $p(\bm{\eta})$ being the prior distribution, serving as a regularization term, and $p({\sf\Lambda}^{\omega}|\bm{\eta})$ serving as the likelihood function. The reconstruction can be carried out by finding the maximum a posteriori (MAP) estimation:
\begin{equation}\label{eqn:Bayes}
\bm{\eta}^* = \argmax_{\bm{\eta}} p(\bm{\eta} | {\sf\Lambda}^{\omega})\,.
\end{equation}
The prior distribution is usually computed based on expected properties, such as sparse representation in Fourier space, of piecewise constant scatterers.

Computing this probability is intractable, but one can nevertheless sample from it. A standard strategy is to design a Markov chain whose invariant measure recovers the target distribution. In this context, the target distribution is the posterior distribution~\eqref{eqn:post}. If the Markov chain has this property, any random initialization for $\bm{\eta}$, after going through the chain along long enough pseudo-time, can potentially be viewed as a sample from the target distribution. There are many choices for designing this Markov chain, and one of the most popular is the Langevin-type:
{\begin{equation}\label{eq:langevin_type}
d\bm{\eta} = \nabla_{\bm{\eta}} \log p(\bm{\eta} | {\sf\Lambda}^{\omega}) dt + \sqrt{2} dW_t\,,
\end{equation}}
where $dW_t$ is a Wiener process.

As made evident in \eqref{eq:langevin_type}, knowing the score function $\nabla_{\bm{\eta}} \log p(\bm{\eta} | {\sf\Lambda}^{\omega})$ is crucial for sampling from the posterior distribution. It is often rare to find examples where the score function can be computed explicitly. Numerically, one seeks to find its numerical approximation. A typical assumption involves considering a Gaussian approximation to the misfit; specifically, assuming $\Sigma$, a positive definite matrix, is the covariance matrix of the measurement error, we derive:
\begin{equation}\label{eqn:gaussian_misfit}
p({\sf\Lambda}^{\omega} | \bm{\eta}) \propto \exp{\left(-\frac{1}{2} (\mathcal{F}^{\omega}_d[\bm{\eta}] - {\sf\Lambda}^{\omega})^\intercal \Sigma^{-1} (\mathcal{F}^{\omega}_d[\bm{\eta}] - {\sf\Lambda}^{\omega})\right) }\,.
\end{equation}
Throughout the paper, we simplify the formulation by assuming $\Sigma = \sigma^2\bm{I}$. In this case, we can integrate~\eqref{eqn:post} with \eqref{eqn:gaussian_misfit} to have
{\begin{equation}\label{eq:langevin_type_bayes}
\begin{aligned}
d\bm{\eta} &=  \nabla_{\bm{\eta}} \log p({\sf\Lambda}^{\omega} | \bm{\eta}) \, dt + \nabla_{\bm{\eta}} \log p(\bm{\eta}) \, dt + \sqrt{2} dW_t\,,\\
&=-\frac{1}{2\sigma^2}\nabla_{\bm{\eta}} \| \mathcal{F}_d^{\omega}[\bm{\eta}] - {\sf\Lambda}^{\omega} \|^2 \,dt + \nabla_{\bm{\eta}} \log p(\bm{\eta})\,dt+\sqrt{2} dW_t\,,
\end{aligned}
\end{equation}}
where the two terms in the velocity field respectively represent the gradient flow induced by the misfit function in \eqref{eq:data_misfit}, and a regularization term.

However, in what follows, we argue that we can learn this conditional score function \textit{directly} leveraging state-of-the-art generative AI tools.

\section{Denoising Diffusion Probabilistic Modeling (DDPM)}\label{sec:diffusion_model_preliminary}
The goal of score-based generative models is to be able to sample from a target data distribution using a sample from an easy-to-sample distribution, such as high-dimensional Gaussian, with a progressive transformation of the sample.
Theoretically, this procedure is backed by a simple observation that sequentially corrupting a sample of any distribution with increasing noise produces a sample drawn from a Gaussian distribution. Score-based generative {models seek to} revert this process and produce a sample from the target distribution by {increasingly ``denoising''} a Gaussian sample.  

Two mathematically equivalent computational frameworks {proposed for this task have become highly popular in recent years}: score matching with Langevin dynamics (SMLD)~\cite{song2019_score_based}, which estimates the \textit{score} (i.e., the gradient of the log probability density with respect to data), and denoising diffusion probabilistic models (DDPM)~\cite{Ho_DDPM2020}, which trains a sequence of probabilistic models to {sequentially} reverse the noise corruption of the data.

In this section we introduce the main ideas behind DDPM; from its mathematical foundation, to how we can extend it for a target conditional distribution, including some practical considerations ~\cite{song2021scorebased,karras2022elucidating,Luo2022UnderstandingDM} that render the algorithm more efficient.

\subsection{Mathematical Foundations}\label{sec:OU}

The mathematical foundation of DDPM lies on the well-known fact that a relatively simple stochastic process can map an arbitrary distribution ($p_\data$) to a humble Gaussian Normal. To see this, we start off with the classical Ornstein–Uhlenbeck (OU) process~\cite{OU_process},
\begin{equation}\label{eq:sde_langevin}
    \,d\bm{\eta}_t = -\bm{\eta}_t \,dt + \sqrt{2}\,dW_t\,,\quad \bm{\eta}_0\sim p_\data\,,
\end{equation}
where $W_t$ is a Brownian motion. The Feynman-Kac formula~\cite{Feynman-Kac_formula} suggests that the law of $\bm{\eta}$ solves the following Fokker-Planck equation:
\begin{equation}\label{eq:F-P_langevin}
    \partial_{t}p_t(\bm{\eta} ) = \nabla_{\bm{\eta}}\cdot(\bm{\eta}  p_t(\bm{\eta} )) + \Delta_{\bm{\eta}} p_t(\bm{\eta} )\,,\quad\text{with}\quad p_0=p_\data\,.
\end{equation}
{After a simple computation one can show that the limiting density} as $t\to\infty$ {satisfies},
\begin{equation}
p_{\infty}{(\bm{\eta})} \propto e^{-\frac{\|\bm{\eta}\|^{2}}{2}} \propto \mathcal{N}(\bm{\eta};\bm{0},\bm{I})\,,
\end{equation}
where $\mathcal{N}$ stands for normal distribution with mean and variance presented as the last two parameters. {This derivation implies that we can build a stochastic map from $p_\data$ to a standard Gaussian by solving~\eqref{eq:F-P_langevin}}. Equivalently, a sample $\bm{\eta}$ drawn from $p_\data$ will {become a sample from a Gaussian distribution through the OU process.}

DDPM seeks to revert this process. From~\eqref{eq:F-P_langevin}, by letting $t = T-\tau$, a simple derivation gives:
\begin{equation}\label{eq:F-P_reversed}
\begin{aligned}
    \partial_{\tau}p_{T-\tau}(\bm{\eta}) &= -\nabla_{\bm{\eta}}\cdot(\bm{\eta}  p_{T-\tau}(\bm{\eta})) - \Delta_{\bm{\eta}}p_{T-\tau}(\bm{\eta})\\
    &= -\nabla_{\bm{\eta}}\cdot(p_{T-\tau}(\bm{\eta})(\bm{\eta}  + 2\nabla_{\bm{\eta}}\log{p_{T-\tau}}(\bm{\eta}))) + \Delta_{\bm{\eta}}p_{T-\tau}(\bm{\eta})\\
    &= -\nabla_{\bm{\eta}}\cdot(p_{T-\tau}(\bm{\eta})(\bm{\eta}  + 2\nabla_{\bm{\eta}}\log{p_{t}}(\bm{\eta}))) + \Delta_{\bm{\eta}}p_{T-\tau}(\bm{\eta})\,,
\end{aligned}
\end{equation}
where $\nabla_{\bm{\eta}}\log{p_t}$ {is usually called the \emph{score function}} of the process. These dynamics can be represented by the associated SDE as well:
\begin{equation}\label{eq:sde_backwards_tau}
       d\tilde{\bm{\eta}}_{T-\tau}  = (\tilde{\bm{\eta}}_{T-\tau}  + 2\nabla_{\bm{\eta}}\log{p_{t}}(\tilde{\bm{\eta}}_{T-\tau}))\,d\tau + \sqrt{2}\,dW_{T-\tau}'\,.
\end{equation}
Change the time variable $t=T-\tau$ back:
\begin{equation}\label{eq:sde_backwards}
       d\tilde{\bm{\eta}}_t  = -(\tilde{\bm{\eta}}_t  + 2\nabla_{\bm{\eta}}\log{p_{t}}(\tilde{\bm{\eta}}_t ))\,dt + \sqrt{2}\,dW_t^{\prime}\,,
\end{equation}
where $W_t^{\prime}$ is a Brownian motion. Clearly, $p_{T-\tau}$ reverts the process of $p_t$ and thus maps a Gaussian normal distribution (in the $T\to\infty$ limit) to $p_\data$. Therefore, a sample $\tilde{\bm{\eta}}_T\sim\mathcal{N}(\,\cdot\,;\bm{0},\bm{I})$ and runs through~\eqref{eq:sde_backwards} approximately produces a sample from $p_{\text{data}}$ at $t=0$.
 
\subsection{Practical Considerations}\label{sec:fg_formulation}
It is straightforward to see that the OU process studied in Section~\ref{sec:OU} is not the only dynamics that {link} the target distribution $p_\data$ to $\mathcal{N}(\,\cdot\,;\bm{0},\bm{I})$. To speed up the dynamics, one has the freedom to adjust the velocity field and the strength of the Brownian motion. In particular, define $f$ as the drift coefficient, and $g$ the diffusion coefficient, we let $\bm{\eta}$ solve
\begin{equation}\label{eqn:forwardSDE}
    d\bm{\eta}_t  = f(t)\bm{\eta}_t  \, dt + g(t) \,dW_t\,,\quad\bm{\eta}_0\sim p_\data\,,
\end{equation}
with $W_t$ being the standard Wiener process. Then its law satisfies:
\begin{equation}\label{eqn:FP_fg}
\partial_t p_t(\bm{\eta}) = -\nabla_{\bm{\eta}} \cdot \left( \bm{\eta} f(t) p_t(\bm{\eta}) \right) + \frac{1}{2} g^2(t) \Delta_{\bm{\eta}} p_t(\bm{\eta})\,,\quad\text{with}\quad p_0=p_\data\,.
\end{equation}
The solution to this PDE is:
\begin{equation}\label{eq:solution_F-P_langevin}
p_{t}(\bm{\eta}) = \int p_{\data}(\bm{\eta}_0)p_{0t}(\bm{\eta}|\bm{\eta}_0)d\bm{\eta}_0 \propto (p_{\data}*\mathcal{N}(\,\cdot\,;\bm{0},\sigma^2(t)\bm{I}))\left(\frac{\bm{\eta}}{s(t)}\right)\,,   
\end{equation}
where the second equation comes from the change of variables, and $p_{0t}$ is the Green's function:
\begin{equation}\label{eqn:perturbation_kernel}
    p_{0t}(\bm{\eta}|\bm{\eta}_0) = \mathcal{N}(\bm{\eta}; s(t)\bm{\eta}_0, s^2(t) \sigma^2(t)\bm{I})\,.
\end{equation}
The $(s,\sigma)$ pair is uniquely determined by the $(f,g)$ pair:
\begin{equation}
\begin{cases}
s(t) = \exp\left(\int_0^t f(\xi) \, d\xi\right)\\
\sigma(t) = \sqrt{\int_0^t \frac{g(\xi)^2}{s(\xi)^2} \, d\xi}
\end{cases}\,,\quad\text{and equivalently}\quad\begin{cases}
    f(t)=\frac{\dot{s}(t)}{s(t)}\\
    g(t) = s(t)\sqrt{2\dot\sigma(t)\sigma(t)}
\end{cases}\,.
\end{equation}
The flexibility of adjusting $f$ and $g$ allows us to seek for dynamics that can drive $p_\data$ to a Gaussian faster than the simple OU process. In particular, suppose we wish $p_T\approx\mathcal{N}(\,\cdot\,;\bm{0},C^2\bm{I})$ at a finite $T$ for a prefixed $C$, one only needs to set: 
\begin{equation}
\lim_{t\to T}\sigma(t)=C\,,\quad \lim_{t\to T}s(t) = 0\quad\text{so that}\quad \lim_{t\to T}p_{t}(\bm{\eta}) =(\delta_0\ast \mathcal{N}(\,\cdot\,;\bm{0},C^2\bm{I}))(\bm{\eta})=\mathcal{N}(\bm{\eta};\bm{0},C^2\bm{I})\,.
\end{equation}
Having $s(t=T)=0$ can introduce singularity in $p_{\data}\left(\frac{\bm{\eta}}{s(t)}\right)$. To avoid numerical difficulties, we can relax it to be a very small number. One such example is to set
\begin{equation}\label{eqn_s_sigma_example}
s(t) = \frac{1}{\sqrt{\sigma^2(t)+1}}\,,\quad\text{with}\quad \sigma(t) = \tan\left( t_{\text{max}}t/T\right)\,,\quad t_{\max}=\arctan(C)\,,
\end{equation}
so that at $t=T$, $\sigma(T)=C$ and $s(T)=\frac{1}{\sqrt{C^2+1}}\ll1$ for $C\gg 1$. This noise schedule $\sigma(t)$ was proposed in~\cite{nichol2021improveddenoisingdiffusionprobabilistic}. The scaling factor $s(t)$ follows from the variance-preserving formulation proposed in~\cite{song2021scorebased}. These are the strategies we follow in our simulations. 

All these choices of drift and diffusion coefficients provide links between the target distribution $p_\data$ with a Gaussian. So similar to~\eqref{eq:F-P_reversed} and \eqref{eq:sde_backwards}, the reverse process that depends on the knowledge of the score function can be written as:
\begin{equation}\label{eqn:reverse_f_g_score}
d\tilde{\bm{\eta}}_t = \left[f(t)\tilde{\bm{\eta}}_t -g^2(t)\nabla_{\bm{\eta}}\log p_t(\tilde{\bm{\eta}}_t)\right] \, dt + g(t)\, dW_t^{\prime}\,,\quad\tilde{\bm{\eta}}_T\sim\mathcal{N}(\,\cdot\,;\bm{0},C^2\bm{I})\,.
\end{equation}
It provides a sample at $t=0$: $\tilde{\bm{\eta}}_0\sim p_\data$.

\subsection{Score function learning}\label{sec:DDPMformulation}
The success of running~\eqref{eqn:reverse_f_g_score} to process a desired sample hinges on the availability of the score function. However, it is typically unknown and needs to be learned from data in the offline training stage. To do so, it is conventional to parameterize it using a neural network and learn the weights using samples of the target distribution. 

To formulate the learning process through an objective function, we will recognize that the score function can be re-written by a conditional mean, and theoretically this conditional mean serves as a global optimizer of a specially designed loss function. To see so, we first rewrite~\eqref{eq:solution_F-P_langevin}. For any $\bm{\eta}_0\sim p_\data$:
\begin{equation} \label{eq:added_noise}
\bm{\eta}_{t} = s(t)\bm{\eta}_{0}+ \bm{\varepsilon}_{t}\,,\quad\text{with}\quad \bm{\varepsilon}_{t}\sim \mathcal{N}(\,\cdot\,;\bm{0},s^2(t)\sigma^{2}(t)\bm{I})\,.
\end{equation}
This presentation essentially writes $\bm{\eta}_t$ as a noised version of $\bm{\eta}_0$. To denoise $\bm{\eta}_t$ back to $\bm{\eta}_0$, we deploy the Tweedie's formula~\cite{Efron2011TweediesFA,vincent2011connection}\footnote{Tweedie's formula states that given $\bm{z}\sim\mathcal{N}(\,\cdot\,;\bm{\mu}_{\bm{z}},\bm{\Sigma}_{\bm{z}})$ we have $\mathbb{E}[\bm{\mu}_{\bm{z}}|\bm{z}] = \bm{z} + \bm{\Sigma}_{\bm{z}}\nabla_{\bm{z}}\log{p}(\bm{z})$.}:
\begin{equation}\label{eq:tweedie_application}
\mathbb{E}[\bm{\eta}_{0}|\bm{\eta}_{t}] = \frac{\bm{\eta}_{t}}{s(t)} + s(t)\sigma^{2}(t)\nabla_{\bm{\eta}}\log{p_{t}(\bm{\eta}_t)}\quad\Rightarrow\quad\nabla_{\bm{\eta}}\log{p_{t}}(\bm{\eta}_t) = \frac{\mathbb{E}[\bm{\eta}_{0}|\bm{\eta}_{t}] - \bm{\eta}_{t}/s(t)}{s(t)\sigma^{2}(t)}\,.
\end{equation}
Here the conditional mean is defined as:
\begin{equation}\label{eq:minimizer_proof}
\mathbb{E}[\bm{\eta}_0|\bm{\eta}_t] = \int \bm{\eta}_0 p(\bm{\eta}_0|\bm{\eta}_t)d\bm{\eta}_0 = \frac{1}{p_t(\bm{\eta}_t)}\int \bm{\eta}_0 p_{0t}(\bm{\eta}_t|\bm{\eta}_0)p_\data(\bm{\eta}_0)d\bm{\eta}_0\,,
\end{equation}
where we have used the Bayes' formula in the second equation. Noting this conditional mean is to recover the original signal $\bm{\eta}_0$ conditioned on a noisy version $\bm{\eta}_t$, and thus the term is interpreted as a \textit{denoiser}. Equation~\eqref{eq:tweedie_application} suggests that the computation of the score function can now be translated to the computation of $\mathbb{E}[\bm{\eta}_{0}|\bm{\eta}_{t}]$ in~\eqref{eq:minimizer_proof}. 

Remarkably, this conditional mean~\eqref{eq:minimizer_proof} is closely related to the optimizer of the following objective functional: 
\begin{equation} \label{eq:denoiser_minimization}
 \mathcal{L}(D;\sigma) = \mathbb{E}_{\bm{\eta} \sim p_{\text{data}}} \mathbb{E}_{\bm{n} \sim \mathcal{N}(\,\cdot\,;\bm{0}, \sigma^2 \mathbf{I})} \left\| D(\bm{\eta} + \bm{n}) - \bm{\eta} \right\|_2^2\,.
\end{equation}
For a fixed $\sigma$,  \eqref{eq:denoiser_minimization} maps a function of $\bm{\eta}$ to a non-negative number. Noting its quadratic and convex form, we derive its optimizer through the first critical condition. For every $t$, with $\sigma(t)$ fixed, setting:
 
\begin{equation}\label{eqn:first_order}
\left.\frac{\delta \mathcal{L}}{\delta D}\right|_{\mathbb{E}[\bm{\eta}_0|s(t)\cdot]}=0\quad\Rightarrow\quad \mathbb{E}\left[\bm{\eta}_0|s(t)\cdot\right]=\argmin \mathcal{L}(D;\sigma(t))\,.
\end{equation}
Numerically, given samples of $p_\data$ and $\bm{n}$ drawn from a Gaussian distribution, the $\mathbb{E}$ in $\mathcal{L}$ is replaced by its empirical mean to define $\mathcal{L}_e$. Letting $D_{\bm{\Theta}}$, a neural network parameterized by $\bm{\Theta}$, be the minimizer of this empirical objective functional,
\begin{equation}\label{eq:denoiser_opt}
D_{\bm{\Theta}}(\,\cdot\,; \sigma)=\argmin_{D_{\bm{\Theta}'}}\mathcal{L}_e(D_{\bm{\Theta}'};\sigma )\,,
 \end{equation}
then considering~\eqref{eqn:first_order}, $\forall t$, we have:
\begin{equation}
\mathbb{E}\left[\bm{\eta}_0| \bm{\eta}_t\right]\approx D_{\bm{\Theta}}\left(\frac{\bm{\eta}_t}{s(t)}; \sigma(t)\right)\,.
\end{equation}
The $\approx$ sign accounts for the failure of finding {the} global optimum of~\eqref{eq:denoiser_opt}, lack of approximation power of the neural network feasible set, and the empirical approximation of $\mathcal{L}$ by $\mathcal{L}_e$. 

This numerical conditional mean $D_{\bm{\Theta}}\left(\frac{\bm{\eta}_t}{s(t)}; \sigma(t)\right)$ then is integrated {into} the score-function formula~\eqref{eq:tweedie_application} to enter the online stage for drawing a sample. We prepare $\bm{\eta}_T\sim \mathcal{N}(\,\cdot\,;\bm{0},C^2\bm{I})$ and run the following dynamics:
\begin{equation}
 d\bm{\eta}_t = \left[\frac{\dot{s}(t)}{s(t)}\bm{\eta}_t -2s^2(t)\dot{\sigma}(t) \sigma(t) \frac{D_{\bm{\Theta}}(\bm{\eta}_t/s(t);\sigma(t)) - \bm{\eta}_t/s(t)}{s(t)\sigma^{2}(t)}\right] \, dt + s(t) \sqrt{2 \dot{\sigma}(t) \sigma(t)}\, dW_t^{\prime}\,,
\end{equation}
from $t=T$ back to $t=0$. The output $\bm{\eta}_0$ provides an approximate sample from $p_{\text{data}}$.

\subsection{Conditional Diffusion}\label{sec:conditional_diffusion_model}
In the context of the inverse scattering problem, as presented in Section~\ref{sec:bayesian_sampling}, and~\eqref{eq:langevin_type}, we are given the scattering data ${\sf\Lambda}^\omega$, and aim to draw a sample from the target distribution $p(\,\cdot\,|{\sf\Lambda}^\omega)$. As a consequence, both the offline training and online drawing processes are conducted in this conditioned setting.

In the offline training stage, we run the optimization~\eqref{eq:denoiser_opt} with $p_\data$ replaced by $p(\,\cdot\,|{\sf\Lambda}^\omega)$. The output of the optimization formulation provides the approximation to the score function:
\begin{equation}\label{eq:denoiser_score}
    D_{\bm{\Theta}}\left(\frac{\bm{\eta}}{s(t)};{\sf\Lambda}^\omega,\sigma(t)\right) \approx \frac{\bm{\eta}}{s(t)} + s(t)\sigma^{2}(t)\nabla_{\bm{\eta}}\log{p_t}(\bm{\eta}|{\sf\Lambda}^\omega)\,.
\end{equation}

With this knowledge in hand, in the online drawing, we prepare $\bm{\eta}_{T}\sim \mathcal{N}(\,\cdot\,;\bm{0},C^{2}\bm{I})$, and run
reverse-time SDE~\cite{song2021scorebased} as
\begin{equation}\label{eqn:reverseSDE}
d\bm{\eta}_t = \left[\frac{\dot{s}(t)}{s(t)}\bm{\eta}_t -2s^2(t)\dot{\sigma}(t) \sigma(t) \frac{D_{\bm{\Theta}}(\bm{\eta}_t/s(t);{\sf\Lambda}^\omega,\sigma(t)) - \bm{\eta}_t/s(t)}{s(t)\sigma^{2}(t)}\right] \, dt + s(t) \sqrt{2 \dot{\sigma}(t) \sigma(t)}\, dW_t^{\prime}\,,
\end{equation}
from $t=T$ up to $t=0$. The output is a sample from the target $p(\,\cdot\,|{\sf\Lambda}^\omega)$. The choices of $(s(t),\sigma(t))$ are consistent {with} those previously defined in~\eqref{eqn_s_sigma_example}.

\section{Wideband Back-Projection Diffusion Model}\label{sec:methodology}
In this section, we integrate all the techniques presented above and we tailor them to solve the inverse scattering problem.
More specifically, we aim to infer $\bm{\eta}$ from the knowledge of $\{{\sf\Lambda}^{\omega}\}_{\omega \in \bar{\Omega}}$ using the Bayesian framework introduced in Section \ref{sec:bayesian_sampling}. This involves drawing a sample from the posterior distribution using the DDPM framework:
\begin{equation}\label{eqn:drawing_post}
\bm{\eta}^\ast\sim p(\bm{\eta} |\{{\sf\Lambda}^{\omega}\}_{\omega \in \bar{\Omega}})\,.
\end{equation}
The dataset used to learn this posterior distribution is denoted
\begin{equation}
(\bm{\eta},\{{\sf\Lambda}^{\omega}\}_{\omega \in \bar{\Omega}})\sim p_\data\,,
\end{equation}
and, when the context is clear, $p_\data(\bm{\eta})$ denotes the marginal distribution. For the sake of brevity, we will omit the frequency $\omega$ from the discussion unless necessary. Particularly, in the presence of the wideband data $\{{\sf\Lambda}^\omega\}_{\omega\in\bar{\Omega}}$, with an abuse of notation, we denote
\begin{equation}
{\sf\Lambda}=\{{\sf\Lambda}^{\omega}\}_{\omega \in \bar{\Omega}}\quad\text{ and }\quad \bm{\alpha}_{\sf\Lambda}=\{\bm{\alpha}^\omega_{{\sf\Lambda}}\}_{\omega \in \bar{\Omega}}\,.
\end{equation} 
We recall that the inverse scattering problem, as presented in Section~\ref{sec:FBP}, is composed of two stages: the back-scattering and the filtering, see~\eqref{eqn:FBP}. These two stages exhibit very different structures, as discussed in Propositions~\ref{prop2} and~\ref{prop3}. This structural difference suggests these two operations should be treated separately, thus inducing our factorization into a two-stage reconstruction:

\begin{itemize}
    \item[Stage 1:] mimics the back-scattering operation and deterministically reconstructs the intermediate field $\bm{\alpha}_{\sf\Lambda}$, the discrete form of:
    \begin{equation}\label{eqn:compute_alpha}
        {\bm{\alpha}_{\Lambda}}(\bm{y})=F^* \Lambda(\bm{y})=\int_{[0,2\pi]^2} e^{i\omega(\bm{r}-\bm{s})\cdot \bm{y}}\Lambda(r,s)\,dr\,ds\,,
    \end{equation}
    
    \item[Stage 2:] mimics the filtering process and draws a sample $\bm{\eta}$ using DDPM, as defined by:
    \begin{equation}\label{eqn:alpha_condition_reverse_SDE}
    d\bm{\eta} = \left[f(t)\bm{\eta} - g^2(t)\nabla_{\bm{\eta}}\log p_t(\bm{\eta}|\bm{\alpha}_{\sf\Lambda})\right] \, dt + g(t)\, dW_t^{\prime}\,,
    \end{equation}
    where $\nabla_{\bm{\eta}}\log p_t(\bm{\eta}|\bm{\alpha}_{\sf\Lambda})$ will be termed the physics-aware score function.
\end{itemize}
Proposition~\ref{prop:Fstar_invertible} states that $F^\ast$ is injective, thus it is invertible within its range. Assuming its discrete version $\sfF^\ast$ enjoys the same property, then the conditional distribution {satisfies} $p(\bm{\eta}|{\sf\Lambda})=p(\bm{\eta}|(\sfF^\ast)^{-1}\bm{\alpha}_{\sf\Lambda})$. Throughout the paper, we shorten the notation to be $p(\bm{\eta}|\bm{\alpha}_{\sf\Lambda})$.

This two-stage separation will be implemented in both the offline learning stage and the online sampling stage. In the offline learning stage, two neural networks will be independently developed to capture the back-scattering and filtering {processes} respectively. The composite neural network maps the given data ${\sf\Lambda}$ to the physics-aware score function $\nabla_{\bm{\eta}}\log p_t(\bm{\eta}|\bm{\alpha}_{\sf\Lambda})$. The weights of the entire neural network are learned from $p_\data$. Then, in the online stage,  given any ${\sf\Lambda}$, we can produce a sample $\bm{\eta}$ by running~\eqref{eqn:compute_alpha}-\eqref{eqn:alpha_condition_reverse_SDE} with the learned physics-aware score function $\nabla_{\bm{\eta}}\log p_t(\bm{\eta}|\bm{\alpha}_{\sf\Lambda})$.

To best follow the structure of the back-scattering and the filtering operators, we need to integrate the rotational equivariance and translational equivariance into the neural network architecture. To build in rotational equivariance to represent the back-scattering operator is straightforward, and we discuss it in Section~\ref{sec:back_proj_representation}. The translational equivariance of the filtering operator needs to be translated to the associated property for the score function, as seen in~\eqref{eqn:alpha_condition_reverse_SDE}, and is detailed in Section~\ref{sec:score_function_representation}. Finally, to enhance training efficiency and quality, we employ an off-the-shelf preconditioned framework~\cite{karras2022elucidating}, which is elaborated upon in Section~\ref{sec:architecture}. The diagram of the approach is illustrated in Figure~\ref{fig:diagram_approach}.

\subsection{Representation of the Back-Scattering Operator}\label{sec:back_proj_representation}
According to Proposition~\ref{prop2}, the back-scattering operator is {rotationally} equivariant. Therefore, we are to design a neural network, denoted as $F_{\bm{\Theta}_1}$, that preserves this rotational equivariance property to achieve:
\begin{equation}\label{eq:F_theta_1}
\bm{\alpha}_{\sf\Lambda}\approx F_{\bm{\Theta}_1}(\sf\Lambda)\,.\end{equation}
Several choices are available:

\paragraph{Uncompressed Rotationally Equivariant Model (EquiNet)}
The back-scattering operator for each frequency $\omega$, when expressed in polar coordinates $\bm{y} = (\rho\cos\theta,\rho\sin\theta)$, is formulated as:
\begin{equation}\label{eqn:polar_alpha}
[(F^\omega)^* \Lambda](\theta,\rho) = \int_{[0,2\pi]} \underbrace{e^{i\omega\rho\cos(r)}}_{\text{kernel } K^\omega}\bigg(\int_{[0,2\pi]}\underbrace{e^{-i\omega\rho\cos(s)}}_{\text{kernel } \overline{K^\omega}}\Lambda^\omega(r+\theta,s+\theta)\,ds\bigg)\,dr\,.
\end{equation}
A notable observation detailed in~\cite{equivariant} is that the rotational equivariance is preserved regardless of the form of $K^\omega$ and thus the integral kernel $K^\omega  = e^{i\omega\rho\cos(t)}$ can be replaced by any other function.  {Numerically, this integral kernel is modeled using trainable parameters, and the application of $(F^\omega)^*$ is then approximated by a neural network.} This whole approach of utilizing the formulation of~\eqref{eqn:polar_alpha} to preserve rotational equivariance is hence termed ``EquiNet.''

\paragraph{Compressed Rotationally Equivariant Model (B-EquiNet)} 
B-EquiNet is an extension of EquiNet and aims at reducing computational complexity, with the ``B'' standing for butterfly, drawing inspiration from the butterfly factorization~\cite{BF}. This factorization is an economical presentation of a two-dimensional function and saves memory costs. The authors in~\cite{equivariant} studied the butterfly structure of the integral kernel $K^\omega = e^{i\omega\rho\cos(t)}$ and integrated this structure in building the NN representation.

\begin{remark}
Other NN architectures have also been investigated in the literature, and we present a couple of choices:
\begin{itemize}
    \item Wideband Butterfly Network (WideBNet): WideBNet~\cite{MLZ}  leverages computational savings from both the butterfly factorization and Cooley-Tukey FFT~\cite{BF,Cooley_Tukey:1965}. The work examines the structure of the integral kernel shown in~\eqref{eqn:F_ast}, and approximates it using a full Butterfly Network while incorporating data at each frequency in a hierarchical fashion following the natural dyadic decomposition.
     
    \item SwitchNet: SwitchNet~\cite{Khoo_YingSwitchNet:2019} leverages the inherent low-rank properties of the problem. Specifically, sufficiently small square submatrices of the discrete back-scattering operator are numerically low-rank. This property inspires a low-complexity factorization of the operator, which can be viewed as an incomplete butterfly factorization.
\end{itemize}
Neither of these architectures preserves the rotational equivariance.
\end{remark}

\subsection{Representation of the Physics-Aware Score Function}\label{sec:score_function_representation}

Recall~\eqref{eqn:FBP} and the translational equivariance property (Proposition~\ref{prop3}) of the filtering operator, it is natural for us to believe that the score function in DDPM used to filter information in the intermediate media also needs to exhibit certain symmetric features. Given the complex relation between the map and the score function, it is not immediately clear how these features translate. We discuss the condition and symmetric properties needed for the score function in Section~\ref{sec:score_function_symmetry}. To numerically capture this property, we propose using a CNN-based representation. These numerical strategies are discussed in Appendix~\ref{sec:CNN}.

\subsubsection{Symmetry of the Physics-Aware Score Function}\label{sec:score_function_symmetry}
We show that when both the target conditional distribution $p_\data$ and the likelihood function $p_0(\alpha|\bm{\eta})$ are {translationally} invariant, the physics-aware score function $\nabla_{\bm{\eta}} \log p_t(\bm{\eta}|\bm{\alpha}_{\sf\Lambda})$ should be {translationally equivariant}. 

Notably, the classical definition of translational equivariance (as was presented in~\cite{equivariant}) applies to the continuous setting, whereas the score function pertains to discrete objects. We provide an analogous definition for translational symmetry in Definition~\ref{def:translational_symmetry} {and we provide the symmetry property of the score function in Theorem~\ref{thm:score_function_TE}}.

\begin{definition}[translation operator]\label{def:translation_operator}
For any $\bm{a}\in\N^2_\eta$, define the translation operator $ T_{\bm{a}} $ as a map between matrices $T_a : \R^{n_\eta\times n_\eta} \rightarrow \R^{n_\eta\times n_\eta}$ such that for any $v\in\R^{n_\eta\times n_\eta}$ and $\bm{y}\in\N_\eta^2$:
\begin{equation}
(T_{\bm{a}} v)_{\bm{y}} = v_{\tau_{-\bm{a}}(\bm{y})}\,,
\end{equation}
where $\tau_a : \N_\eta^2 \rightarrow \N_\eta^2 $ is the coordinates translation map:
\begin{equation}\label{eqn:translation_map}
\tau_{\bm{a}}(\bm{y}) = (\bm{y}+\bm{a}) \mod n_\eta\quad\forall \bm{y}\in\N_\eta^2\,.
\end{equation}
The modulo operation is applied element-wise.
\end{definition}

\begin{definition}\label{def:translational_symmetry}

A function $\mathcal{P}$ is said to be translationally invariant if the output does not change when its arguments are acted {upon} by $T_{\bm{a}}$ for any $\bm{a}\in\N_\eta^2$. Examples are:
\begin{itemize}
\item $\mathcal{P}: \R^{n_\eta\times n_\eta}\rightarrow \mathbb{R} $ acting on $v\in \R^{n_\eta\times n_\eta}$ is translationally invariant if
\begin{equation}
\mathcal{P}(T_{\bm{a}} v) = \mathcal{P}(v)\,, \quad \forall \bm{a} \in \N_\eta^2\,.
\end{equation}
    \item $\mathcal{P}: (\R^{n_\eta\times n_\eta})^{\otimes 2} \rightarrow \mathbb{R} $ acting on $ (v,w)\in (\R^{n_\eta\times n_\eta})^{\otimes 2}$ is translationally invariant if
\begin{equation}
\mathcal{P}(T_{\bm{a}} v, T_{\bm{a}} w) = \mathcal{P}(v,w)\,, \quad \forall \bm{a} \in \N_\eta^2\,.
\end{equation}
\end{itemize}

{An} operator $\mathcal{Q} : (\R^{n_\eta\times n_\eta})^{\otimes 2} \rightarrow \R^{n_\eta\times n_\eta} $ is said to be translationally equivariant if, for any $(v,w)\in (\R^{n_\eta\times n_\eta})^{\otimes 2}$
\begin{equation}
\mathcal{Q}(T_{\bm{a}} v,T_{\bm{a}} w) = T_{\bm{a}}\mathcal{Q}(v,w)\,, \quad \forall \bm{a} \in \N_\eta^2\,.
\end{equation}
\end{definition}
These definitions apply to discrete quantities (matrices on $\mathbb{R}^{n_\eta\times n_\eta}$), and they mimic those defined for continuous quantities~\cite{equivariant}. Specific attention should be paid to the $\mod$ operator in~\eqref{eqn:translation_map}, which suggests the use of periodic boundary condition for simulation.

\begin{thm}\label{thm:score_function_TE}
With $\bm{\alpha}$, $\bm{\eta}$ and $p_t$ defined above, if $\mathcal{P}(\bm{\alpha},\bm{\eta})=p_0(\bm{\alpha} | \bm{\eta})$ and $p_\data(\bm{\eta})$ are both translationally invariant, then the physics-aware score function $\nabla_{\bm{\eta}}\log p_t(\bm{\eta}|\bm{\alpha})$ is translationally equivariant. More specifically,
assume: $p_0(T_{\bm{a}}\bm{\alpha} | T_{\bm{a}}\bm{\eta})=p_0(\bm{\alpha} | \bm{\eta})$
and $p_\data(T_{\bm{a}}\bm{\eta}) = p_\data(\bm{\eta})$ for all $\bm{a} \in \N_\eta^2$, then for $p_t \in C^1(\mathbb{R}^{n_\eta \times n_\eta})^{\otimes 2}$ and $p_t > 0$:
\begin{equation}\label{eqn:score_function_TE}
\nabla_{\bm{\eta}}\log p_t(T_{\bm{a}}\bm{\eta}| T_{\bm{a}}\bm{\alpha}) = T_{\bm{a}}\nabla_{\bm{\eta}}\log p_t(\bm{\eta}|\bm{\alpha}) \quad \forall \bm{a} \in \mathbb{N}_\eta^2\,,
\end{equation}
i.e. $\mathcal{Q}(\bm{\eta},\bm{\alpha})=\nabla_{\bm{\eta}}\log p_t(\bm{\eta}|\bm{\alpha})$ is {translationally} equivariant.
    \begin{proof}
    The proof of this theorem is included in Appendix~\ref{sec:appendixB}.
    \end{proof}
\end{thm}
It should be noted that the assumption of translational invariance holds true in many cases. One such example occurs when the conditional probability $p_0(\bm{\alpha} | \bm{\eta})$ takes the form: $p_0(\bm{\alpha} | \bm{\eta}) \propto \exp{\left(\frac{1}{2\sigma^2} \|\sfF^*\sfF\bm{\eta} - \bm{\alpha}\|^2\right)}$, a variant of~\eqref{eqn:gaussian_misfit}.

\subsection{The Flowchart of the Training with Preconditioning}\label{sec:architecture}

{Section~\ref{sec:back_proj_representation} details the specific architectures used for numerical back-scattering. Furthermore, as suggested by Theorem~\ref{thm:score_function_TE}, we design a translationally equivariant CNN-based representation, as detailed in Appendix~\ref{sec:CNN}, for the filtering process. We now integrate these choices into the training process, as outlined below.}

Recall from Sections~\ref{sec:DDPMformulation}-\ref{sec:conditional_diffusion_model} that the denoiser is identified through the optimization formulation. Adapting this process to our context, we define the objective functional:
\begin{equation}
\mathcal{L}(D): D(\bm{\eta},{\sf\Lambda},\sigma)\to\mathbb{R}
\end{equation}
using
\begin{equation}
\mathcal{L}(D)=\mathbb{E}_{\sigma \sim p_{\text{train}}}\mathbb{E}_{(\bm{\eta},{\sf\Lambda}) \sim p_{\text{data}}} \mathbb{E}_{\bm{n} \sim \mathcal{N}(\,\cdot\,;\bm{0}, \sigma^2 I)} \left[\lambda(\sigma)\left\| D_{\bm{\Theta}}(\bm{\eta}+\bm{n} , {\sf\Lambda}, \sigma) - \bm{\eta}\right\|^2_2\right]\,,
\end{equation}
for a given noise level distribution $\sigma\sim p_{\text{train}}$ and a weight $\lambda(\sigma)$.

To conduct the minimization, we restrict ourselves to the feasible set of function space spanned by neural networks of the following form
\begin{equation}\label{eq:feasible_set}
    \begin{aligned}
        \mathcal{A} = \{D_{\bm{\Theta}}: \Theta&=\Theta_1\bigcup\Theta_2\,,\text{with}\\ &D_{\bm{\Theta}}(\bm{\eta}, {\sf\Lambda}, \sigma) = c_{\text{skip}}(\sigma)\bm{\eta} + c_{\text{out}}(\sigma) S_{\bm{\Theta}_2}\left(c_{\inc}(\sigma)\bm{\eta},F_{\bm{\Theta}_1}({\sf\Lambda}),c_{\text{noise}}(\sigma)\sigma\right)\}\,,
    \end{aligned}
\end{equation}
where $c_{\text{skip}},c_{\text{out}},c_{\inc}$, and $c_{\text{noise}}$ are predefined coefficients, termed as preconditioning of the network. $F_{\bm{\Theta}_1}$ represents the back-scattering component of the inversion, and either EquiNet or B-EquiNet will be deployed to code $F_{\bm{\Theta}_1}$, as was done in~\eqref{eq:F_theta_1}. Similarly, CNN-based representation will be used for $S_{\bm{\Theta}_2}$. This guarantees
$S_{\bm{\Theta}_2}(\bm{\eta}, \bm{\alpha}, \sigma)$ satisfies translational equivariance:
    \begin{equation}
        T_{\bm{a}} S_{\bm{\Theta}_2}(\bm{\eta}, \bm{\alpha}, \sigma) = S_{\bm{\Theta}_2}(T_{\bm{a}} \bm{\eta}, T_{\bm{a}} \bm{\alpha}, \sigma) \text{\quad for all $\bm{a} \in \mathbb{N}_\eta^2$}\,.
    \end{equation}
This choice of $S_{\bm{\Theta}_2}$ automatically guarantees that all functions in the feasible set~\eqref{eq:feasible_set} satisfy the translational equivariance for the denoiser, and as a consequence, the approximation to the score-function, see~\eqref{eq:denoiser_score}, is also {translationally equivariant}, as required by Theorem~\ref{thm:score_function_TE}. The whole NN architecture used to represent $D_{\bm{\Theta}}(\bm{\eta}, {\sf\Lambda}, \sigma)$ is summarized in Algorithm~\ref{alg:denoiser}:
\begin{algorithm}[h!]
\caption{Neural Architecture of the Denoiser}\label{alg:denoiser}
\begin{algorithmic}[1]
\Procedure{$D_{\bm{\Theta}}$}{$\bm{\eta}, {\sf\Lambda}, \sigma$}
  \State $\bm{\alpha} \gets  F_{\bm{\Theta}_1}(\sf\Lambda)$
  \State $\bm{s} \gets S_{\bm{\Theta}_2}\left(c_{\inc}(\sigma)\bm{\eta},\bm{\alpha},c_{\text{noise}}(\sigma)\sigma\right)$
  \State \Return $c_{\text{skip}}(\sigma)\bm{\eta} + c_{\text{out}}(\sigma)\bm{s}$
\EndProcedure
\end{algorithmic}
\end{algorithm}

\section{Numerical Examples}\label{sec:numerical_examples} 
The architecture for our {Back-Projection Diffusion} model is factorized into two neural networks applied in tandem: $F_{\bm{\Theta}_1}$ and $S_{\bm{\Theta}_2}$. This motivates us to name the models by joining the names of each component. Our main models are called \textbf{EquiNet-CNN} and \textbf{B-EquiNet-CNN}, where the latent intermediate field representation $\bm{\alpha}_{\sf\Lambda} \approx F_{\bm{\Theta}_1}({\sf\Lambda})$ is instantiated by EquiNet and B-EquiNet models (discussed in Section~\ref{sec:back_proj_representation}), and $S_{\bm{\Theta}_2}$ is instantiated by a CNN-based representation (introduced in Appendix~\ref{sec:CNN}). {We have made the code and datasets publicly available in a GitHub repository\footnote{\url{https://github.com/borongzhang/back_projection_diffusion}}.}

{We} present the training/optimization formulation in Appendix~\ref{sec:formulation_optimization}, and we introduce the evaluation metrics in Section~\ref{sec:metrics}. We provide details on the datasets that are used for training in Section~\ref{sec:datasets}. In Section~\ref{sec:baselines}, we introduce the {state-of-the-art} ML-based deterministic and classical methods that we use as baselines to benchmark our methodology. {For details on the software and hardware stack used for the experiments, see Appendix~\ref{sec:soft_hardware}.}

We perform an extensive suite of benchmarks to demonstrate the properties of our framework as mentioned in the introduction. In what follows, we summarize each of the benchmarks.

\textbf{Performance Comparison} (Section~\ref{sec:comparison_vs_deterministic}): We show that EquiNet-CNN and B-EquiNet-CNN considerably outperform other state-of-the-art deterministic methods and classical methods on the three synthetic datasets. 

\textbf{Parameter Efficiency and Performance across Resolutions} (Section~\ref{sec:complexity_resolution}): We demonstrate that the number of trainable parameters in EquiNet-CNN and B-EquiNet-CNN scales favorably with increasing resolution (and the number of unknowns to reconstruct), showcasing their parameter efficiency, which refers to the ability of a model to achieve high performance with a relatively small number of parameters.
EquiNet-CNN achieves high reconstruction accuracy, even for a challenging MRI Brain dataset (\cite{fastmri1,fastmri2}). We show that the quality of reconstruction increases as the resolution of the training data (and the probing frequency) increases, resulting in images with more fine-grained details.

\textbf{Sample Complexity} (Section~\ref{sec:sample_complexity}): We highlight the low-sample complexity of EquiNet-CNN by training the model with different training {datasets} with increasing {numbers} of samples. Remarkably, when trained on only 2000 data points, it achieves higher accuracy on a dataset with strong multiple back-scattering than the deterministic baselines.

\textbf{Posterior Distribution} (Section~\ref{sec:posterior_distribution}): We demonstrate that EquiNet-CNN captures the posterior distribution well, with the Relative Root Mean Square Error (RRMSE) between ground truth data and data sampled from the far-field patterns showing that the modes of the error distribution align with manual pixel-level adjustments of scatterers.

\textbf{Cycle Skipping} (Section~\ref{sec:cycle_skipping}): We showcase the stability of EquiNet-CNN by training it using only data at the highest frequency. We demonstrate that the cycle skipping phenomenon that often affects classical methods is noticeably mitigated by EquiNet-CNN.

\textbf{Ablation Study} (Section~\ref{sec:ablation_study}): We present an ablation study using different combinations of back-scattering architectures and representations for the physics-aware score function. 
EquiNet-CNN and B-EquiNet-CNN demonstrate a lower number of parameters while {achieving competitive performance based on the prescribed metrics.}

\textbf{Inverse Crime and Noisy Inputs} (Section~\ref{sec:noise_inverse_crime}): We demonstrate the robustness of EquiNet-CNN against noise and varying levels of epistemic uncertainty in data. To avoid the infamous inverse crime, we show that our methods {produce} high-quality reconstruction even when the input data was generated using solvers with different stencils and when stochastic noise was added to it. 

\textbf{Mixed Dataset and Generalization} (Section~\ref{sec:mixed_data}): When trained with all three synthetic datasets mixed together, we show that EquiNet-CNN can generate samples for each dataset with high accuracy, significantly outperforming other baseline deterministic models. We also evaluate the performance of EquiNet-CNN on out-of-distribution datasets, where the class of scatterers in the dataset is not present in the within-distribution training dataset.

\subsection{Metrics}\label{sec:metrics}
In this part, we present the metrics that we used to measure the error of our results. For {a} detailed description of the {metrics}, see Appendix~\ref{appendix:metrics}

\begin{itemize}
    \item Relative Root Mean Square Error (RRMSE): This metric quantifies the relative misfit between the generated samples and the ground truth for each element from the testing set. The average is then taken across the testing set.
    
    \item Sinkhorn Divergence (SD): An optimal transport-based metric, the Sinkhorn Divergence computes the distance between the ground truth's distribution and the estimated distribution. It involves computing a reference distance between the training set and the testing set, which is then compared to the distance between the testing set and the generated samples.
    
    \item Mean Energy Log Ratio (MELR): This metric assesses the quality of our samples by measuring the log-ratio of the energy spectrum (via Fourier modes) between the generated samples and the ground truth.
    
    \item Continuous Ranked Probability Score (CRPS): Used to measure the accuracy of our probabilistic model, this metric computes the difference between the estimated probabilities and the actual outcomes in the ground truth.
\end{itemize}

\subsection{Datasets}\label{sec:datasets}
The datasets consist of pairs of perturbation and scattering data $(\bm{\eta}, {\sf\Lambda})$ from different distributions. The perturbations are sampled randomly from a predetermined distribution, and their corresponding far-field patterns at three different frequencies, following a dyadic decomposition, are obtained by solving~\eqref{eq:scattereqn} numerically. 

The physical domain for the perturbations and intermediate fields was $[-0.5,0.5]^2$ discretized with {an} equispaced grid of $80 \times 80$ points. The {differential} operators were discretized with a tensorized finite {difference} stencil of $8$-th order in each {dimension}, and the radiation boundary {condition} was implemented using the perfectly matched layer (PML)~\cite{Berenger:PML} of order 2 and intensity 80. The sparse linear system was solved using a sparse LU factorization via UMFPACK~\cite{matlabprimer8}. The wideband data was sampled with monochromatic plane waves with frequencies of 2.5, 5, and 10, see Section~\ref{sec:preliminary}, for which the effective wavelength is 8 points per wavelength (PPW). In particular, we use $n_\sca=80$ receivers and sources, where the {receivers' geometry is} aligned with the directions of sources, i.e. 80 equiangular directions.

In our experiments, we first evaluate the effectiveness of the models using 3 different categories of synthetic perturbations: Shepp-Logan, 3-5-10h Triangles, and 10h Overlapping Squares, which {cover} most of the challenges from inverse problems: strong reflections hiding internal structure, small scatterers with features below Nyquist-Sampling rate, {and} scatterers exhibiting strong multiple back-scattering (i.e., waves bouncing back several times).

\begin{itemize}
\item Shepp-Logan: The well-known Shepp-Logan phantom,  which was created in 1974 by Larry Shepp and Benjamin F. Logan to represent a human head~\cite{SL}. The medium has a strong discontinuity modeling an uneven skull, which produced a strong reflection, which in return renders the recovery of the interior features challenging for classical methods.  The perturbations are generated based on randomly chosen scalings, densities, positions, and orientations for the phantoms~\cite{chungshepplogan}. 

\item 3-5-10h Triangles: Right triangles of side length $3$, $5$ and $10$ pixels, which are randomly located and oriented, and it is possible for them to overlap with each other. In this case we test the capacity of the algorithm to image consisting of small scatterers that are slightly below sub-Nyquist in size. The number of triangles is chosen randomly from 1 to 10.

\item 10h Overlapping Squares: 20 overlapping squares of side length $10$ pixels. 
\end{itemize}

\begin{figure}[h!]
    \centering
    \includegraphics[width=0.9\textwidth]{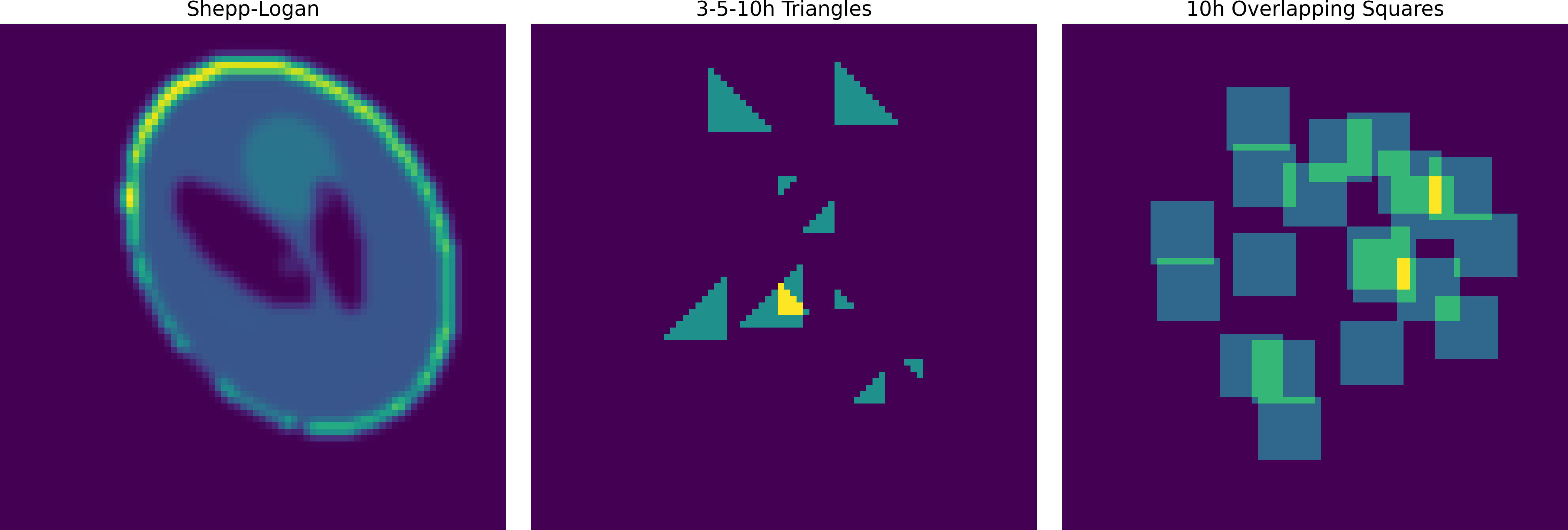} 
    \caption{Examples of the three synthetic perturbations (Shepp-Logan, 3-5-10h Triangles, and 10h Overlapping Squares) used to benchmark the models.}
    \label{fig:synthetic_scatterers}
\end{figure}

Figure~\ref{fig:synthetic_scatterers} showcases one example for each of the three {categories}. 

In addition, we study the scaling of the number of trainable parameters and performance for different resolutions of EquiNet-CNN on the NYU fastMRI Brain data. The Brain MRI images used as our perturbations are obtained from the NYU fastMRI Initiative database~\cite{fastmri1,fastmri2}. We padded, resized, and normalized the perturbations to a native resolution at $n_\eta = 240$ points representing the same physical domain $[-0.5,0.5]^2$.  Then, we down-sampled the perturbations to resolutions $n_\eta = 60, 80, 120$, and $160$.  For the perturbations at resolution $n_\eta = 60$, using the same method as introduced in the beginning of this section, we generated the far-field patterns discretized with $n_\sca=60$ at frequencies 3, 6, and 12, for which the effective wavelength is 5 PPW.  For perturbations of different resolutions, we generated their far-field patterns with $n_\sca=n_\eta$ by sampling at proportionally scaled frequencies, which resulted in the same effective wavelength.  More specifically, for resolutions at $n_\eta = 80, 120$ and $160$, we chose frequencies 4,8, and 16, frequencies 6,12, and 24, and frequencies 8,16, and 32 respectively. 

\begin{figure}[h!]
    \centering
    \includegraphics[width=0.9\textwidth]{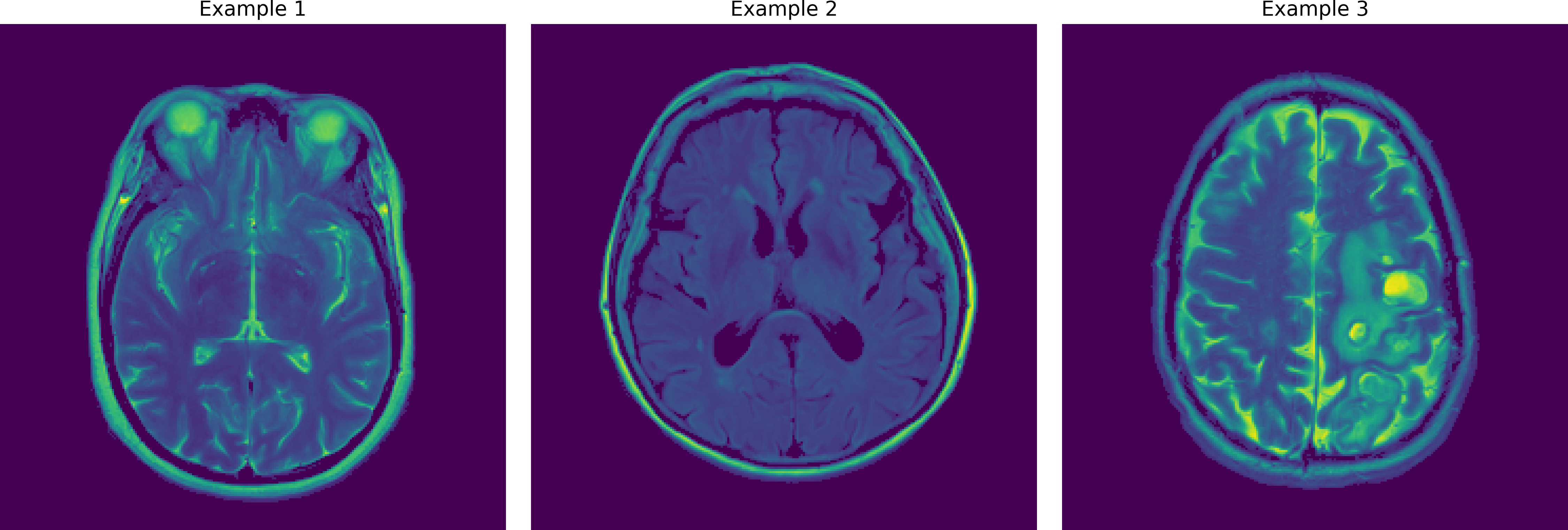}
    \caption{Examples of the Brain MRI perturbations.} 
    \label{fig:brainmri}
\end{figure}

Figure~\ref{fig:brainmri} showcases three examples of the Brain MRI perturbations at the native resolution $240\times240$. 

{We have made the datasets publicly available\footnote{\url{https://doi.org/10.5281/zenodo.14911327}, \url{https://doi.org/10.5281/zenodo.14760123}}.}

\subsection{Baselines} \label{sec:baselines}
We use four state-of-the-art deterministic baselines when comparing our framework, which also leverage the filtered back-projection formula in~\eqref{eqn:FBP}, and they approximate the back-scattering operator and the filtering operator by neural networks. In particular, they all use a CNN to represent the filtering operator, respecting its translational equivariance, see Proposition~\ref{prop3}. They differ primarily in their representations of the back-scattering operator.  We briefly recap their features and detail the training specifics:

\begin{itemize}
    \item SwitchNet~\cite{Khoo_YingSwitchNet:2019} uses an incomplete Butterfly factorization to derive a low-complexity factorizaiton of the back-scattering operator, which is then replaced by a neural network.
    \item WideBNet~\cite{MLZ} utilizes the butterfly factorization and Cooley-Tukey FFT algorithm to design its neural network. 
    \item EquiNet~\cite{equivariant} relies on a change of variable from the integral representation of the scattering operator to write a rotationally equivariant network.
    \item B-EquiNet~\cite{equivariant} follows the same structure of EquiNet, but it relies on a Butterfly network to compress the operators.
\end{itemize}

For deterministic models, we decorate their model name with `(deterministic)' and refer to them as EquiNet (deterministic), B-EquiNet (deterministic), WideBNet (deterministic) and SwitchNet (deterministic). {We have made the code publicly available in a GitHub repository\footnote{\url{https://github.com/borongzhang/ISP_baseline}}.}

\paragraph{Optimization and Hyperparameters}
The deterministic models are trained to minimize the mean square error between the network-produced  perturbations and the ground truth perturbations (used to generate the input data), i.e.,
\begin{equation}
\min_{\bm{\Theta}} \frac{1}{N_s}\sum_{s =1}^{N_s} \| \Phi_{\bm{\Theta}}({\sf\Lambda}^{[s]}) - \eta^{[s]} \|^2\,.
\end{equation}  

For training both SwitchNet (deterministic) and WideBNet (deterministic), the initial learning rate was set as $1 \times 10^{-3}$ and the scheduler was set as Optax’s \texttt{exponential\_decay}~\cite{deepmind2020jax} with a decay rate of 0.95 after every 2000 transition steps, with staircase set to true. The Adam optimizer~\cite{adam} is employed, and we terminate training after 150 epochs. Additionally, we trained both EquiNet (deterministic) and B-EquiNet (deterministic) for 35 epochs using the Adam optimizer with Optax’s \texttt{warmup\_cosine\_decay}~\cite{deepmind2020jax} as our scheduler. The initial learning rate was set to $1 \times 10^{-5}$, gradually increased to a peak of $5 \times 10^{-3}$ over the first 2000 steps, and then decayed to $1 \times 10^{-7}$ by the end of training. 

In addition to the ML-based approach, we also considered PDE-constrained optimization approaches: 

\begin{itemize}
    \item Least Squares~\cite{Colton_Kress:Inverse_Acoustic_and_Electromagnetic_Scattering_Theory}: It uses the Born approximation to fix the background, it then finds the perturbation that minimizes the data misfit in \eqref{eq:data_misfit} with respect to the fixed background. Additionally, in presence of the wideband data, it minimizes the sum of data misfits at all frequencies. 
    \item Full Waveform Inversion (FWI)~\cite{Tarantola:Inversion_of_seismic_reflection_data_in_the_acoustic_approximation}: Similarly to the least-squares approach, it minimizes the same data misfit in \eqref{eq:data_misfit}, but it allows the background to be updated at each iteration. In presence of the wideband data, the optimization process the data hierarchically starting from the lowest frequency and slowly starting to process data at higher frequencies. We performed a sweep of different schedules with different combination of frequencies that provide the best reconstruction.
\end{itemize}

\subsection{Performance Comparison} \label{sec:comparison_vs_deterministic}

We compare reconstructions {of} our three synthetic datasets (Shepp-Logan, 3-5-10h Triangles, and 10h Overlapping Squares) using our main models, EquiNet-CNN and B-EquiNet-CNN, alongside the baseline models introduced in Section~\ref{sec:baselines}. The models are benchmarked using the metrics RRMSE, MELR, and SD {as defined in Section~\ref{sec:metrics}}

Table~\ref{tab:rrmse_melr_results} summarizes the performance of each model on these datasets, from which we observe that EquiNet-CNN and B-EquiNet-CNN considerably outperform {the state-of-the-art ML deterministic methods as well as the classical methods.}

\begin{table}[h!]
\centering
\begin{tabular}{c|c|c|c}
\toprule
Model & RRMSE  & MELR  & SD \\
&  & $(\times 10^{-2})$ & \\
\midrule
\multicolumn{4}{c}{Shepp-Logan (Reference SD: 14.406)} \\
\midrule
EquiNet-CNN               & \textbf{1.323\%} & \textbf{1.564} & \textbf{3.734} \\
B-EquiNet-CNN             & 1.406\% & 1.757 & 3.754 \\
EquiNet (deterministic)   & 1.693\% & 2.827 & 3.786 \\
B-EquiNet (deterministic) & 2.022\% & 2.906 & 3.831 \\
WideBNet (deterministic)  & 3.843\% & 13.255 & 4.085 \\
SwitchNet (deterministic) & 4.305\% & 8.071 & 4.147 \\
FWI                       & 52.041\% & 241.893 & 8.893 \\
Least Squares             & 154.645\% & 408.039 & 21.774\\
\midrule
\multicolumn{4}{c}{3-5-10h Triangles (Reference SD: 2.833)} \\
\midrule
EquiNet-CNN               & \textbf{1.590\%} & 1.385 & \textbf{0.949} \\
B-EquiNet-CNN             & 1.657\% & \textbf{1.318} & 0.952 \\
EquiNet (deterministic)   & 2.741\% & 1.734 & 0.987\\
B-EquiNet (deterministic) & 2.944\% & 2.010 & 0.990 \\
WideBNet (deterministic)  & 17.263\% & 10.582 & 1.294 \\
SwitchNet (deterministic) & 15.084\% & 9.377 & 1.256 \\
FWI                       & 28.637\% & 159.894 & 1.501 \\
Least Squares             & 41.666\% & 81.908 & 14.391\\
\midrule
\multicolumn{4}{c}{10h Overlapping Squares (Reference SD: 11.183)} \\
\midrule
EquiNet-CNN               & \textbf{1.744\%} & \textbf{1.979} & \textbf{3.860} \\
B-EquiNet-CNN             & 2.046\% & 2.683 & 3.894 \\
EquiNet (deterministic)   & 10.891\% & 25.906 & 4.881 \\
B-EquiNet (deterministic) & 9.484\%  & 21.434 & 4.727 \\
WideBNet (deterministic)  & 14.327\% & 44.182 & 5.260 \\
SwitchNet (deterministic) & 20.102\% & 24.295 & 5.917 \\
FWI                       & 38.777\% & 281.057 & 7.948 \\
Least Squares             & 163.037\% & 301.991 & 17.603\\
\bottomrule
\end{tabular}
\caption{Comparison of model performance on three synthetic datasets (Shepp-Logan, 3-5-10h Triangles, and 10h Overlapping Squares) using the metrics RRMSE, MELR, and SD. The table highlights the superior performance of EquiNet-CNN and B-EquiNet-CNN compared to various deterministic methods and classical approaches. The best performance metrics for each dataset are indicated in bold.}
\label{tab:rrmse_melr_results}
\end{table}

\begin{figure}[h!]
    \centering
    \includegraphics[width=0.9\textwidth]{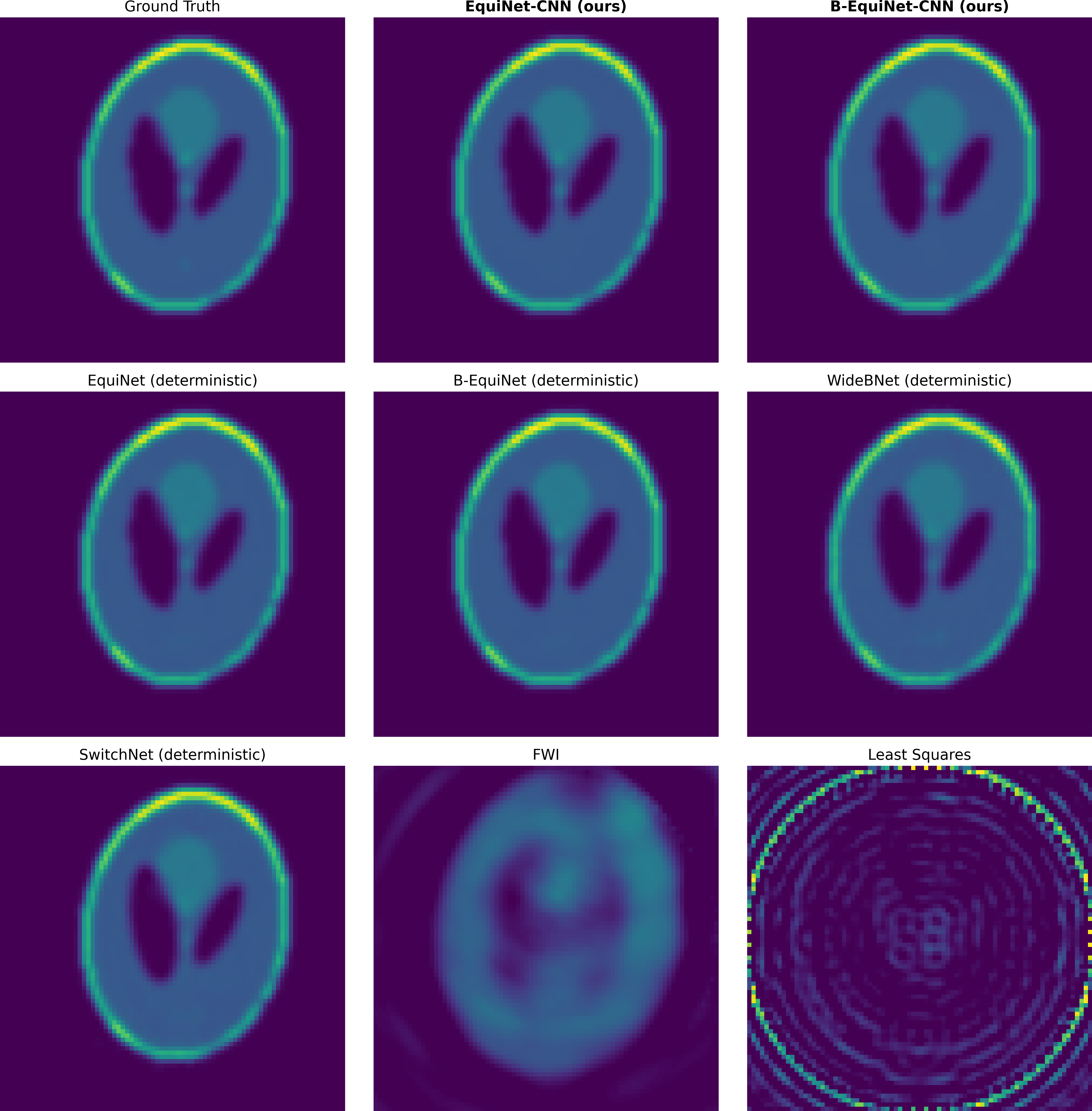}
    \caption{Model reconstructions from EquiNet-CNN, B-EquiNet-CNN, and baselines for the Shepp-Logan dataset. See Figure~\ref{fig:comparison_shepplogan} for a detailed zoom-in comparison.}
    \label{fig:reconstruction_shepplogan}
\end{figure}

\begin{figure}[h!]
    \centering
        \includegraphics[width=0.9\textwidth]{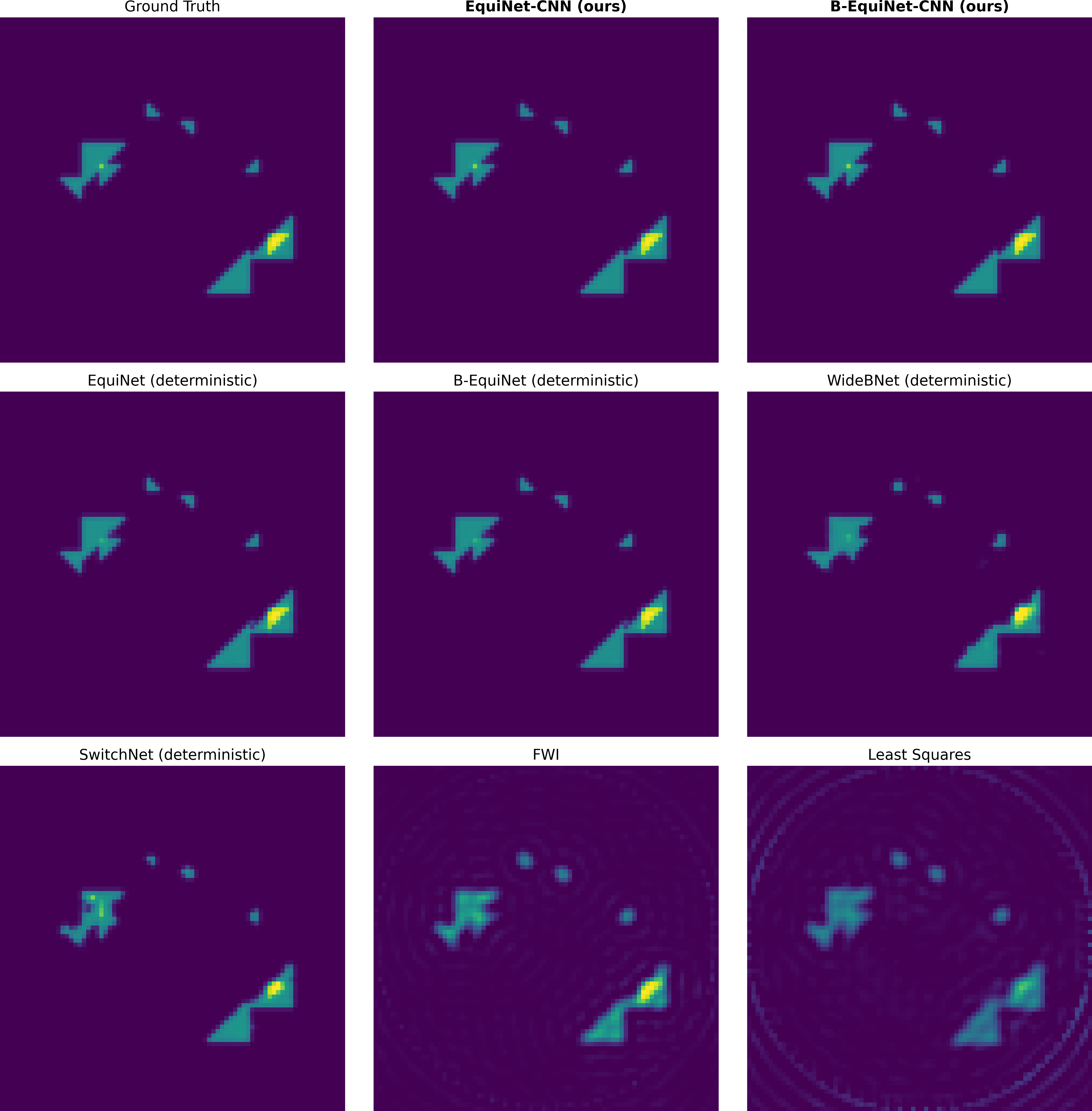}
        \caption{Model reconstructions from EquiNet-CNN, B-EquiNet-CNN, and baselines for the 3-5-10h Triangles dataset. See Figure~\ref{fig:comparison_3510triangles} for a detailed zoom-in comparison.}
    \label{fig:reconstruction_3510triangles}
\end{figure}

\begin{figure}[h!]
    \centering
        \includegraphics[width=0.9\textwidth]{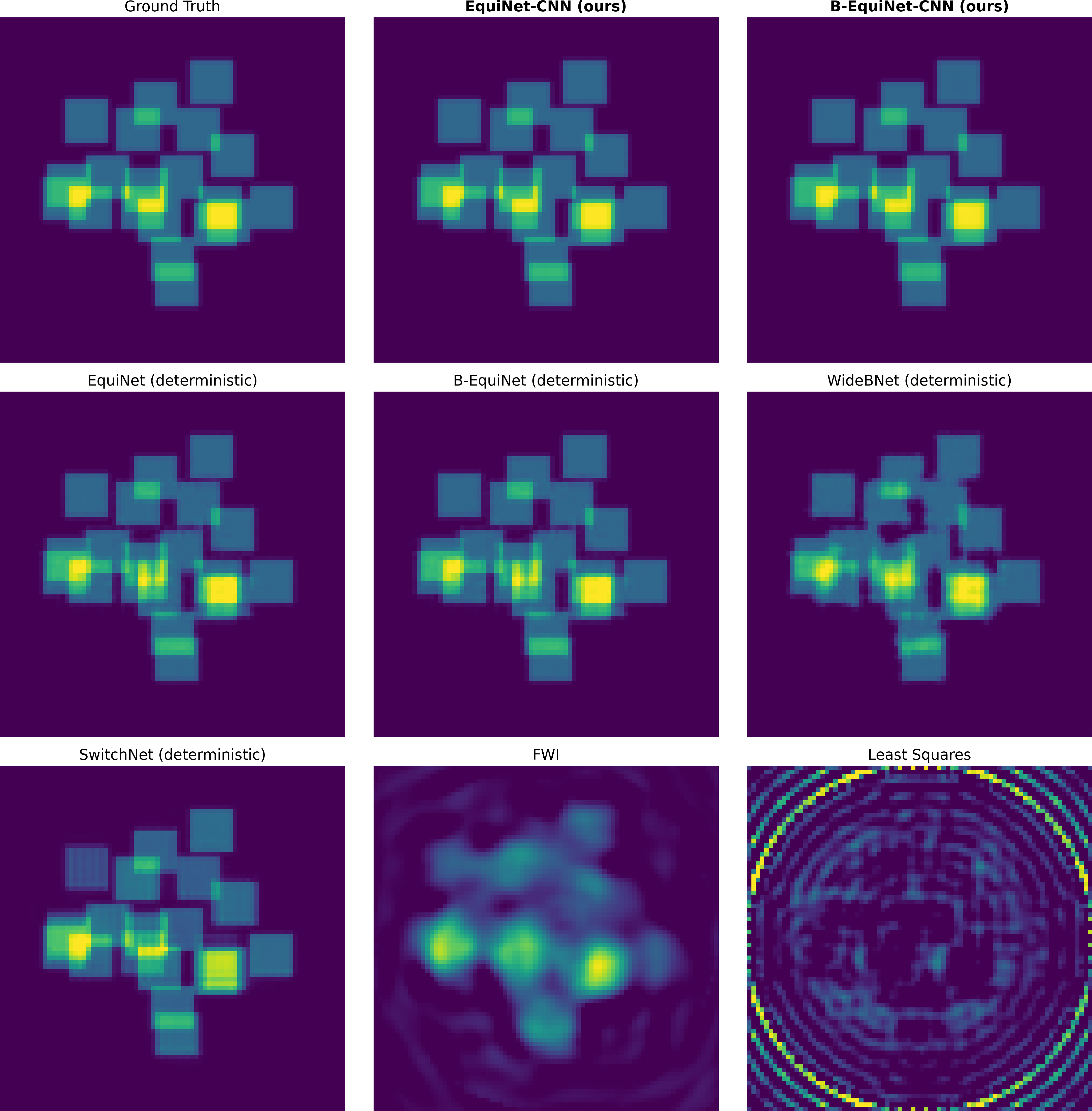}
    \caption{Model reconstructions from EquiNet-CNN, B-EquiNet-CNN, and baselines for the 10h Overlapping Squares dataset. See Figure~\ref{fig:comparison_10hsquares} for a detailed zoom-in comparison.}
    \label{fig:reconstruction_10hsquares}
\end{figure}

Additionally, Figures~\ref{fig:reconstruction_shepplogan}, \ref{fig:reconstruction_3510triangles}, and \ref{fig:reconstruction_10hsquares} visually showcase model reconstructions on three synthetic datasets: Shepp-Logan, 10h Overlapping Squares, and 3-5-10h Triangles. These figures initially present plots of the ground truth and reconstructions. Notably, the reconstructions from Least Squares and Full Waveform Inversion have markedly lower quality than the {others}, prompting a further detailed comparison using region of interest (ROI) plots. 

\begin{figure}[h!]
    \centering
    \includegraphics[width=0.75\textwidth]{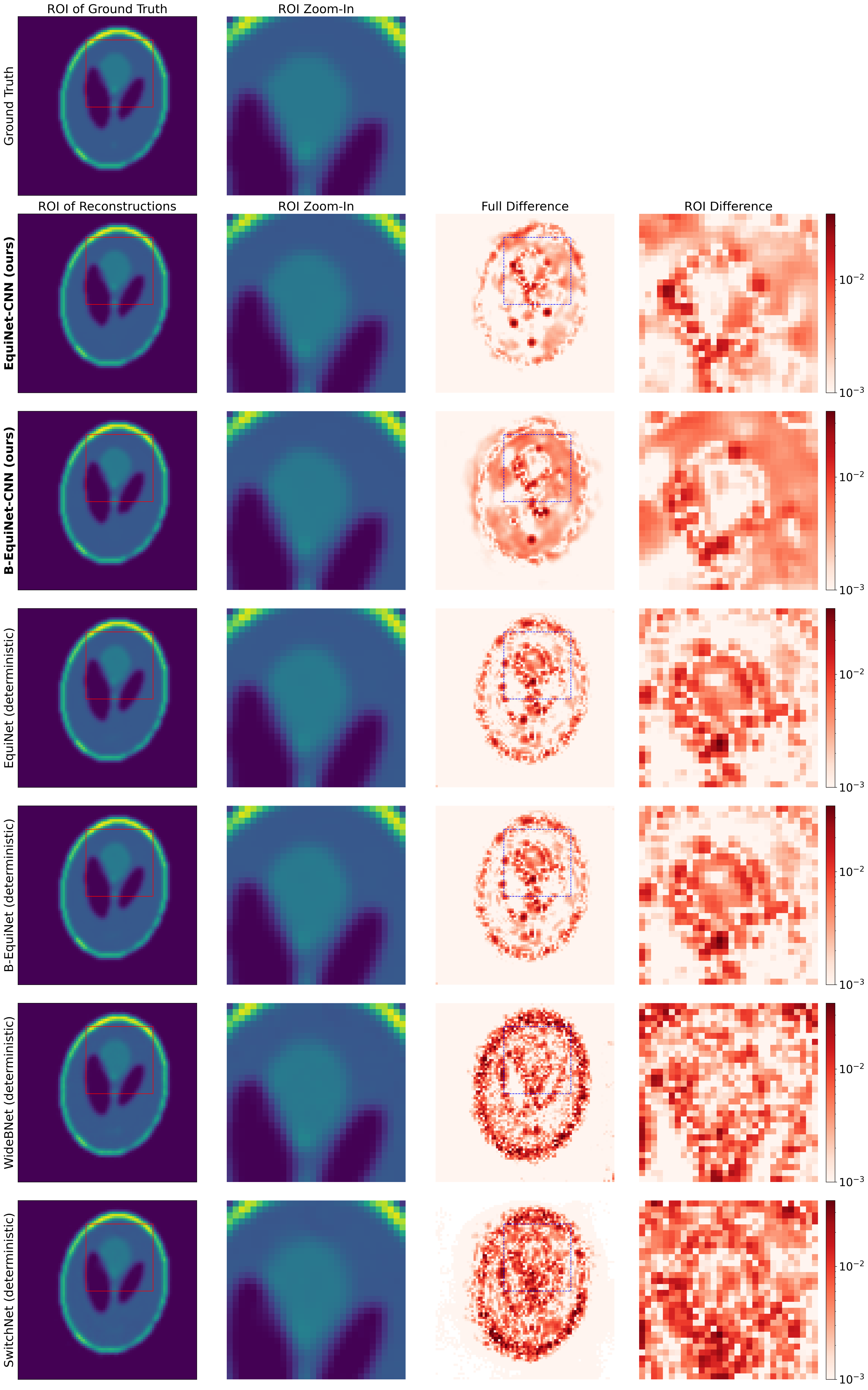}
    \caption{Model reconstructions for the Shepp-Logan dataset, showcasing regions of interest (ROI), the detailed zoom-ins and differences for EquiNet-CNN, B-EquiNet-CNN, and baseline deterministic models.}
    \label{fig:comparison_shepplogan}
\end{figure}

\begin{figure}[h!]
    \centering
        \includegraphics[width=0.75\textwidth]{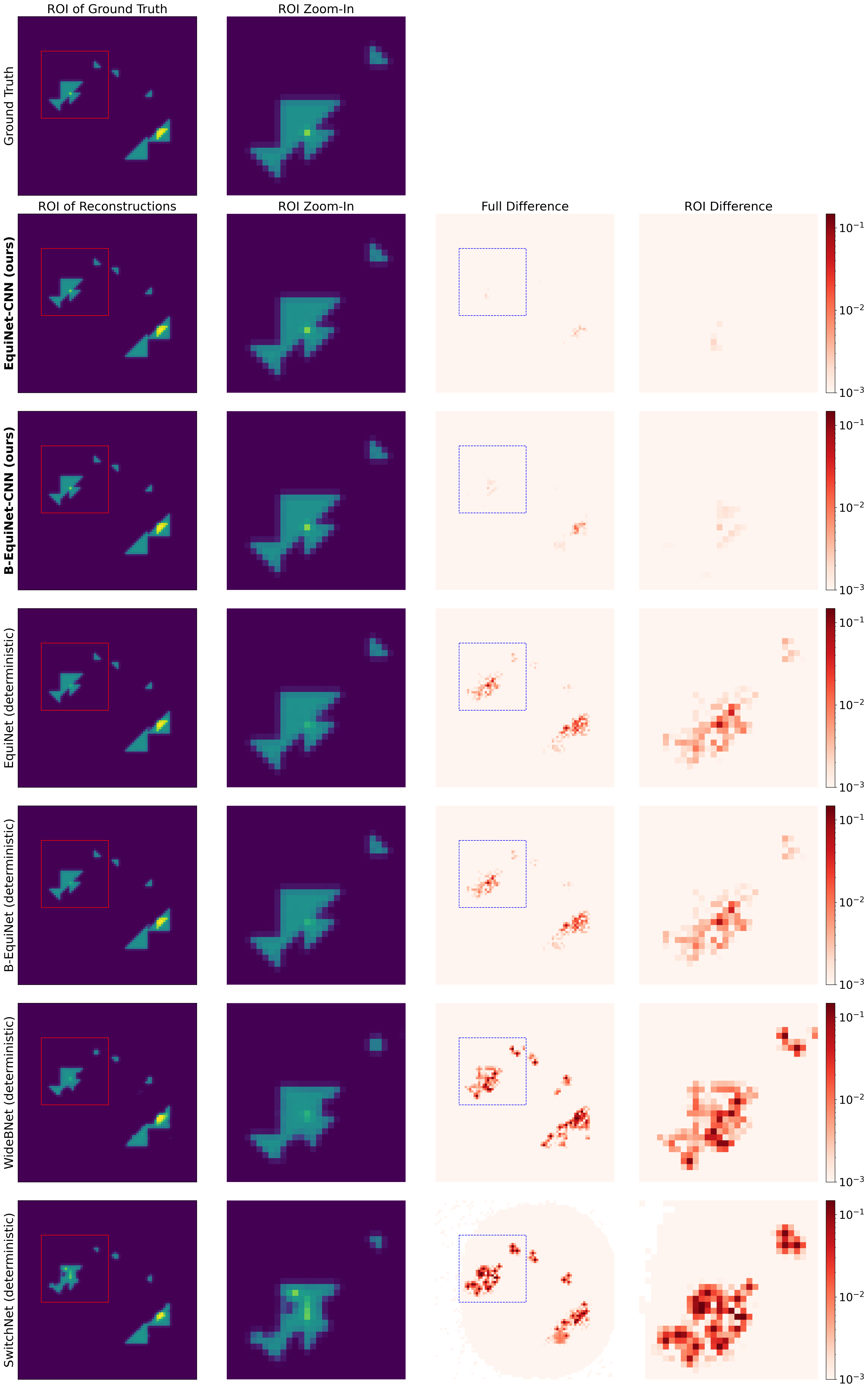}
        \caption{Model reconstructions for the 3-5-10h Triangles dataset, showcasing regions of interest (ROI), the detailed zoom-ins and differences for EquiNet-CNN, B-EquiNet-CNN, and baseline deterministic models.}
    \label{fig:comparison_3510triangles}
\end{figure}

\begin{figure}[h!]
    \centering
        \includegraphics[width=0.75\textwidth]{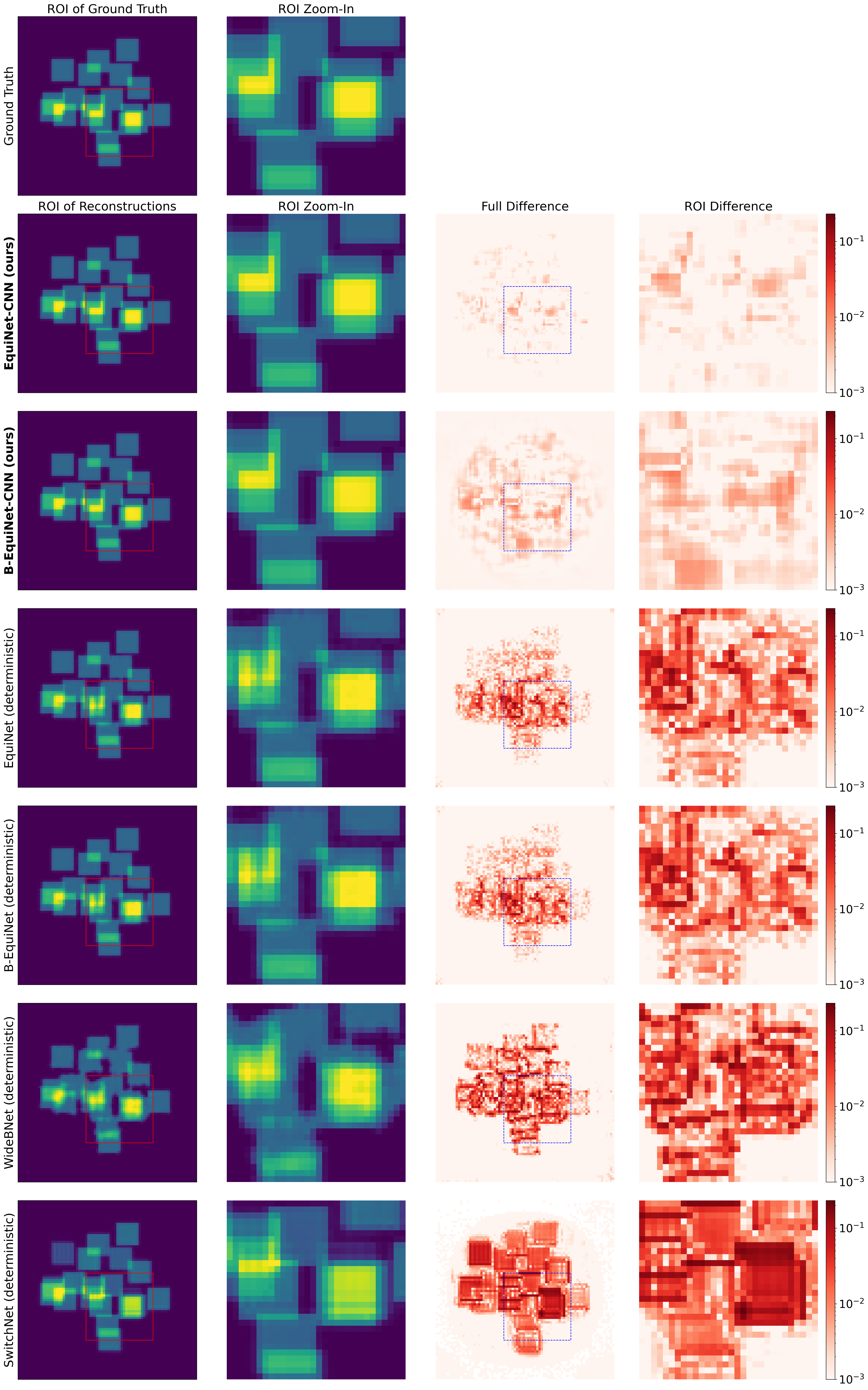}
    \caption{Model reconstructions for the 10h Overlapping Squares dataset, showcasing regions of interest (ROI), the detailed zoom-ins and differences for EquiNet-CNN, B-EquiNet-CNN, and baseline deterministic models.}
    \label{fig:comparison_10hsquares}
\end{figure}

Figures~\ref{fig:comparison_shepplogan}, \ref{fig:comparison_3510triangles}, and \ref{fig:comparison_10hsquares} focus on ROIs to highlight finer and more subtle differences between the reconstructions stemming from EquiNet-CNN, B-EquiNet-CNN, and the deterministic models. The first row of subplots presents the ROI of the ground truth alongside one ROI zoom-in. Subsequent rows display the ROI of reconstructions, ROI zoom-ins, Full Differences, and ROI Differences for each model. EquiNet-CNN and B-EquiNet-CNN exhibit considerably lower errors compared to other state-of-the-art deterministic models.

\subsection{Parameter Efficiency and Performance across Resolutions}\label{sec:complexity_resolution}

In the design\footnote{See Section~\ref{sec:architecture}.} of EquiNet-CNN and the further compressed B-EquiNet-CNN, $S_{\bm{\Theta}_2}$ is a CNN-based representation that maintains a constant number of trainable parameters across all resolutions. Specifically, in our experiments, the CNN-based representation has 374,575 trainable parameters. Therefore, the asymptotic scaling of the number of trainable parameters in the models is entirely determined by $F_{\bm{\Theta}_1}$. The {latter} is summarized in Table~\ref{table:complexity_comparison}, and written relative to $N_{\text{unknown}} \defeq n_\eta^2$, which is the number of grid points used in the  reconstruction after the physical domain is discretized. In our experiments, we use a discretization of the same size for the scattering data, i.e., $N_{\text{unknown}} = n_\eta^2 = n_\sca^2$ (see Section~\ref{sec:discretization}). Additionally, it should be noted that a fixed number of frequencies (in our case, three) {is} used to generate data at all resolutions.

\begin{table}[h!]
\centering
\begin{tabular}{l|c|c}
 \hline
 Complexity & EquiNet-CNN & B-EquiNet-CNN \\
 \hline
 \# Parameters & $\mathcal{O}(N_{\text{unknown}})$ & $\mathcal{O}(\sqrt{N_{\text{unknown}}}\log{N_{\text{unknown}}})$ \\
 \hline
\end{tabular}
\caption{Scaling of the number of trainable parameters with respect to the number of unknowns for B-EquiNet-CNN and EquiNet-CNN.}
\label{table:complexity_comparison}
\end{table}

We test the performance of EquiNet-CNN on the MRI Brain datasets at resolutions $n_\eta=60, 80, 120,$ and $160$. Table~\ref{tab:equinet_cnn_mri_brain} shows the numbers of trainable parameters of $F_{\bm{\Theta}_1}$ and $S_{\bm{\Theta}_2}$, denoted as $|\bm{\Theta}_1|$ and $|\bm{\Theta}_2|$ respectively, as well as the validation RRMSE at the training resolutions. In particular, Table~\ref{tab:equinet_cnn_mri_brain} shows that the number of trainable parameters {scales} sublinearly with the total number of unknowns to reconstruct, demonstrating the model's parameter efficiency. In addition, for the MRI Brain dataset, EquiNet-CNN achieves high accuracy in the reconstruction, which further improves as the resolution of the data (and the frequency of the probing waves) {increases}.

\begin{table}[h!]
\centering
\begin{tabular}{lccc}
\toprule
Resolution ($n_\eta$) & $|\bm{\Theta}_1|$ & $|\bm{\Theta}_2|$  & RRMSE \\ 
\midrule
60  & 87,840  & 374,575  &  5.363\% \\
80  & 155,520 & 374,575  &  5.425\% \\
120 & 348,480 & 374,575  &  5.062\% \\
160 & 618,240 & 374,575  &  4.544\% \\
\bottomrule
\end{tabular}
\caption{Number of trainable parameters and performance of EquiNet-CNN on MRI Brain datasets at different resolutions. Note that $|\bm{\Theta}_2|$ is a constant 374,575 at all resolutions and $|\bm{\Theta}_1| = \mathcal{O}(N_{\text{unknown}})$. }\label{tab:equinet_cnn_mri_brain}
\end{table}

\begin{figure}[h!]
    \centering
    \includegraphics[width=0.75\textwidth]{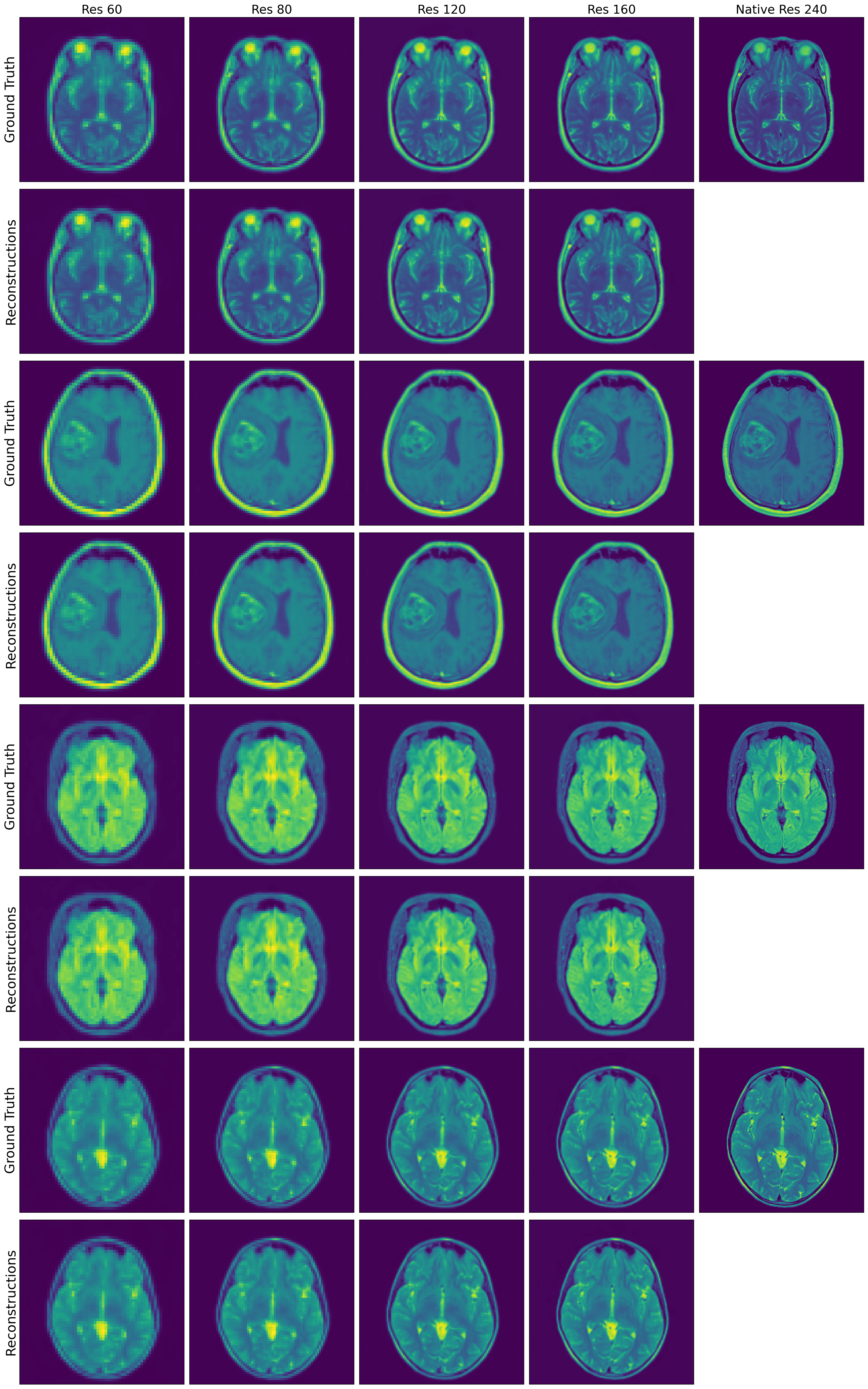}
    \caption{Ground truth perturbations and reconstructions from the MRI Brain dataset are displayed at resolutions 60, 80, 120, and 160. The rightmost column shows the ground truth perturbations at their native resolution of 240.}
    \label{fig:mri_reconstructions}
\end{figure}

Figure~\ref{fig:mri_reconstructions} plots four different ground truth perturbations at resolutions $n_\eta=60,80,120$ and $160$ from the MRI Brain dataset, alongside their reconstructions using EquiNet-CNN. Even-numbered rows showcase ground truth images at native resolution 240 and its downsampled versions at 60, 80, 120, and 160 resolutions. The subsequent odd-numbered rows present corresponding reconstructions by EquiNet-CNN at these resolutions. 

\subsection{Sample Complexity}\label{sec:sample_complexity}
By exploiting the symmetries of the problem, specifically, the rotational equivariance of the back-scattering operator for the latent representation and the translational equivariance of the physics-aware score function, we significantly reduce the number of trainable parameters. With approximately half a million parameters, EquiNet-CNN demonstrates low sample complexity, as detailed in this section. We trained EquiNet-CNN on partial datasets consisting of 1,000, 2,000, 4,000, 8,000, 16,000, and 21,000 data points for a fixed number of training steps at 131,250, which is equivalent to training for 100 epochs on 21,000 data points using a batch size of 16.

\begin{figure}[h!]
    \centering
    \includegraphics[width=0.9\textwidth]{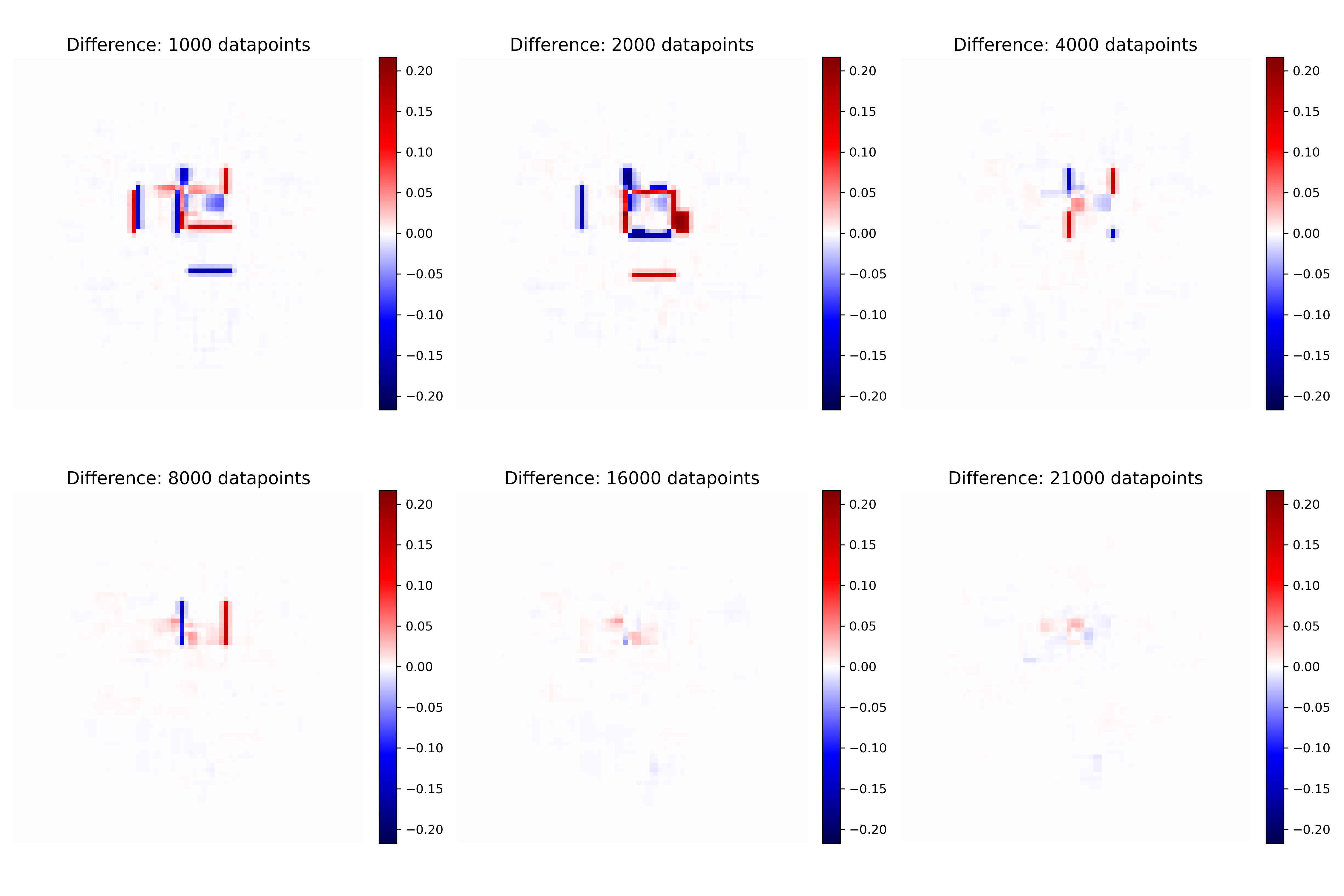}
    \caption{{Error in} reconstructions of 10h Overlapping Squares from EquiNet-CNN models trained on different dataset sizes. Notice that for models trained with fewer data points, the position of the square scatterer in the middle is inaccurate. }
    \label{fig:sample_complexity}
\end{figure}

Table~\ref{tab:sample_complexity} shows the comparison of the reconstructions from the models using metrics such as RRMSE, MELR, SD, and CRPS. With only 2,000 data points for training, EquiNet-CNN already outperforms all baseline models trained on 21,000 data points. The accuracy stagnates when training with more than 8,000 data points. Figure~\ref{fig:sample_complexity} displays a typical {error in} reconstruction from each model on the 10h Overlapping Squares dataset. Note that for models trained with a small number of data points, the positions {of some of the scatterers} are not accurate.

\begin{table}[h!]
\centering
\begin{tabular}{c|c|c|c|c}
\toprule
Dataset Size & RRMSE & MELR & SD & CRPS \\
 & & $(\times 10^{-2})$ & & $(\times 10^{-4})$\\
\midrule
\multicolumn{4}{c}{Shepp-Logan (Reference SD: 14.406)} \\
\midrule
1,000  & 2.085\% & 2.005 & 3.845 & 6.555 \\
2,000  & 1.644\% & 1.752 & 3.783 & 5.698 \\
4,000  & 1.379\% & 1.534 & 3.742 & 4.793 \\
8,000  & 1.376\% & 1.571 & 3.742 & 5.257 \\
16,000 & 1.281\% & 1.531 & 3.727 & 4.623 \\
21,000 & 1.323\% & 1.564 & 3.734 & 4.964 \\

\midrule
\multicolumn{4}{c}{3-5-10h Triangles (Reference SD: 2.833)} \\
\midrule
1,000  & 8.475\%  & 5.034 & 1.129 & 2.606 \\
2,000  & 5.454\%  & 3.833 & 1.049 & 1.783 \\
4,000  & 3.525\%  & 2.845 & 1.002 & 1.062 \\
8,000  & 2.824\%  & 3.047 & 0.981 & 1.204 \\
16,000 & 1.957\%  & 2.303 & 0.961 & 1.024 \\
21,000 & 1.590\%  & 1.385 & 0.949 & 0.815 \\

\midrule
\multicolumn{4}{c}{10h Overlapping Squares (Reference SD: 11.183)} \\
\midrule
1,000  & 12.454\% & 12.041 & 5.080 &  28.953 \\
2,000  & 7.893\%  & 8.392  & 4.568 &  16.678 \\
4,000  & 4.896\%  & 5.464  & 4.224 &  8.469 \\
8,000  & 2.756\%  & 3.054  & 3.982 &  5.268 \\
16,000 & 2.184\%  & 2.375  & 3.915 &  4.404 \\
21,000 & 1.744\%  & 1.979  & 3.860 &  3.916 \\
\bottomrule
\end{tabular}
\caption{{RRMSE, MELR, SD, and CRPS of reconstructions from EquiNet-CNN models trained on partial datasets of various sizes, using a fixed number of 131,250 training steps.}}
\label{tab:sample_complexity}
\end{table}

\subsection{Posterior Distribution}\label{sec:posterior_distribution}

Due to the probabilistic nature of the diffusion model, we test how well the EquiNet-CNN model captures the posterior distribution. 
Given that we do not have a ground truth for the posterior, we analyze the behavior of the reconstruction as we change the data used for training/inference.
As such, we artificially increase uncertainty by training EquiNet-CNN with monochromatic data at frequencies: 2.5, 5, or 10 for the {3-5-10h Triangles} dataset, which due to the multiple back-scattering should be the most sensitive to partial data. Then we pick one data point {from} one evaluation set, and we generate 500 conditional samples following $\bm{\eta} \sim p(\bm{\eta},  {\sf\Lambda})$. Then for each of these samples we compute the data misfit as $\| \mathcal{F}^{\omega}(\eta) - \Lambda^{\omega} \|$ for each of the frequencies.

\begin{figure}[h!]
    \centering
    \includegraphics[width=0.9\textwidth]{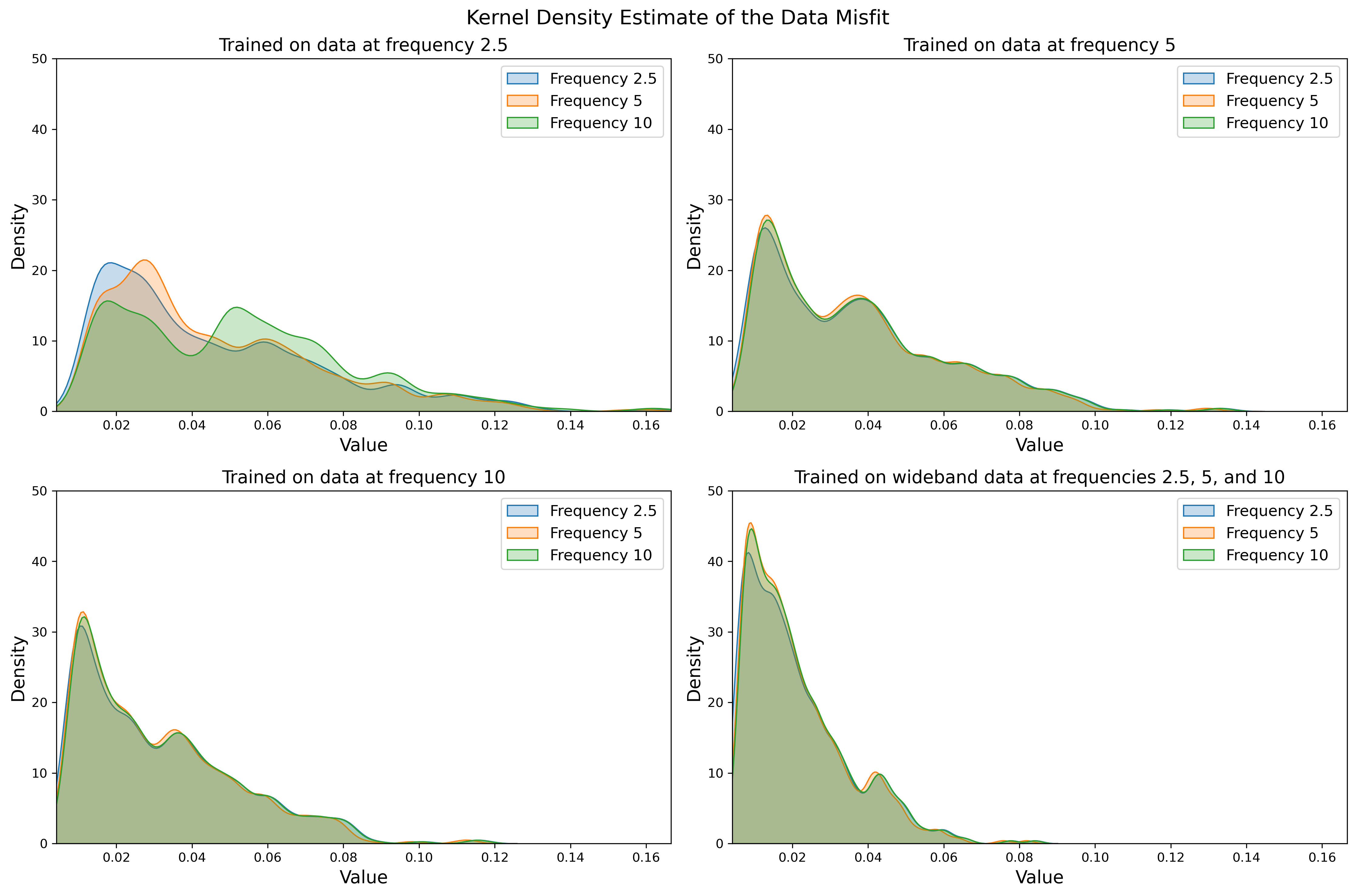}
    \caption{Estimated distributions of the data misfit for EquiNet-CNN {using the 3-5-10h Triangles dataset}, trained on data at a single frequency of 2.5, 5, and 10, as well as at wideband frequencies including 2.5, 5, and 10.}
    \label{fig:data_misfit_3510Tri}
\end{figure}

Table~\ref{tab:data_RRMSE_3510Tri} shows the statistics of the data misfit at frequencies of 2.5, 5, and 10, and Figure~\ref{fig:data_misfit_3510Tri} shows the estimated probability distributions of the data misfit for EquiNet-CNN trained on data at {a single} frequency of 2.5, 5, and 10, as well as at wideband frequencies including 2.5, 5, and 10. We {can} observe from the distributions that there are several modes, which correspond to cases where some squares in the reconstruction are shifted by one or more pixels from the ground truth. As reference, Table~\ref{tab:perturbation_errors} records the data misfit induced by manually moving a square in the ground truth by $n$ pixels; the errors match the locations of the modes in the distribution. As expected, using single low-frequency data produces the biggest uncertainty, and the wide-band data produces the least.

{We further extend the experiments to the Shepp-Logan and 10h Overlapping Squares datasets. The results are included in Appendix~\ref{sec:extension_posterior_distribution}: statistics of data misfits are recorded in Tables~\ref{tab:data_RRMSE_Shepp_Logan} and \ref{tab:data_RRMSE_Squares}; and  distributions are illustrated in Figures~\ref{fig:data_misfit_shepp_logan} and \ref{fig:data_misfit_Squares}.}

\begin{table}[h!]
\centering
\begin{tabular}{cccccc}
\toprule
\multicolumn{6}{c}{Trained on Data at Frequency 2.5} \\
\midrule
Frequency & Mean (\%) & Median (\%) & Min (\%) & Max (\%)  & Std  (\%)\\
\midrule
2.5& 4.546 & 3.666 & 1.037 & 16.659 & 2.918 \\
5  & 4.606 & 3.834 & 1.102 & 16.158 & 2.778 \\
10 & 5.232 & 5.031 & 1.134 & 16.561 & 2.964 \\
\end{tabular}

\begin{tabular}{cccccc}
\toprule
\multicolumn{6}{c}{Trained on Data at Frequency 5} \\
\midrule
Frequency & Mean (\%) & Median (\%) & Min (\%) & Max (\%) & Std (\%) \\
\midrule
2.5&  3.708 & 3.278 & 0.771 & 13.430 & 2.429\\
5  &  3.637 & 3.206 & 0.853 & 13.039 & 2.331\\
10 &  3.729 & 3.278 & 0.886 & 13.372 & 2.390\\
\end{tabular}

\begin{tabular}{cccccc}
\toprule
\multicolumn{6}{c}{Trained on Data at Frequency 10} \\
\midrule
Frequency & Mean (\%) & Median (\%) & Min (\%) & Max (\%) & Std (\%) \\
\midrule
2.5 & 3.126 & 2.631 & 0.498 & 11.697 & 2.080\\
5   & 3.067 & 2.577 & 0.609 & 11.361 & 1.999\\
10  & 3.149 & 2.646 & 0.624 & 11.659 & 2.050\\
\end{tabular}

\begin{tabular}{cccccc}
\toprule
\multicolumn{6}{c}{Trained on Data at Wideband Frequencies including 2.5, 5, and 10} \\
\midrule
Frequency & Mean (\%) & Median (\%) & Min (\%) & Max (\%) & Std (\%) \\
\midrule
2.5 & 2.109 & 1.749 & 0.416 & 8.428 & 1.370\\
5   & 2.094 & 1.749 & 0.542 & 8.190 & 1.308\\
10  & 2.152 & 1.799 & 0.548 & 8.406 & 1.341\\
\bottomrule
\end{tabular}
\caption{Statistics of data misfit at different frequencies for 500 samples generated using EquiNet-CNN on 1 data point from the 3-5-10h Triangles dataset.}
\label{tab:data_RRMSE_3510Tri}
\end{table}

\begin{table}[h!]
    \centering
    \begin{tabular}{cccc}
        \hline  Pixels$\backslash$Frequency &  2.5 &  5 &  10 \\
        \hline
        1 & 2.164 & 5.398 & 10.160 \\
        2 & 4.279 & 10.334 & 17.584 \\
        3 & 6.299 & 14.431 & 21.071 \\
        4 & 8.183 & 17.429 & 21.587\\
        5 & 9.895 & 19.237 & 21.431\\
        \hline
    \end{tabular}
    \caption{Relative magnitude of data misfit induced by manually shifting a triangle in a ground truth datapoint from the 3-5-10h Triangles dataset by $n$ pixels.}\label{tab:perturbation_errors}
\end{table}

\subsection{Cycle Skipping}\label{sec:cycle_skipping}
When training with only high-frequency data, classical methods like FWI often encounter a phenomenon known as cycle skipping, where the algorithm converges to a local minimum. We demonstrate that the EquiNet-CNN model significantly mitigates cycle skipping. As such, we trained the model on the Shepp-Logan, 3-5-10h Triangles, and 10h Overlapping Squares datasets using data only at a high frequency of 10. Table~\ref{tab:datasets_vs_errors_highfreq} shows the metrics, RRMSE, MELR, SD, and CRPS of the reconstructions. From table~\ref{tab:datasets_vs_errors_highfreq} we can observe that for the Shepp-Logan and 3-5-10h Triangles, training with data only at a frequency of 10, the model yields results comparable to those obtained with wideband data. As expected from the previous section, the reconstruction of 10h Overlapping {Squares} deteriorates when using only high-frequency data, due to the stronger {back-scattering; however, the error} remains relatively small. 

\begin{table}[h!]
\centering
\begin{tabular}{lcccccc}
\toprule
 & \multicolumn{2}{c}{Shepp-Logan} & \multicolumn{2}{c}{3-5-10h Triangles} & \multicolumn{2}{c}{10h Overlapping Squares} \\
\cmidrule(r){2-3} \cmidrule(r){4-5} \cmidrule(r){6-7}
Metric $\backslash$ Frequency & 2.5-5-10 & 10 &  2.5-5-10 & 10 & 2.5-5-10 & 10 \\
\midrule
RRMSE   & 1.414\% & 1.738\%  &  1.590\% & 1.955\% &  1.744\% & 4.993\% \\
MELR ($\times10^{-2}$)  & 1.847 & 2.045 & 1.385 & 1.969 & 1.979 & 4.603 \\
SD                      & 3.745 & 3.792 & 0.949 & 0.955  & 3.860 & 4.242 \\  
CRPS ($\times10^{-4}$)  &  5.550 & 8.287 &  0.812 & 1.135 & 3.916 & 13.335 \\
\bottomrule
\end{tabular}
\caption{RRMSE, MELR, SD, and CRPS of reconstructions from EquiNet-CNN trained using wideband data at frequencies of 2.5, 5, and 10, or using data at a frequency of 10.}
\label{tab:datasets_vs_errors_highfreq}
\end{table}

\begin{figure}[h!]
    \centering
    \includegraphics[width=0.9\textwidth]{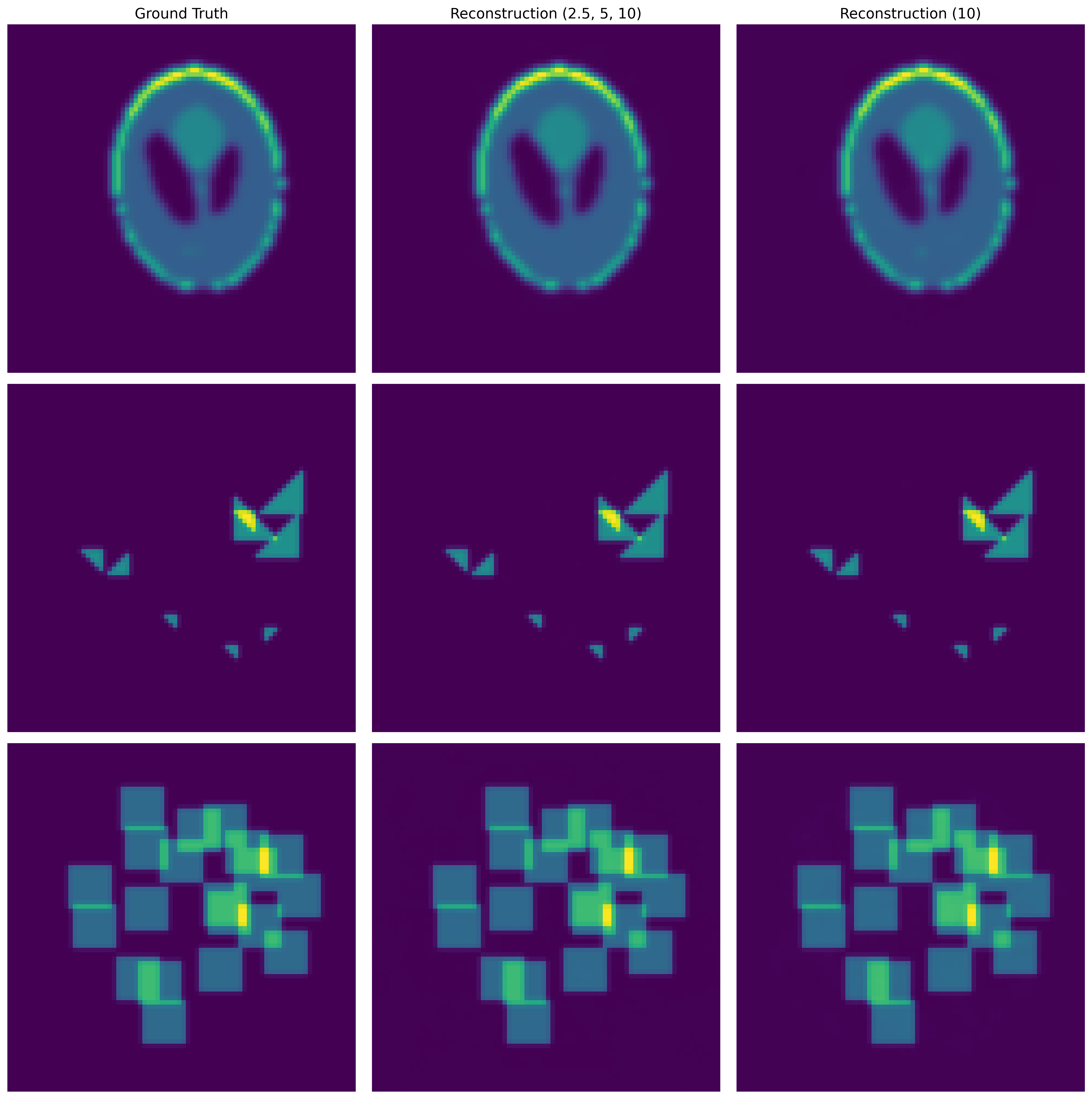}
    \caption{Comparison of reconstructions from the EquiNet-CNN trained with either wideband data at frequencies of 2.5, 5, and 10, or data at the highest frequency of 10. }
    \label{fig:cycle_skipping}
\end{figure}

Figure~\ref{fig:cycle_skipping} plots the reconstructions from EquiNet-CNN trained with either wideband data or data at the highest frequency.  In both cases, the errors are not noticeable to the naked eye.

\subsection{Ablation Studies}\label{sec:ablation_study}
Our main method relies on computing a latent representation of the intermediate field before using it to perform conditional sampling, where this latent representation is a neural network whose architecture is inspired by the back-scattering operators, see~\eqref{eqn:F_ast}, which in return mimics the physics of wave propagation.

We show that this step is crucial for the behavior of our algorithm, thus we consider a few variants of our main model EquiNet-CNN and B-EquiNet-CNN, which {modify} different parts of the algorithmic pipeline.

\paragraph{{Diffusion Models}} we use {a} regular state-of-the-art conditional diffusion model~\cite{karras2022elucidating}, to which we feed the discretized far-field patterns data directly. The neural architecture is a U-ViT network, see Remark~\ref{rem:UViT}. We refer to the model as None-U-ViT. {Additionally, we consider a CNN-based representation (see Appendix~\ref{sec:CNN}) to approximate the score function without a latent representation, which we refer to as None-CNN.}

\paragraph{Back-Projection CNN Diffusion Models} In this class, we find four models: B-EquiNet-CNN, SwitchNet-CNN, WideBNet-CNN, and Analytical-CNN. Similar to our main model EquiNet-CNN, the first three models preprocess the condition, i.e. the far-field patterns data, by using latent intermediate field representations of the corresponding deterministic models, whereas the Analytical-CNN {uses} the analytical expression of the back-scattering operator $F^\ast$ in~\eqref{eqn:F_ast} to compute the latent intermediate field representation. Then, all of them use the CNN-based representation, introduced in Appendix~\ref{sec:CNN}, to represent the physics-aware score function.

\paragraph{Back-Projection U-ViT Diffusion Model} we use the latent representation stemming from EquiNet to represent the intermediate field, but {we} use a U-ViT to represent the physics-aware score function, instead of the CNN-based networks used by the other models.

For each model we run a similar benchmark as the one in Section \ref{sec:comparison_vs_deterministic}, whose results are {summarized} in Table~\ref{tab:ablation_study}. {Across the table, EquiNet-CNN, B-EquiNet-CNN, and EquiNet-U-ViT outperform other variants.} We point out that using the analytical expression is competitive for datasets that are simple or that don't have complicated multiple back-scattering. EquiNet-U-ViT has comparable performance with EquiNet-CNN and B-EquiNet-CNN while having about 10 times more trainable parameters. {Without the latent representation, None-CNN performs the worst among all the variants.}

We compare the number of trainable parameters recorded in Table~\ref{tab:trainable_parameters}, by which we observe the main advantage of using symmetries in the construction of the network, as the number of {parameters} is around an order of {magnitude} lower for EquiNet-CNN and B-EquiNet-CNN while {still achieving competitive performance}.
Also, EquiNet-CNN and B-EquiNet-CNN outperform a pure U-ViT network, which underscores the advantages of the factorization introduced in this paper. 

\begin{table}[h!]
\centering
\begin{tabular}{c|c|c|c|c}
\toprule
Model & RRMSE & MELR & SD & CRPS \\
 & & $(\times 10^{-2})$ & & $(\times 10^{-4})$\\
\midrule
\multicolumn{5}{c}{Shepp-Logan (Reference SD: 14.406)} \\
\midrule
EquiNet-CNN    & 1.323\% & 1.564 & 3.734 & 4.964 \\
B-EquiNet-CNN  & 1.406\% & 1.757 & 3.754 & 5.318 \\
WideBNet-CNN   & 5.271\% & 4.881 & 4.294 & 17.982 \\
SwitchNet-CNN  & 1.943\% & 1.925 & 3.828 & 6.232 \\
Analytical-CNN & 2.371\% & 2.488 & 3.885 & 10.097 \\
EquiNet-U-ViT  & \textbf{0.928\%} & \textbf{1.132} & \textbf{3.682} & \textbf{3.406} \\
\midrule
None-CNN       &  12.870\% & 7.652  & 5.332  & 48.228 \\
None-U-ViT     &  2.317\% & 2.238  &  3.876  & 8.066  \\
\midrule
\multicolumn{5}{c}{3-5-10h Triangles (Reference SD: 2.833)} \\
\midrule
EquiNet-CNN    & \textbf{1.590\%} & 1.385 & \textbf{0.949} & 0.815 \\
B-EquiNet-CNN  & 1.657\% & 1.318 & 0.952 & 0.821 \\
WideBNet-CNN   & 17.197\% & 7.304 & 1.314 & 4.361 \\
SwitchNet-CNN  & 8.033\% & 3.881 & 1.112 & 1.858 \\
Analytical-CNN & 6.434\% & 3.462 & 1.060 & 2.065 \\
EquiNet-U-ViT  & 1.624\% & \textbf{1.191} & 0.957  & \textbf{0.393} \\
\midrule
None-CNN       &  133.757\% & 35.756 & 2.850  &  57.462 \\
None-U-ViT     &  10.001\% &  4.712 &  1.152  &  2.137 \\
\midrule
\multicolumn{5}{c}{10h Overlapping Squares (Reference SD: 11.183)} \\
\midrule
EquiNet-CNN    & 1.744\% & 1.979 & 3.860 & \textbf{3.916} \\
B-EquiNet-CNN  &  2.046\% & 2.683 & 3.894 & 5.298 \\
WideBNet-CNN   & 18.010\% & 17.595 & 5.715 & 48.399 \\
SwitchNet-CNN  &  10.644\% & 10.526 & 4.899 & 21.345 \\
Analytical-CNN & 11.946\% & 11.660  & 5.037 & 24.422 \\
EquiNet-U-ViT  & \textbf{1.458\%} &  \textbf{1.596}  & \textbf{3.831}  &  3.988 \\
\midrule
None-CNN       &  97.189\% & 29.341 &  11.637 & 462.454  \\
None-U-ViT     &  6.188\% & 6.485  &  4.378  & 11.002  \\
\bottomrule
\end{tabular}
\caption{Comparison of model performance on three synthetic datasets (Shepp-Logan, 3-5-10h
Triangles, and 10h Overlapping Squares) using the metrics RRMSE, MELR, SD, and CRPS.  The best performance metrics for each dataset are indicated in bold. }
\label{tab:ablation_study}
\end{table}

\begin{table}[h!]
\centering
\begin{tabular}{lcc}
\toprule
Model & Number of Trainable Parameters \\
\midrule
EquiNet-CNN       & 530,095\\
B-EquiNet-CNN     & 434,479\\
WideBNet-CNN      & 2,287,025 \\
SwitchNet-CNN     & 9,679,273\\
Analytical-CNN    & 378,793 \\
EquiNet-U-ViT     & 5,753,351\\
\midrule
None-CNN          & 418,221\\
None-U-ViT        & 5,746,445\\
\bottomrule
\end{tabular}
\caption{Number of trainable parameters for the {different} models.}
\label{tab:trainable_parameters}
\end{table}

\subsection{Inverse Crime and Noisy Inputs}\label{sec:noise_inverse_crime}

In general, the input data will have a certain amount of uncertainty, either epistemic, due to incomplete knowledge of the physical system, or stochastic, such as slight movement {of} the receivers. 

Thus, we showcase the resilience of our method to small changes in the data distributions stemming from uncertainties {in} the physical models or stochastic noise in the inputs.

\begin{figure}[h!]
    \centering
    \includegraphics[width=\textwidth]{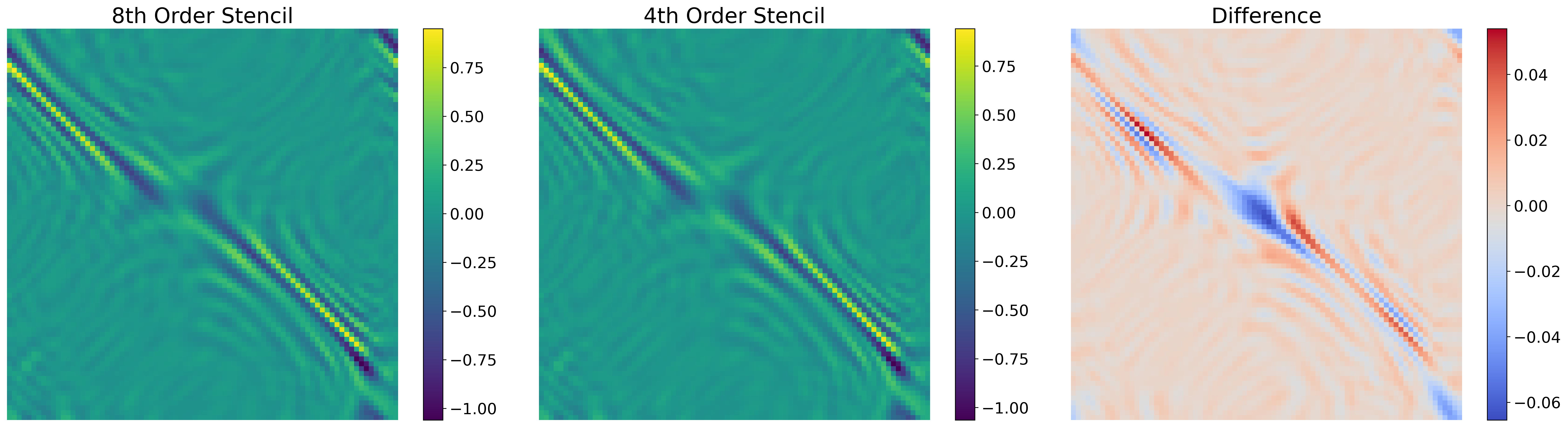}
    \caption{Comparison of the scattering data at a frequency of 10 using 8th and 4th order stencils, and the difference between them.}
    \label{fig:Lambda_comparison}
\end{figure}

Using the same numerical method for generating the training data, and performing the {reconstruction} is usually called the \textit{inverse crime} in the inverse problem community. In statistical terms, this issue is related to overfitting the model to a particular distribution. To {assess} if our framework suffers from this issue, we fed our trained models input data produced by a numerical solver using a different finite differences discretization, i.e., using a different stencil. Figure~\ref{fig:Lambda_comparison} depicts the scattering data $\Lambda$ at frequency 10, and {shows} the difference between $\Lambda$ generated using an 8th order stencil and {a} 4th order stencil. In this case, we can observe that there are differences in the amplitude of the largest reflections by around 5\%, although the plot does not show any large phase errors.
We point out that, due to dispersion errors, the differences become more apparent with fields at higher frequency\footnote{This is a well studied issue with Helmholtz solvers}. 

Table~\ref{tab:model_stability} records the RRMSE of the reconstruction when using within-distribution and out-of-distribution input data, i.e., data generated with {the same} stencil (order = 8) as the training data, and data using a different stencil (order 4 or 6). We can observe that across the different {distributions} of scatterers used in this work, the difference is minimal in the error. 
In summary, Table~\ref{tab:model_stability} demonstrates that the model EquiNet-CNN is stable under moderate perturbation of the input distribution induced by using a different numerical solver.
\begin{table}[h!]
    \centering
    \begin{tabular}{c|c|c|c}
        \hline
        Dataset/Stencil & 4th Order Stencil & 6th Order Stencil & 8th Order Stencil \\
        \hline
        Shepp-Logan           &  1.581\% & 1.402\%  &  1.414\% \\
        3-5-10h Triangles   & 1.537\%  & 1.550\%  &  1.590\% \\
        10h Overlapping Squares  &  1.788\% &  1.773\% &  1.744\% \\
        \hline
    \end{tabular}
    \caption{RRMSE comparisons for EquiNet-CNN across different datasets and stencil orders, demonstrating the model's stability with within-distribution and out-of-distribution inputs.}
    \label{tab:model_stability}
\end{table}

\begin{figure}[h!]
    \centering
    \includegraphics[width=\textwidth]{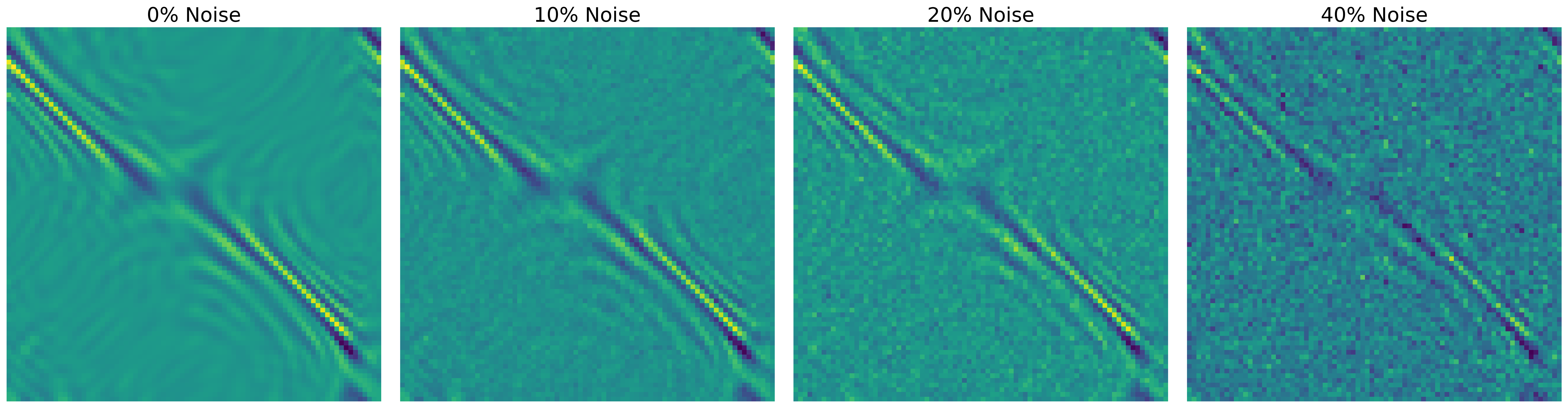}
    \caption{Comparison of data at a frequency of 10, noise-free, and with different levels of noise (10\%, 20\%, 40\%). We can observe that beyond 20\% noise, most of the waves stemming from multiple scattering reflections are drowned by the noise.}
    \label{fig:data_noise_comparison}
\end{figure}

For the case of stochastic uncertainty, we feed our model with scattering data artificially corrupted by different noise levels. The data is corrupted using additive Gaussian noise following,
\begin{equation}
    {\sf\Lambda}_{\text{noised}}  =  {\sf\Lambda} + \gamma_{\text{noise}} \sigma_{{\sf\Lambda}} \bm{\varepsilon}
\end{equation}
where, $\bm{\varepsilon}$ follows a unit normal distribution, $\sigma_{{\sf\Lambda}}$ is the standard deviation of the scattering data, and $\gamma_{\text{noise}} $ is the noise level.
Examples of these noised scattering data, ${\sf\Lambda}_{\text{noised}}$, are showcased in Figure~\ref{fig:data_noise_comparison}. From the figure we can observe that the largest reflections and transmission waves are still legible, although for noise levels beyond 20\%, i.e., $\gamma_{\text{noise}} = 0.2$, most of the multiple back-scattered waves are drowned by the noise. 

\begin{figure}[h!]
    \centering
    \includegraphics[width=0.9\textwidth]{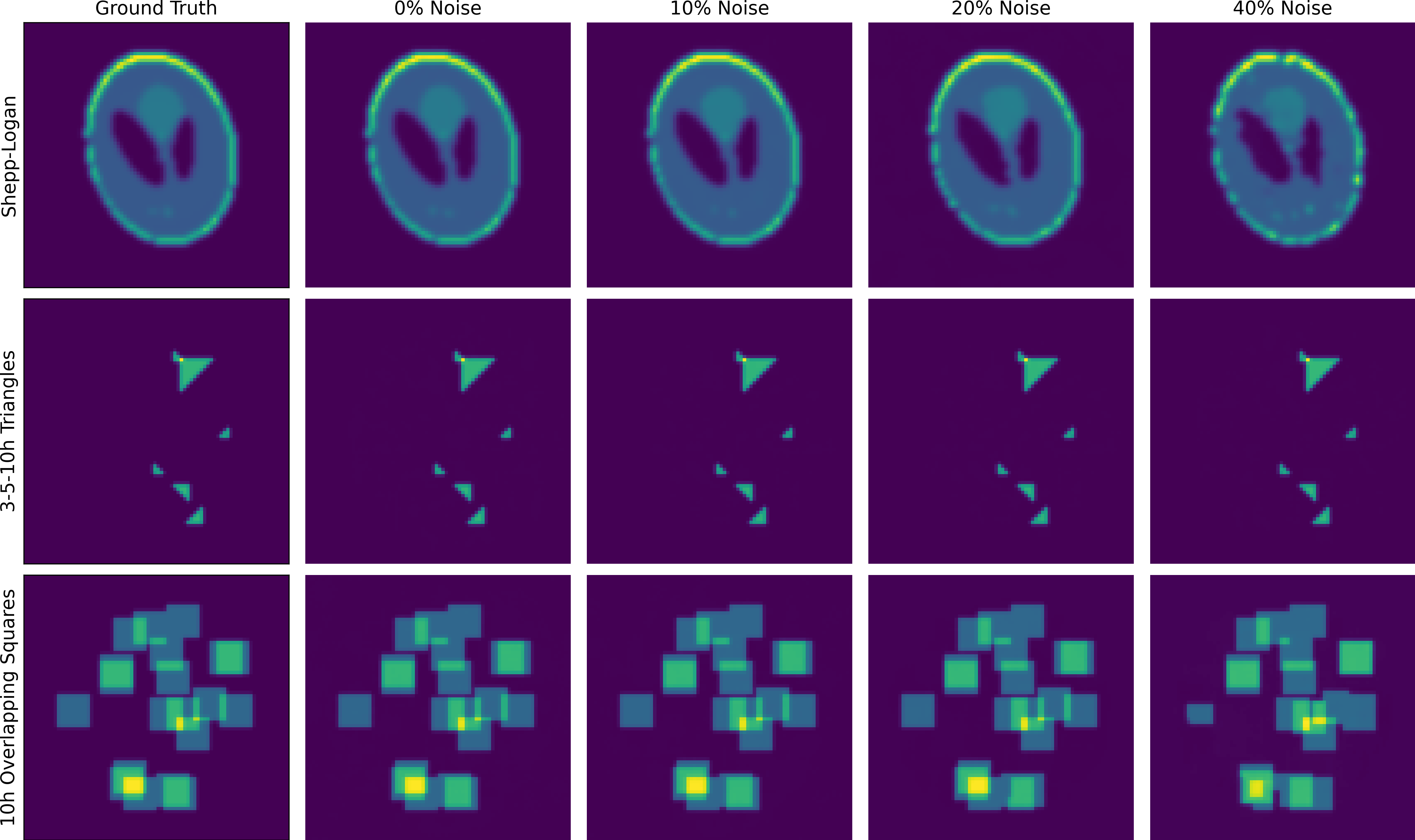}
    \caption{Comparison of reconstructions from EquiNet-CNN on the three synthetic datasets as noise increases from 0\% to 40\%.}
    \label{fig:noise_comparison}
\end{figure}

We perform a similar benchmark to the one presented in Section \ref{sec:comparison_vs_deterministic} but {we} add noise at 4 different {levels} (10\%, 20\%, and 40\%) to the testing data for each of the three synthetic datasets, and we compare the RRMSE, MELR, SD, and CRPS of the resulting reconstructions. The results are summarized in 
Table~\ref{tab:noise_vs_rrmse}, which shows how the reconstruction using EquiNet-CNN deteriorates as the noise increases. To provide visual cues on how the noise corrupts the {reconstruction} we provide some samples of the reconstruction in Figure~\ref{fig:noise_comparison}, which shows typical reconstruction at different levels of noise from the three synthetic datasets.
\begin{table}[h!]
\centering
\begin{tabular}{c|c|c|c|c}
\toprule
Noise Level & RRMSE & MELR & SD & CRPS\\
 & & $(\times 10^{-2})$ & & $(\times 10^{-4})$\\
\midrule
\multicolumn{5}{c}{Shepp-Logan (Reference SD: 14.406)} \\
\midrule
0\% Noise  & 1.323\% & 1.564 & 3.734  & 4.964\\
10\% Noise & 3.355\% & 2.483 & 4.017  & 9.595\\
20\% Noise & 7.422\% & 4.344 & 4.576  & 30.890\\
40\% Noise & 14.827\% & 8.536 & 5.575 & 77.136\\
\midrule
\multicolumn{5}{c}{3-5-10h Triangles (Reference SD: 2.833)} \\
\midrule
0\% Noise  &1.590\% & 1.385 & 0.949 & 0.812\\
10\% Noise &1.564\% & 1.161 & 0.949 & 0.768\\
20\% Noise &1.618\% & 1.203 & 0.950 & 0.806\\
40\% Noise &3.522\% & 2.053 & 0.977 & 1.474\\
\midrule
\multicolumn{5}{c}{10h Overlapping Squares (Reference SD: 11.183)} \\
\midrule
0\% Noise  & 1.744\% & 1.979 & 3.860 &  3.916\\
10\% Noise & 2.675\% & 2.913 & 3.973 &  5.073\\
20\% Noise & 8.464\% & 8.426 & 4.627 &  15.114 \\
40\% Noise &24.761\% & 21.530 & 6.454&  104.699\\
\bottomrule
\end{tabular}
\caption{Model performance of EquiNet-CNN on three synthetic datasets as noise levels increase from 10\% to 40\%. We evaluate using the metrics RRMSE, MELR, SD, and CRPS.}
\label{tab:noise_vs_rrmse}
\end{table}
We can observe that even for relatively high levels of noise, the algorithm is still able to reconstruct the main features. 

\begin{figure}[h!]
    \centering
    \includegraphics[width=0.9\textwidth]{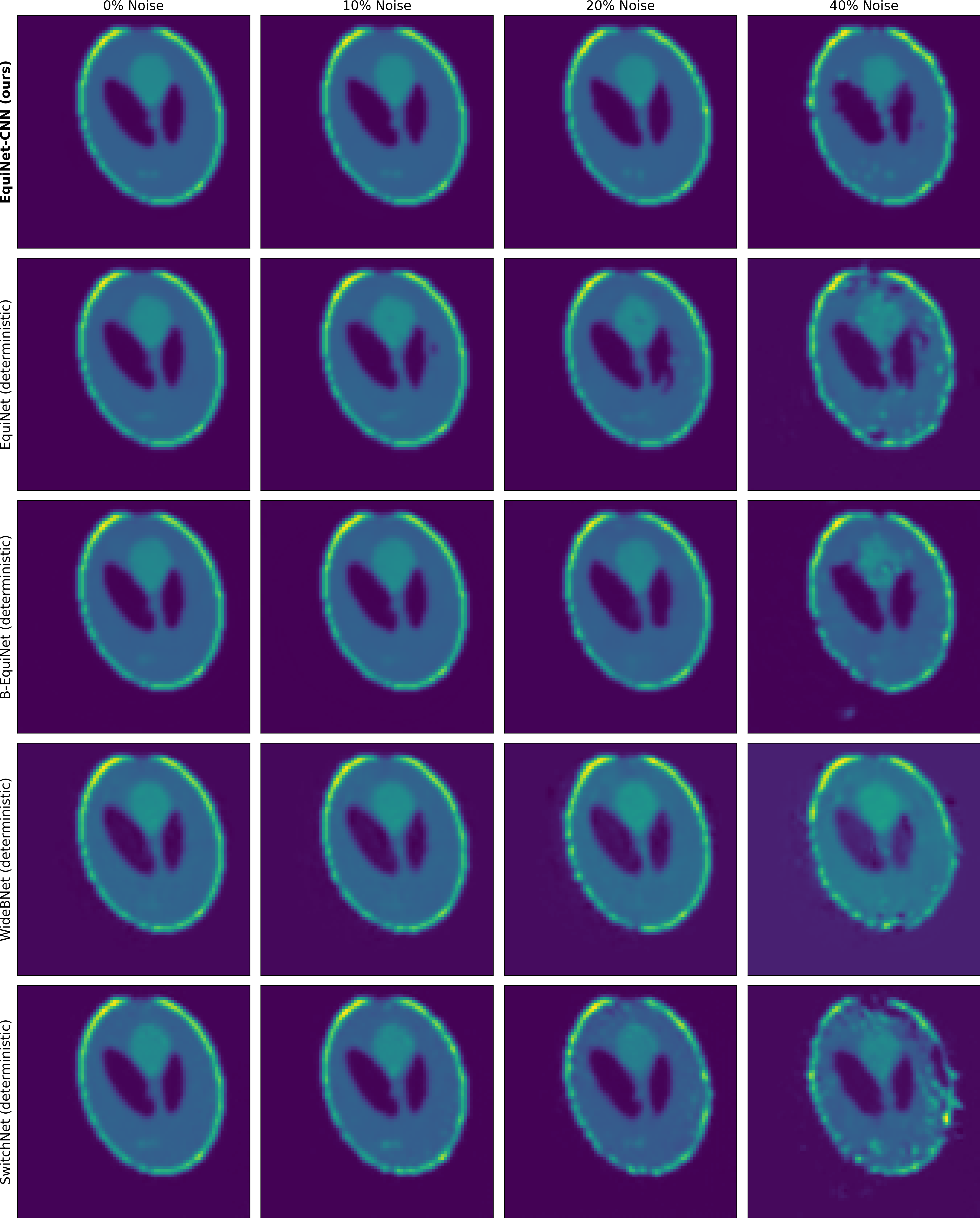}
    \caption{Comparison of reconstructions from EquiNet-CNN and baseline deterministic models on Shepp-Logan dataset as noise increases from 0\% to 40\%.}
    \label{fig:model_comparison_noise}
\end{figure}

To put these results in context, we perform the same procedure using some of the baselines, whose results are summarized in Table~\ref{tab:rrmse_comparison_datasets}, which records the RRMSE of reconstructions between EquiNet-CNN and the deterministic models: EquiNet, B-EquiNet, WideBNet, and SwitchNet. From this Table, we can observe that our methodology has overall the lowest reconstruction error for all levels of noise. In addition, Figure~\ref{fig:model_comparison_noise} shows a typical reconstruction using EquiNet-CNN and baseline deterministic models for the Shepp-Logan dataset at different levels of noise. In this case, we can observe that even though the error is large, we can still observe many of the main features of the Shepp-Logan phantom.

\begin{table}[h!]
\centering
\begin{tabular}{lcccc}
\toprule
\multicolumn{5}{c}{Shepp-Logan} \\
\midrule
Model & 0\% Noise &  10\% Noise & 20\% Noise & 40\% Noise \\
\midrule
EquiNet-CNN & \textbf{1.323\%} & \textbf{3.355\%} & \textbf{7.422\%} & \textbf{14.827\%}\\
EquiNet (deterministic)    & 1.693\%   & 3.549\% & 7.512\% & 17.357\% \\
B-EquiNet (deterministic)  & 2.022\%   & 4.030\% & 7.801\% & 15.738\% \\
WideBNet (deterministic)   & 3.843\%   & 7.417\% & 13.227\%  &  26.019\% \\
SwitchNet (deterministic)  & 4.305\%   & 6.173\% & 10.022\% & 21.111\% \\
\midrule
\multicolumn{5}{c}{3-5-10h Triangles} \\
\midrule
Model & 0\% Noise &  10\% Noise & 20\% Noise & 40\% Noise \\
\midrule
EquiNet-CNN &\textbf{1.590\%}& \textbf{1.564\%} & \textbf{1.618\%} & \textbf{3.522\%} \\
EquiNet (deterministic)    & 2.741\%   & 4.143\% & 4.508\% & 5.623\% \\
B-EquiNet (deterministic)  & 2.944\%   & 4.028\% & 4.430\% & 7.596\% \\
WideBNet (deterministic)   & 17.263\%  & 19.632\% & 21.422\% & 30.021\% \\
SwitchNet (deterministic)  & 15.084\%  & 16.981\% & 18.859\% & 23.230\% \\
\midrule
\multicolumn{5}{c}{10h Overlapping Squares} \\
\midrule
Model & 0\% Noise &  10\% Noise & 20\% Noise & 40\% Noise \\
\midrule
EquiNet-CNN                & \textbf{1.744\%} & \textbf{2.675\%} & \textbf{8.464\%} &\textbf{25.379\%} \\
EquiNet (deterministic)    & 10.891\% & 12.046\% & 15.653\% & 27.732\% \\
B-EquiNet (deterministic)  & 9.484\%  & 12.536\% & 20.129\% & 36.117\% \\
WideBNet (deterministic)   & 14.327\% & 17.048\% & 23.730\% & 39.998\% \\
SwitchNet (deterministic)  & 20.102\% & 22.217\% &27.243\%  &40.178\% \\
\bottomrule
\end{tabular}
\caption{Comparison of RRMSE of reconstructions for Shepp-Logan, 3-5-10h Triangles, and 10h Overlapping Squares datasets between EquiNet-CNN and deterministic models (EquiNet, B-EquiNet, WideBNet, and SwitchNet) at different noise levels. The best performance metrics for each dataset are indicated in bold.}
\label{tab:rrmse_comparison_datasets}
\end{table}

\subsection{Mixed datasets and Generalization}\label{sec:mixed_data}
We also consider how well our methodology behaves with more complex distributions and also with out-of-distribution samples.

To assess the behavior of our networks with more complex distributions, we train our network, along with some of the baselines, using a mix of all synthetic datasets: Shepp-Logan, 3-5-10h Triangles, and 10h Overlapping Squares. We then compute the RRMSE, MELR, and SD of the respective reconstructions for each dataset separately, which are shown in Table~\ref{tab:rrmse_comparison_mixed_datasets}. From Table~\ref{tab:rrmse_comparison_mixed_datasets}, we can observe that EquiNet-CNN is able to generate samples for each dataset with relatively high accuracy when training with the mixed dataset, in contrast with other deterministic models. This observation indicates that EquiNet-CNN has, in general, higher capacity than the alternatives, and it is capable of learning more distributions of scatterers with higher intrinsic dimensionality when compared to other ML-based methods.

\begin{table}[h!]
\centering
\begin{tabular}{c|c|c|c}
\toprule
Model & RRMSE & MELR & SD \\
 & & $(\times 10^{-2})$ & \\
\midrule
\multicolumn{4}{c}{Shepp-Logan (Reference SD: 14.406)} \\
\midrule
EquiNet-CNN               &\textbf{1.583\%} & \textbf{1.899} & \textbf{3.774}\\
EquiNet (deterministic)   &4.360\% &16.033 & 4.155\\
B-EquiNet (deterministic) &3.581\% & 8.748 & 4.047\\
WideBNet (deterministic)  &6.971\% & 28.073 & 4.512 \\
SwitchNet (deterministic) &6.279\% & 21.042 & 4.407\\
\midrule
\multicolumn{4}{c}{3-5-10h Triangles (Reference SD: 2.833)} \\
\midrule
EquiNet-CNN               &\textbf{6.161\%} & \textbf{3.533} & \textbf{1.044}\\
EquiNet (deterministic)   &19.055\% & 30.432& 1.297\\
B-EquiNet (deterministic) &12.404\% & 13.696 & 1.177 \\
WideBNet (deterministic)  &36.077\% &117.196 & 1.591 \\
SwitchNet (deterministic) &32.460\% & 93.747 & 1.520 \\
\midrule
\multicolumn{4}{c}{10h Overlapping Squares (Reference SD: 11.183)} \\
\midrule
EquiNet-CNN               &\textbf{2.579\%} & \textbf{2.857} & \textbf{3.962}\\
EquiNet (deterministic)   &12.692\% & 35.055 & 5.078\\
B-EquiNet (deterministic) &10.687\% & 26.418& 4.860\\
WideBNet (deterministic)  &19.039\% & 104.101 &5.765 \\
SwitchNet (deterministic) &15.277\% & 42.314 & 5.365 \\
\bottomrule
\end{tabular}
\caption{Comparison of RRMSE, MELR, and SD of reconstructions for each dataset (Shepp-Logan, 3-5-10h Triangles, and 10h Overlapping Squares) between EquiNet-CNN and deterministic models (EquiNet, B-EquiNet, WideBNet, and SwitchNet). All of the models are trained with the mixed dataset. The best performance metrics for each dataset are indicated in bold.}
\label{tab:rrmse_comparison_mixed_datasets}
\end{table}

\begin{figure}[h!]
    \centering
    \includegraphics[width=0.9\textwidth]{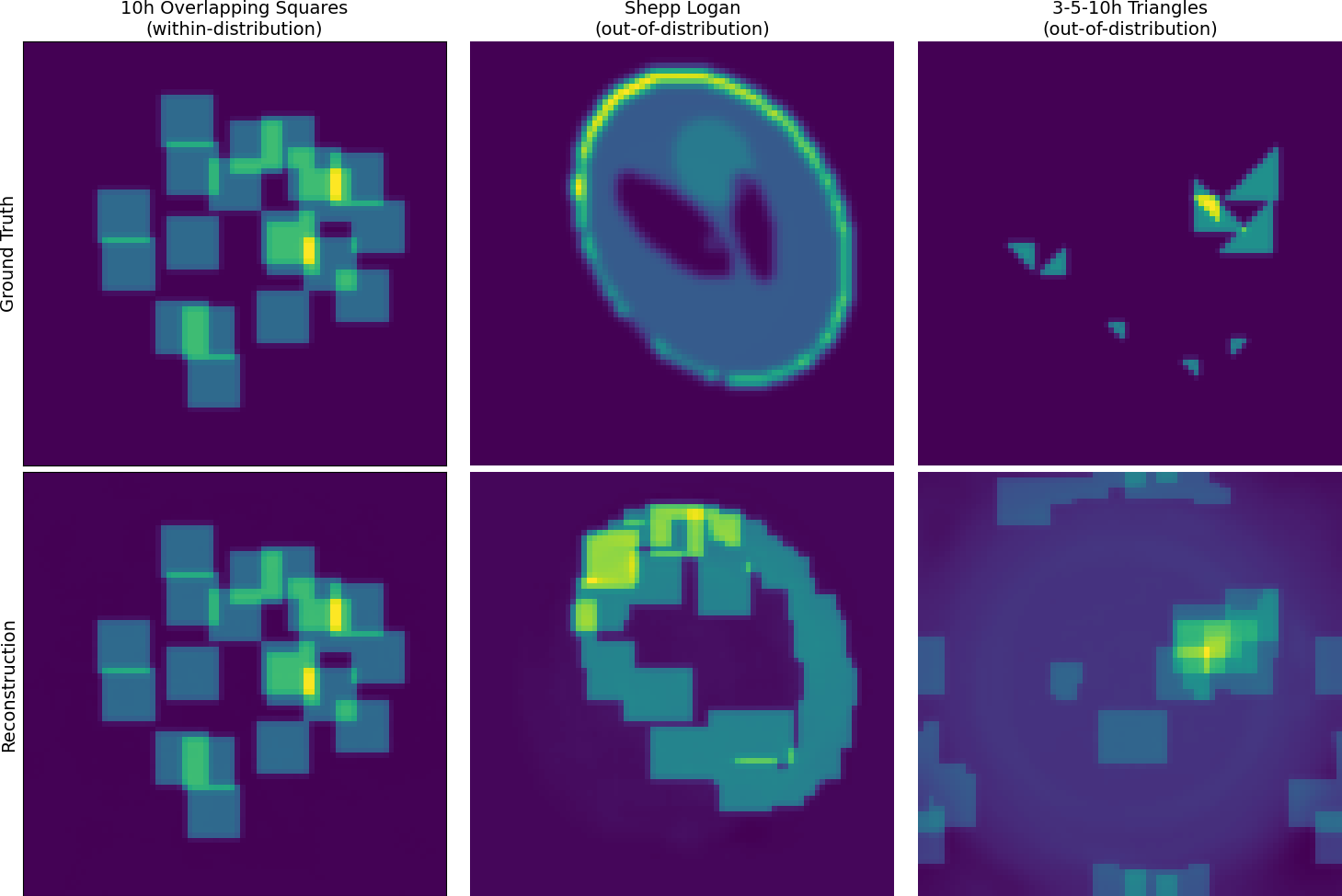}
    \caption{Reconstructions of out-of-distribution Shepp-Logan and 3-5-10h Triangles datasets from EquiNet-CNN trained on within-distribution 10h Overlapping Squares dataset.} 
    \label{fig:generalization}
\end{figure}

We also consider performance of EquiNet-CNN on datasets that are out-of-distribution, in contrast to the within-distribution training dataset.  For such assesment, we trained EquiNet-CNN on the 10h Overlapping Squares dataset and we tested the model on the out-of-distrbution Shepp-Logan and 3-5-10h Triangles dataset.  Figure~\ref{fig:generalization} shows reconstructions of the model from both within-distribution 3-5-10h Triangles dataset and out-of-distribution Shepp-Logan and 10h Overlapping Squares datasets.  
Although, our model, given its statistical nature, generalizes poorly, it is still able to locate and provide some features of the Shepp-Logan phantom.

\section{Conclusion}

By factorizing the score function according to the filtered back-projection formula into back-projection steps that compute a latent representation and a conditional scoring function conditioned on this latent representation, we provide a powerful yet straightforward framework that combines generative AI methods with analytical knowledge of the underlying physical problem. By designing networks that carefully exploit this knowledge, we produce state-of-the-art reconstructions with built-in uncertainty that outperform existing methods, even in the notoriously difficult multiple scattering cases.


{Despite outperforming other methods with increasingly complex scatterer distributions (Section \ref{sec:mixed_data}), our methodology remains constrained by the assumption that the scatterers follow a distribution with low intrinsic dimensionality, leading to generalization limitations. This indicates that our model may not fully capture the underlying physics of wave propagation beyond the underlying distribution. However, for applications such as radar, biomedical imaging, and materials science, where these assumptions are valid, our approach demonstrates superior performance compared to competing techniques.}

Future research directions include how to \emph{explicitly} reincorporate the physics back into the sampling process by directly leveraging the PDE, and how to automatically check if an input is out-of-distribution, thus re-weighting the prior during the sampling process.

\section{Acknowledgement}
The work of Q.~L. is supported by the National Science Foundation under the grant DMS-2308440. The work of M.~G. is partially supported by the National Science Foundation under the grant DMS-2012292. The views expressed in the article do not necessarily represent the views of any funding agencies. The authors are grateful for the support. 

\bibliography{neurips_2024}

\begin{thebibliography}{100}

\bibitem{Beylkin_Burridge:1990}
G.~Beylkin and R.~Burridge.
\newblock Linearized inverse scattering problems in acoustics and elasticity.
\newblock {\em Wave Motion}, 12(1):15 -- 52, 1990.

\bibitem{Tarantola:Inversion_of_seismic_reflection_data_in_the_acoustic_approximation}
A.~Tarantola.
\newblock Inversion of seismic reflection data in the acoustic approximation.
\newblock {\em GEOPHYSICS}, 49(8):1259--1266, 1984.

\bibitem{Virieux_Operto:An_overview_of_full-waveform_inversion_in_exploration_geophysics}
J.~Virieux and S.~Operto.
\newblock An overview of full-waveform inversion in exploration geophysics.
\newblock {\em GEOPHYSICS}, 74:WCC1--WCC26, 2009.

\bibitem{Schomberg:1978}
H.~Schomberg.
\newblock An improved approach to reconstructive ultrasound tomography.
\newblock {\em J. of Phys. D: Appl. Phys.}, 11(15):L181--L185, oct 1978.

\bibitem{Cheney_SAR:2001}
M.~Cheney.
\newblock A mathematical tutorial on synthetic aperture radar.
\newblock {\em SIAM Rev.}, 43(2):301--312, 2001.

\bibitem{Pettit:2015}
J.~R. Pettit, A.~E. Walker, and M.~J.~S. Lowe.
\newblock Improved detection of rough defects for ultrasonic nondestructive evaluation inspections based on finite element modeling of elastic wave scattering.
\newblock {\em IEEE T. Ultrason. Ferr.}, 62:1797--1808, 2015.

\bibitem{Rawlinson:2010}
N.~Rawlinson, S.~Pozgay, and S.~Fishwick.
\newblock Seismic tomography: A window into deep {E}arth.
\newblock {\em Phys. Earth Planet. Int.}, 178(3-4):101--135, 2 2010.

\bibitem{abbe1873beitrage}
Ernst Abbe.
\newblock Beitr{\"a}ge zur theorie des mikroskops und der mikroskopischen wahrnehmung.
\newblock {\em Archiv f{\"u}r mikroskopische Anatomie}, 9(1):413--468, 1873.

\bibitem{garnier2016passive}
J.~Garnier and G.~Papanicolaou.
\newblock {\em Passive Imaging with Ambient Noise}.
\newblock Cambridge Monographs on Applied and Computational Mathematic. Cambridge University Press, 2016.

\bibitem{rayleigh1879xxxi}
Rayleigh.
\newblock Xxxi. investigations in optics, with special reference to the spectroscope.
\newblock {\em The London, Edinburgh, and Dublin Philosophical Magazine and Journal of Science}, 8(49):261--274, 1879.

\bibitem{Hahner_Hohage2001_inverse_problem_estimates}
Peter Hähner and Thorsten Hohage.
\newblock New stability estimates for the inverse acoustic inhomogeneous medium problem and applications.
\newblock {\em SIAM J. Math. A.}, 33(3):670--685, 2001.

\bibitem{Leeuwen_Herrmann:2013}
T.~van Leeuwen and F.~J. Herrmann.
\newblock {Mitigating local minima in full-waveform inversion by expanding the search space}.
\newblock {\em Geophys. J. Int.}, 195(1):661--667, 07 2013.

\bibitem{MLZ}
Matthew Li, Laurent Demanet, and Leonardo Zepeda-N\'{u}\~{n}ez.
\newblock Wide-band butterfly network: Stable and efficient inversion via multi-frequency neural networks.
\newblock {\em Multiscale Modeling \& Simulation}, 20(4):1191--1227, 2022.

\bibitem{ZepedaDemanet:the_method_of_polarized_traces}
L.~Zepeda-N\'u\~nez and L.~Demanet.
\newblock The method of polarized traces for the {2D} {H}elmholtz equation.
\newblock {\em J. Comput. Phys.}, 308:347 -- 388, 2016.

\bibitem{zepeda2019method}
Leonardo Zepeda-N{\'u}nez, Adrien Scheuer, Russell~J Hewett, and Laurent Demanet.
\newblock The method of polarized traces for the 3d helmholtz equation.
\newblock {\em Geophysics}, 84(4):T313--T333, 2019.

\bibitem{cakoni2005qualitative}
Fioralba Cakoni and David Colton.
\newblock {\em Qualitative methods in inverse scattering theory: An introduction}.
\newblock Springer Science \& Business Media, 2005.

\bibitem{chung2011adaptive}
Eric Chung, Jianliang Qian, Gunther Uhlmann, and Hongkai Zhao.
\newblock An adaptive phase space method with application to reflection traveltime tomography.
\newblock {\em Inverse problems}, 27(11):115002, 2011.

\bibitem{Pratt:Seismic_waveform_inversion_in_the_frequency_domain;_Part_1_Theory_and_verification_in_a_physical_scale_model}
R.~G. Pratt.
\newblock Seismic waveform inversion in the frequency domain; part 1: {T}heory and verification in a physical scale model.
\newblock {\em GEOPHYSICS}, 64(3):888--901, 1999.

\bibitem{chen1997inverse}
Yu~Chen.
\newblock Inverse scattering via heisenberg's uncertainty principle.
\newblock {\em Inverse problems}, 13(2):253, 1997.

\bibitem{appelo2020waveholtz}
Daniel Appelo, Fortino Garcia, and Olof Runborg.
\newblock Waveholtz: Iterative solution of the helmholtz equation via the wave equation.
\newblock {\em SIAM Journal on Scientific Computing}, 42(4):A1950--A1983, 2020.

\bibitem{taus2020sweeps}
Matthias Taus, Leonardo Zepeda-N{\'u}{\~n}ez, Russell~J Hewett, and Laurent Demanet.
\newblock L-sweeps: A scalable, parallel preconditioner for the high-frequency helmholtz equation.
\newblock {\em Journal of Computational Physics}, 420:109706, 2020.

\bibitem{zepeda2016fast}
Leonardo Zepeda-N{\'u}nez and Hongkai Zhao.
\newblock Fast alternating bidirectional preconditioner for the 2d high-frequency lippmann--schwinger equation.
\newblock {\em SIAM Journal on Scientific Computing}, 38(5):B866--B888, 2016.

\bibitem{gander2019class}
Martin~J Gander and Hui Zhang.
\newblock A class of iterative solvers for the helmholtz equation: Factorizations, sweeping preconditioners, source transfer, single layer potentials, polarized traces, and optimized schwarz methods.
\newblock {\em Siam Review}, 61(1):3--76, 2019.

\bibitem{melia2024multi}
Owen Melia, Olivia Tsang, Vasileios Charisopoulos, Yuehaw Khoo, Jeremy Hoskins, and Rebecca Willett.
\newblock Multi-frequency progressive refinement for learned inverse scattering.
\newblock {\em Journal of Computational Physics}, 527:113809, 2025.

\bibitem{Butterfly-Net2}
Zhongshu Xu, Yingzhou Li, and Xiuyuan Cheng.
\newblock {Butterfly-Net2: Simplified Butterfly-Net and F}ourier transform initialization.
\newblock In Jianfeng Lu and Rachel Ward, editors, {\em Proceedings of The First Mathematical and Scientific Machine Learning Conference}, volume 107 of {\em Proceedings of Machine Learning Research}, pages 431--450, Princeton University, Princeton, NJ, USA, 20--24 Jul. 2020. PMLR.

\bibitem{Khoo_YingSwitchNet:2019}
Y.~Khoo and L.~Ying.
\newblock Switch{N}et: A neural network model for forward and inverse scattering problems.
\newblock {\em SIAM J. Sci. Comput.}, 41(5):A3182--A3201, 2019.

\bibitem{YL}
Yuwei Fan and Lexing Ying.
\newblock Solving inverse wave scattering with deep learning.
\newblock {\em Annals of Mathematical Sciences and Applications}, 7(1):23--48, 2022.

\bibitem{Fprinciple_NeuroIPS}
Zhi-Qin~John Xu, Yaoyu Zhang, and Yanyang Xiao.
\newblock Training behavior of deep neural network in frequency domain.
\newblock In Tom Gedeon, Kok~Wai Wong, and Minho Lee, editors, {\em Neural Information Processing}, pages 264--274, Cham, 2019. Springer International Publishing.

\bibitem{equivariant}
Borong Zhang, Leonardo Zepeda-N{\'u}{\~n}ez, and Qin Li.
\newblock Solving the wide-band inverse scattering problem via equivariant neural networks.
\newblock {\em Journal of Computational and Applied Mathematics}, page 116050, 2024.

\bibitem{li2021accurate}
Matthew Li, Laurent Demanet, and Leonardo Zepeda-N{\'u}{\~n}ez.
\newblock Accurate and robust deep learning framework for solving wave-based inverse problems in the super-resolution regime.
\newblock {\em arXiv preprint arXiv:2106.01143}, 2021.

\bibitem{tarantola1982inverse}
Albert Tarantola, Bernard Valette, et~al.
\newblock Inverse problems= quest for information.
\newblock {\em Journal of geophysics}, 50(1):159--170, 1982.

\bibitem{Ho_DDPM2020}
Jonathan Ho, Ajay Jain, and Pieter Abbeel.
\newblock Denoising diffusion probabilistic models.
\newblock In {\em Proceedings of the 34th International Conference on Neural Information Processing Systems}, NIPS '20, Red Hook, NY, USA, 2020. Curran Associates Inc.

\bibitem{tarantola2005inverse}
Albert Tarantola.
\newblock {\em Inverse problem theory and methods for model parameter estimation}.
\newblock SIAM, 2005.

\bibitem{Stuart_2010:inverse_problems}
A.~M. Stuart.
\newblock Inverse problems: A bayesian perspective.
\newblock {\em Acta Numerica}, 19:451–559, 2010.

\bibitem{implicit_neural_vlasic_2022}
Tin Vlašić, Hieu Nguyen, Amirehsan Khorashadizadeh, and Ivan Dokmanic.
\newblock Implicit neural representation for mesh-free inverse obstacle scattering.
\newblock In {\em Proceedings of the 56th Asilomar Conference on Signals, Systems, and Computers}, pages 947--952, 10 2022.

\bibitem{Kawar_2022}
Bahjat Kawar, Michael Elad, Stefano Ermon, and Jiaming Song.
\newblock Denoising diffusion restoration models.
\newblock In {\em Proceedings of the 36th International Conference on Neural Information Processing Systems}, NIPS '22, Red Hook, NY, USA, 2022. Curran Associates Inc.

\bibitem{khorashadizadeh2023deep}
AmirEhsan Khorashadizadeh, Vahid Khorashadizadeh, Sepehr Eskandari, Guy~AE Vandenbosch, and Ivan Dokmani{\'c}.
\newblock Deep injective prior for inverse scattering.
\newblock {\em IEEE Transactions on Antennas and Propagation}, 2023.

\bibitem{chung2022come}
Hyungjin Chung, Byeongsu Sim, and Jong~Chul Ye.
\newblock Come-closer-diffuse-faster: Accelerating conditional diffusion models for inverse problems through stochastic contraction.
\newblock In {\em Proceedings of the IEEE/CVF Conference on Computer Vision and Pattern Recognition}, pages 12413--12422, 2022.

\bibitem{song2021solving}
Yang Song, Liyue Shen, Lei Xing, and Stefano Ermon.
\newblock Solving inverse problems in medical imaging with score-based generative models.
\newblock {\em arXiv preprint arXiv:2111.08005}, 2021.

\bibitem{chung2022improving}
Hyungjin Chung, Byeongsu Sim, Dohoon Ryu, and Jong~Chul Ye.
\newblock Improving diffusion models for inverse problems using manifold constraints.
\newblock {\em Advances in Neural Information Processing Systems}, 35:25683--25696, 2022.

\bibitem{pmlr-v202-finzi23a}
Marc~Anton Finzi, Anudhyan Boral, Andrew~Gordon Wilson, Fei Sha, and Leonardo Zepeda-Nunez.
\newblock User-defined event sampling and uncertainty quantification in diffusion models for physical dynamical systems.
\newblock In Andreas Krause, Emma Brunskill, Kyunghyun Cho, Barbara Engelhardt, Sivan Sabato, and Jonathan Scarlett, editors, {\em Proceedings of the 40th International Conference on Machine Learning}, volume 202 of {\em Proceedings of Machine Learning Research}, pages 10136--10152. PMLR, 23--29 Jul 2023.

\bibitem{Vaswani_2017:attention_is_all_you_need}
Ashish Vaswani, Noam Shazeer, Niki Parmar, Jakob Uszkoreit, Llion Jones, Aidan~N. Gomez, Lukasz Kaiser, and Illia Polosukhin.
\newblock Attention is all you need.
\newblock 2017.

\bibitem{dasgupta2024conditional}
Agnimitra Dasgupta, Harisankar Ramaswamy, Javier~Murgoitio Esandi, Ken Foo, Runze Li, Qifa Zhou, Brendan Kennedy, and Assad Oberai.
\newblock Conditional score-based diffusion models for solving inverse problems in mechanics.
\newblock {\em arXiv preprint arXiv:2406.13154}, 2024.

\bibitem{baldassari2024conditional}
Lorenzo Baldassari, Ali Siahkoohi, Josselin Garnier, Knut Solna, and Maarten~V de~Hoop.
\newblock Conditional score-based diffusion models for bayesian inference in infinite dimensions.
\newblock {\em Advances in Neural Information Processing Systems}, 36, 2024.

\bibitem{fbp}
D.~Colton and R.~Kress.
\newblock {\em Integral Equation Methods in Scattering Theory}.
\newblock Society for Industrial and Applied Mathematics, Philadelphia, PA, 2013.

\bibitem{Virieux_FWI:2017}
J.~Virieux, A.~Asnaashari, R.~Brossier, L.~M\'etivier, A.~Ribodetti, and W.~Zhou.
\newblock {\em 6. An introduction to full waveform inversion}, pages R1--1--R1--40.
\newblock Society of Exploration Geophysicists, 2017.

\bibitem{fastmri1}
Florian Knoll, Jure Zbontar, Anuroop Sriram, Matthew~J. Muckley, Mary Bruno, Aaron Defazio, Marc Parente, Krzysztof~J. Geras, Joe Katsnelson, Hersh Chandarana, Zizhao Zhang, Michal Drozdzalv, Adriana Romero, Michael Rabbat, Pascal Vincent, James Pinkerton, Duo Wang, Nafissa Yakubova, Erich Owens, C.~Lawrence Zitnick, Michael~P. Recht, Daniel~K. Sodickson, and Yvonne~W. Lui.
\newblock fastmri: A publicly available raw k-space and dicom dataset of knee images for accelerated mr image reconstruction using machine learning.
\newblock {\em Radiology: Artificial Intelligence}, 2(1):e190007, 2020.
\newblock PMID: 32076662.

\bibitem{fastmri2}
Jure Zbontar, Florian Knoll, Anuroop Sriram, Tullie Murrell, Zhengnan Huang, Matthew~J. Muckley, Aaron Defazio, Ruben Stern, Patricia Johnson, Mary Bruno, Marc Parente, Krzysztof~J. Geras, Joe Katsnelson, Hersh Chandarana, Zizhao Zhang, Michal Drozdzal, Adriana Romero, Michael Rabbat, Pascal Vincent, Nafissa Yakubova, James Pinkerton, Duo Wang, Erich Owens, C.~Lawrence Zitnick, Michael~P. Recht, Daniel~K. Sodickson, and Yvonne~W. Lui.
\newblock fastmri: An open dataset and benchmarks for accelerated mri, 2019.

\bibitem{Hormander:FIO}
Lars H\"ormander.
\newblock {\em The Analysis of Linear Partial Differential Operators. {IV}: {F}ourier Integral Operators}, volume~63 of {\em Classics in Mathematics}.
\newblock Springer, Berlin, 2009.

\bibitem{karras2022elucidating}
Tero Karras, Miika Aittala, Timo Aila, and Samuli Laine.
\newblock Elucidating the design space of diffusion-based generative models, 2022.

\bibitem{Plessix_2006:ajoint_state}
R.-E. Plessix.
\newblock {A review of the adjoint-state method for computing the gradient of a functional with geophysical applications}.
\newblock {\em Geophysical Journal International}, 167(2):495--503, 11 2006.

\bibitem{SYMES1991147}
William~W. Symes.
\newblock A differential semblance algorithm for the inverse problem of reflection seismology.
\newblock {\em Computers \& Mathematics with Applications}, 22(4):147--178, 1991.

\bibitem{Borges_Gillman_Greengard:2017}
C.~Borges, A.~Gillman, and L.~Greengard.
\newblock High resolution inverse scattering in two dimensions using recursive linearization.
\newblock {\em SIAM J. Imaging Sci.}, 10(2):641--664, 2017.

\bibitem{Kirsch}
Andreas Kirsch.
\newblock {\em An Introduction to the Mathematical Theory of Inverse Problems}.
\newblock 2021.

\bibitem{Colton_Kress:Integral_Equation_Methods_in_Scattering_Theory}
D.~Colton and R.~Kress.
\newblock {\em Integral Equation Methods in Scattering Theory}.
\newblock Society for Industrial and Applied Mathematics, Philadelphia, PA, 2013.

\bibitem{degroot2012probability}
Morris~H. DeGroot and Mark~J. Schervish.
\newblock {\em Probability and Statistics}.
\newblock Addison-Wesley, 2012.

\bibitem{song2019_score_based}
Yang Song and Stefano Ermon.
\newblock {\em Generative modeling by estimating gradients of the data distribution}.
\newblock Curran Associates Inc., Red Hook, NY, USA, 2019.

\bibitem{song2021scorebased}
Yang Song, Jascha Sohl-Dickstein, Diederik~P. Kingma, Abhishek Kumar, Stefano Ermon, and Ben Poole.
\newblock Score-based generative modeling through stochastic differential equations, 2021.

\bibitem{Luo2022UnderstandingDM}
Calvin Luo.
\newblock Understanding diffusion models: A unified perspective.
\newblock {\em ArXiv}, abs/2208.11970, 2022.

\bibitem{OU_process}
G.~E. Uhlenbeck and L.~S. Ornstein.
\newblock On the theory of the brownian motion.
\newblock {\em Phys. Rev.}, 36:823--841, Sep 1930.

\bibitem{Feynman-Kac_formula}
M.~Kac.
\newblock On distributions of certain wiener functionals.
\newblock {\em Transactions of the American Mathematical Society}, 65(1):1--13, 1949.

\bibitem{nichol2021improveddenoisingdiffusionprobabilistic}
Alex Nichol and Prafulla Dhariwal.
\newblock Improved denoising diffusion probabilistic models, 2021.

\bibitem{Efron2011TweediesFA}
Bradley Efron.
\newblock Tweedie’s formula and selection bias.
\newblock {\em Journal of the American Statistical Association}, 106:1602 -- 1614, 2011.

\bibitem{vincent2011connection}
Pascal Vincent.
\newblock A connection between score matching and denoising autoencoders.
\newblock {\em Neural computation}, 23(7):1661--1674, 2011.

\bibitem{BF}
Yingzhou Li, Haizhao Yang, Eileen~R. Martin, Kenneth~L. Ho, and Lexing Ying.
\newblock Butterfly factorization.
\newblock 13(2):714--732, jan 2015.

\bibitem{Cooley_Tukey:1965}
J.~W. Cooley and J.~W. Tukey.
\newblock An algorithm for the machine calculation of complex {F}ourier series.
\newblock {\em Math. Comput.}, 19(90):297--301, 1965.

\bibitem{Berenger:PML}
J.-P. B\'erenger.
\newblock A perfectly matched layer for the absorption of electromagnetic waves.
\newblock {\em J. Comput. Phys.}, 114(2):185--200, 1994.

\bibitem{matlabprimer8}
Timothy~A. Davis.
\newblock {\em MATLAB Primer, Eighth Edition}.
\newblock CRC Press, Inc., USA, 8th edition, 2010.

\bibitem{SL}
L.~A. Shepp and B.~F. Logan.
\newblock The fourier reconstruction of a head section.
\newblock {\em IEEE Transactions on Nuclear Science}, 21(3):21--43, 1974.

\bibitem{chungshepplogan}
Matthias Chung.
\newblock Random-shepp-logan-phantom.
\newblock \url{https://github.com/matthiaschung/Random-Shepp-Logan-Phantom}, 2018.

\bibitem{deepmind2020jax}
DeepMind, Igor Babuschkin, Kate Baumli, Alison Bell, Surya Bhupatiraju, Jake Bruce, Peter Buchlovsky, David Budden, Trevor Cai, Aidan Clark, Ivo Danihelka, Antoine Dedieu, Claudio Fantacci, Jonathan Godwin, Chris Jones, Ross Hemsley, Tom Hennigan, Matteo Hessel, Shaobo Hou, Steven Kapturowski, Thomas Keck, Iurii Kemaev, Michael King, Markus Kunesch, Lena Martens, Hamza Merzic, Vladimir Mikulik, Tamara Norman, George Papamakarios, John Quan, Roman Ring, Francisco Ruiz, Alvaro Sanchez, Laurent Sartran, Rosalia Schneider, Eren Sezener, Stephen Spencer, Srivatsan Srinivasan, Milo\v{s} Stanojevi\'{c}, Wojciech Stokowiec, Luyu Wang, Guangyao Zhou, and Fabio Viola.
\newblock The {D}eep{M}ind {JAX} {E}cosystem, 2020.

\bibitem{adam}
Diederik~P. Kingma and Jimmy Ba.
\newblock Adam: {A} method for stochastic optimization.
\newblock In Yoshua Bengio and Yann LeCun, editors, {\em 3rd International Conference on Learning Representations, {ICLR} 2015, San Diego, CA, USA, May 7-9, 2015, Conference Track Proceedings}, 2015.

\bibitem{Colton_Kress:Inverse_Acoustic_and_Electromagnetic_Scattering_Theory}
D.~Colton and R.~Kress.
\newblock {\em Inverse Acoustic and Electromagnetic Scattering Theory}.
\newblock Springer-Verlag New York, New York, PA, 3 edition, 2013.

\bibitem{colton2003linear}
David Colton, Houssem Haddar, and Michele Piana.
\newblock The linear sampling method in inverse electromagnetic scattering theory.
\newblock {\em Inverse problems}, 19(6):S105, 2003.

\bibitem{colton1998inverse}
David~L Colton and Rainer Kress.
\newblock {\em Inverse acoustic and electromagnetic scattering theory}, volume~93.
\newblock Springer, 1998.

\bibitem{Oldham1906}
Richard~Dixon Oldham.
\newblock The constitution of the interior of the {E}arth, as revealed by earthquakes.
\newblock {\em Quarterly Journal of the Geological Society}, 62(1-4):456--475, 1906.

\bibitem{Gutenberg1914}
B.~Gutenberg.
\newblock {Ueber Erdbebenwellen. VII A. Beobachtungen an Registrierungen von Fernbeben in Göttingen und Folgerung über die Konstitution des Erdkörpers (mit Tafel)}.
\newblock {\em {Nachrichten von der Gesellschaft der Wissenschaften zu Göttingen, Mathematisch-Physikalische Klasse}}, 1914:125--176, 1914.

\bibitem{Backus_Gilbert:1968}
G.~Backus and F.~Gilbert.
\newblock {The Resolving Power of Gross Earth Data}.
\newblock {\em Geophys. J. Int.}, 16(2):169--205, 10 1968.

\bibitem{stefanov2019travel}
Plamen Stefanov, Gunther Uhlmann, Andras Vasy, and Hanming Zhou.
\newblock Travel time tomography.
\newblock {\em Acta Mathematica Sinica, English Series}, 35(6):1085--1114, 2019.

\bibitem{Sebudandi_1993_travel_time_tomography}
Ch. Sebudandi and Ph.-L. Toint.
\newblock {Non-linear optimization for seismic traveltime tomography}.
\newblock {\em Geophysical Journal International}, 115(3):929--940, 12 1993.

\bibitem{EngquistYing:Sweeping_PML}
B.~Engquist and L.~Ying.
\newblock Sweeping preconditioner for the {H}elmholtz equation: moving perfectly matched layers.
\newblock {\em Multiscale Model. Sim.}, 9(2):686--710, 2011.

\bibitem{Symes_Carazzone:1991}
W.~W. Symes and J.~J. Carazzone.
\newblock Velocity inversion by differential semblance optimization.
\newblock {\em GEOPHYSICS}, 56(5):654--663, 1991.

\bibitem{li2016full}
Yunyue~Elita Li and Laurent Demanet.
\newblock Full-waveform inversion with extrapolated low-frequency data.
\newblock {\em Geophysics}, 81(6):R339--R348, 2016.

\bibitem{ChenDingLiZepeda}
Shi Chen, Zhiyan Ding, Qin Li, and Leonardo Zepeda-N\'{u}\~{n}ez.
\newblock High-frequency limit of the inverse scattering problem: Asymptotic convergence from inverse helmholtz to inverse liouville.
\newblock {\em SIAM Journal on Imaging Sciences}, 16(1):111--143, 2023.

\bibitem{data_driven_internal}
Liliana Borcea, Josselin Garnier, Alexander~V. Mamonov, and J\"{o}rn Zimmerling.
\newblock Waveform inversion with a data driven estimate of the internal wave.
\newblock {\em SIAM Journal on Imaging Sciences}, 16(1):280--312, 2023.

\bibitem{Fichtner2011}
Andreas Fichtner and Jeannot Trampert.
\newblock Resolution analysis in full waveform inversion.
\newblock {\em Geophysical Journal International}, 187(3):1604--1624, Oct 2011.

\bibitem{deBuhan:2017}
M.~de~Buhan and M.~Darbas.
\newblock Numerical resolution of an electromagnetic inverse medium problem at fixed frequency.
\newblock {\em Comput. Math. Appl.}, 74(12):3111 -- 3128, 2017.

\bibitem{dielectric_2019}
Peipei Ran, Yingying Qin, and Dominique Lesselier.
\newblock Electromagnetic imaging of a dielectric micro-structure via convolutional neural networks.
\newblock In {\em 2019 27th European Signal Processing Conference (EUSIPCO)}, pages 1--5, 2019.

\bibitem{Guo_2022}
Rui Guo, Zhichao Lin, Tao Shan, Xiaoqian Song, Maokun Li, Fan Yang, Shenheng Xu, and Aria Abubakar.
\newblock Physics embedded deep neural network for solving full-wave inverse scattering problems.
\newblock {\em IEEE Transactions on Antennas and Propagation}, 70(8):6148--6159, 2022.

\bibitem{ZHOU_warm_start}
Mo~Zhou, Jiequn Han, Manas Rachh, and Carlos Borges.
\newblock A neural network warm-start approach for the inverse acoustic obstacle scattering problem.
\newblock {\em Journal of Computational Physics}, 490:112341, 2023.

\bibitem{Ongie_2020}
Gregory Ongie, Ajil Jalal, Christopher~A. Metzler, Richard~G. Baraniuk, Alexandros~G. Dimakis, and Rebecca Willett.
\newblock Deep learning techniques for inverse problems in imaging.
\newblock {\em IEEE Journal on Selected Areas in Information Theory}, 1(1):39--56, 2020.

\bibitem{Gilton2021DeepEA}
Davis Gilton, Greg Ongie, and Rebecca~M. Willett.
\newblock Deep equilibrium architectures for inverse problems in imaging.
\newblock {\em IEEE Transactions on Computational Imaging}, 7:1123--1133, 2021.

\bibitem{ZehuiZhou}
Zehui Zhou.
\newblock On the recovery of two function-valued coefficients in the helmholtz equation for inverse scattering problems via neural networks.
\newblock {\em Advances in Computational Mathematics}, 51(1):12, 2025.

\bibitem{goodfellow2020generative}
Ian Goodfellow, Jean Pouget-Abadie, Mehdi Mirza, Bing Xu, David Warde-Farley, Sherjil Ozair, Aaron Courville, and Yoshua Bengio.
\newblock Generative adversarial networks.
\newblock {\em Communications of the ACM}, 63(11):139--144, 2020.

\bibitem{Radford2015UnsupervisedRL}
Alec Radford, Luke Metz, and Soumith Chintala.
\newblock Unsupervised representation learning with deep convolutional generative adversarial networks.
\newblock {\em CoRR}, abs/1511.06434, 2015.

\bibitem{Kingma2014a}
Diederik~P. Kingma and Max Welling.
\newblock {Auto-encoding variational Bayes}.
\newblock {\em 2nd International Conference on Learning Representations, ICLR 2014 - Conference Track Proceedings}, (Ml):1--14, 2014.

\bibitem{kingmaNEURIPS2018}
Durk~P Kingma and Prafulla Dhariwal.
\newblock Glow: Generative flow with invertible 1x1 convolutions.
\newblock In S.~Bengio, H.~Wallach, H.~Larochelle, K.~Grauman, N.~Cesa-Bianchi, and R.~Garnett, editors, {\em Advances in Neural Information Processing Systems}, volume~31. Curran Associates, Inc., 2018.

\bibitem{pmlr-v70-bora17a}
Ashish Bora, Ajil Jalal, Eric Price, and Alexandros~G. Dimakis.
\newblock Compressed sensing using generative models.
\newblock In Doina Precup and Yee~Whye Teh, editors, {\em Proceedings of the 34th International Conference on Machine Learning}, volume~70 of {\em Proceedings of Machine Learning Research}, pages 537--546. PMLR, 06--11 Aug 2017.

\bibitem{pmlr-v139-kelkar21a}
Varun~A Kelkar and Mark Anastasio.
\newblock Prior image-constrained reconstruction using style-based generative models.
\newblock In Marina Meila and Tong Zhang, editors, {\em Proceedings of the 38th International Conference on Machine Learning}, volume 139 of {\em Proceedings of Machine Learning Research}, pages 5367--5377. PMLR, 18--24 Jul 2021.

\bibitem{karras_gans}
T.~Karras, S.~Laine, and T.~Aila.
\newblock A style-based generator architecture for generative adversarial networks.
\newblock {\em IEEE Transactions on Pattern Analysis and Machine Intelligence}, 43(12):4217--4228, dec 2021.

\bibitem{ThanhTung2020CatastrophicFA}
Hoang Thanh-Tung and T.~Tran.
\newblock Catastrophic forgetting and mode collapse in gans.
\newblock {\em 2020 International Joint Conference on Neural Networks (IJCNN)}, pages 1--10, 2020.

\bibitem{Kothari2021TrumpetsIF}
Konik Kothari, AmirEhsan Khorashadizadeh, Maarten~V. de~Hoop, and Ivan Dokmani'c.
\newblock Trumpets: Injective flows for inference and inverse problems.
\newblock In {\em Conference on Uncertainty in Artificial Intelligence}, 2021.

\bibitem{pmlr-v119-asim20a}
Muhammad Asim, Max Daniels, Oscar Leong, Ali Ahmed, and Paul Hand.
\newblock Invertible generative models for inverse problems: mitigating representation error and dataset bias.
\newblock In Hal~Daumé III and Aarti Singh, editors, {\em Proceedings of the 37th International Conference on Machine Learning}, volume 119 of {\em Proceedings of Machine Learning Research}, pages 399--409. PMLR, 13--18 Jul 2020.

\bibitem{whang2020compressed}
Jay Whang, Qi~Lei, and Alex Dimakis.
\newblock Compressed sensing with invertible generative models and dependent noise.
\newblock In {\em NeurIPS 2020 Workshop on Deep Learning and Inverse Problems}, 2020.

\bibitem{pmlr-v202-liu23au}
Tianci Liu, Tong Yang, Quan Zhang, and Qi~Lei.
\newblock Optimization for amortized inverse problems.
\newblock In Andreas Krause, Emma Brunskill, Kyunghyun Cho, Barbara Engelhardt, Sivan Sabato, and Jonathan Scarlett, editors, {\em Proceedings of the 40th International Conference on Machine Learning}, volume 202 of {\em Proceedings of Machine Learning Research}, pages 22289--22319. PMLR, 23--29 Jul 2023.

\bibitem{herrmann_WISE}
Ziyi Yin, Rafael Orozco, Mathias Louboutin, and Felix~J. Herrmann.
\newblock Wise: Full-waveform variational inference via subsurface extensions.
\newblock {\em Geophysics}, 89(4):A23--A28, 2024.

\bibitem{herrmann_WISER}
Ziyi Yin, Rafael Orozco, and Felix~J. Herrmann.
\newblock Wiser: multimodal variational inference for full-waveform inversion without dimensionality reduction, 2024.

\bibitem{orozco2024aspireiterativeamortizedposterior}
Rafael Orozco, Ali Siahkoohi, Mathias Louboutin, and Felix~J. Herrmann.
\newblock Aspire: Iterative amortized posterior inference for bayesian inverse problems, 2024.

\bibitem{jacobsen2023cocogen}
Christian Jacobsen, Yilin Zhuang, and Karthik Duraisamy.
\newblock Cocogen: Physically-consistent and conditioned score-based generative models for forward and inverse problems, 2023.

\bibitem{bruna2024posteriorsamplingdenoisingoracles}
Joan Bruna and Jiequn Han.
\newblock Posterior sampling with denoising oracles via tilted transport, 2024.

\bibitem{abramowitz1965handbook}
M.~Abramowitz and I.A. Stegun.
\newblock {\em Handbook of Mathematical Functions: With Formulas, Graphs, and Mathematical Tables}.
\newblock Applied mathematics series. Dover Publications, 1965.

\bibitem{tripleintergralbessel}
A.~Gervois and H.~Navelet.
\newblock {Some integrals involving three Bessel functions when their arguments satisfy the triangle inequalities}.
\newblock {\em Journal of Mathematical Physics}, 25(11):3350--3356, 11 1984.

\bibitem{tancik2020fourier}
Matthew Tancik, Vincent Sitzmann, Julien~N.P. Martel, David~B. Lindell, and Gordon Wetzstein.
\newblock Fourier features let networks learn high frequency functions in low dimensional domains.
\newblock {\em NeurIPS}, 2020.

\bibitem{perez2017filmvisualreasoninggeneral}
Ethan Perez, Florian Strub, Harm de~Vries, Vincent Dumoulin, and Aaron Courville.
\newblock Film: Visual reasoning with a general conditioning layer, 2017.

\bibitem{he2016deep}
Kaiming He, Xiangyu Zhang, Shaoqing Ren, and Jian Sun.
\newblock Deep residual learning for image recognition.
\newblock In {\em Proceedings of the IEEE conference on computer vision and pattern recognition}, pages 770--778, 2016.

\bibitem{iandola2016squeezenetalexnetlevelaccuracy50x}
Forrest~N. Iandola, Song Han, Matthew~W. Moskewicz, Khalid Ashraf, William~J. Dally, and Kurt Keutzer.
\newblock Squeezenet: Alexnet-level accuracy with 50x fewer parameters and <0.5mb model size, 2016.

\bibitem{bao2023worthwordsvitbackbone}
Fan Bao, Shen Nie, Kaiwen Xue, Yue Cao, Chongxuan Li, Hang Su, and Jun Zhu.
\newblock All are worth words: A vit backbone for diffusion models, 2023.

\bibitem{crps_matheson}
James~E. Matheson and Robert~L. Winkler.
\newblock Scoring rules for continuous probability distributions.
\newblock {\em Management Science}, 22(10):1087--1096, 1976.

\bibitem{crps_swirl}
Tilmann Gneiting and Adrian~E Raftery.
\newblock Strictly proper scoring rules, prediction, and estimation.
\newblock {\em Journal of the American Statistical Association}, 102(477):359--378, 2007.

\bibitem{santambrogio2015optimal}
Filippo Santambrogio.
\newblock Optimal transport for applied mathematicians.
\newblock {\em Birk{\"a}user, NY}, 55(58-63):94, 2015.

\bibitem{kantorovich1942translocation}
Leonid~V Kantorovich.
\newblock On the translocation of masses.
\newblock In {\em Dokl. Akad. Nauk. USSR (NS)}, volume~37, pages 199--201, 1942.

\bibitem{cuturi2013sinkhorn}
Marco Cuturi.
\newblock Sinkhorn distances: Lightspeed computation of optimal transport.
\newblock {\em Advances in neural information processing systems}, 26, 2013.

\bibitem{peyre2019computational}
Gabriel Peyr{\'e}, Marco Cuturi, et~al.
\newblock Computational optimal transport: With applications to data science.
\newblock {\em Foundations and Trends{\textregistered} in Machine Learning}, 11(5-6):355--607, 2019.

\bibitem{ramachandran18}
Prajit Ramachandran, Barret Zoph, and Quoc~V. Le.
\newblock Searching for activation functions.
\newblock In {\em 6th International Conference on Learning Representations, {ICLR} 2018, Vancouver, BC, Canada, April 30 - May 3, 2018, Workshop Track Proceedings}. OpenReview.net, 2018.

\bibitem{genevay2018learning}
Aude Genevay, Gabriel Peyr{\'e}, and Marco Cuturi.
\newblock Learning generative models with sinkhorn divergences.
\newblock In {\em International Conference on Artificial Intelligence and Statistics}, pages 1608--1617. PMLR, 2018.

\bibitem{feydy2019interpolating}
Jean Feydy, Thibault S{\'e}journ{\'e}, Fran{\c{c}}ois-Xavier Vialard, Shun-ichi Amari, Alain Trouv{\'e}, and Gabriel Peyr{\'e}.
\newblock Interpolating between optimal transport and {MMD} using sinkhorn divergences.
\newblock In {\em The 22nd International Conference on Artificial Intelligence and Statistics}, pages 2681--2690. PMLR, 2019.

\bibitem{cuturi2022optimal}
Marco Cuturi, Laetitia Meng-Papaxanthos, Yingtao Tian, Charlotte Bunne, Geoff Davis, and Olivier Teboul.
\newblock Optimal transport tools (ott): A {JAX} toolbox for all things {W}asserstein.
\newblock {\em arXiv preprint arXiv:2201.12324}, 2022.

\bibitem{wan2023debias}
Zhong~Yi Wan, Ricardo Baptista, Anudhyan Boral, Yi-Fan Chen, John Anderson, Fei Sha, and Leonardo Zepeda-N{\'u}{\~n}ez.
\newblock Debias coarsely, sample conditionally: Statistical downscaling through optimal transport and probabilistic diffusion models.
\newblock In {\em Thirty-seventh Conference on Neural Information Processing Systems}, 2023.

\bibitem{jax2018github}
James Bradbury, Roy Frostig, Peter Hawkins, Matthew~James Johnson, Chris Leary, Dougal Maclaurin, George Necula, Adam Paszke, Jake Vander{P}las, Skye Wanderman-{M}ilne, and Qiao Zhang.
\newblock {JAX}: composable transformations of {P}ython+{N}um{P}y programs, 2018.

\bibitem{flax2020github}
Jonathan Heek, Anselm Levskaya, Avital Oliver, Marvin Ritter, Bertrand Rondepierre, Andreas Steiner, and Marc van {Z}ee.
\newblock {F}lax: A neural network library and ecosystem for {JAX}, 2023.

\bibitem{Kloeden1992}
Peter~E. Kloeden and Eckhard Platen.
\newblock {\em Numerical Solution of Stochastic Differential Equations}, volume~23 of {\em Stochastic Modelling and Applied Probability}.
\newblock Springer, Berlin, Heidelberg, 1992.

\bibitem{karras2018progressivegrowinggansimproved}
Tero Karras, Timo Aila, Samuli Laine, and Jaakko Lehtinen.
\newblock Progressive growing of gans for improved quality, stability, and variation, 2018.

\end{thebibliography}

\appendix
\renewcommand{\thesection}{\Alph{section}} 
\makeatletter
\renewcommand\@seccntformat[1]{\appendixname\ \csname the#1\endcsname.\hspace{0.5em}}
\makeatother

\section{{Related Works}} \label{sec:related_work}

\subsection{Classical Approaches}\label{sec:classical}

Many non-ML approaches for image reconstruction have been developed over the years. Due to the extensive literature, we focus {on} problems that rely on optimization, and we refer the reader to~\cite{cakoni2005qualitative,colton2003linear,colton1998inverse} for reviews of analysis-based methods.

Among the earliest methods we can find travel-time tomography~\cite{Oldham1906,Gutenberg1914,Backus_Gilbert:1968}, {which} reduces the inverse problem to {the geometrical problem} of finding an {underlying} metric~\cite{stefanov2019travel} by assuming that rays propagate inside the unknown domain satisfying the minimum action principle following the unknown metric. By using the time that each ray takes to travel between points, one can write a non-linear {least-squares} problem~\cite{Sebudandi_1993_travel_time_tomography}, whose solution is used to estimate the {wave speed} inside the medium.
This technique can be cheaply implemented; however, it assumes that the {wave speed} is smooth and the frequency of the propagating waves is high enough to accept a ray approximation. Therefore, the reconstructions quickly deteriorate when highly heterogeneous media or multiple scatterers {are} present.

Following the advent of modern computers, and their increasing capability of numerically solving the underlying PDEs, full-waveform inversion (FWI) was introduced~\cite{Tarantola:Inversion_of_seismic_reflection_data_in_the_acoustic_approximation} in the late 80's. FWI recast the inverse problem as a PDE-constrained optimization problem, where the goal is to minimize the misfit between the real data and synthetic data that comes from the numerical solution of the governing PDE~\cite{Virieux_FWI:2017}. The main advantage with respect to other methods, {lies in its enhanced} capability of handling multiple scattering.

Despite being considered the go-to classical technique for image reconstruction, particularly in geophysical exploration~\cite{Pratt:Seismic_waveform_inversion_in_the_frequency_domain;_Part_1_Theory_and_verification_in_a_physical_scale_model}, {FWI} has some important drawbacks. First, the amount of computational power needed to compute the gradient inside the optimization loop is prohibitive. Even with state-of-the-art solvers~\cite{ZepedaDemanet:the_method_of_polarized_traces,EngquistYing:Sweeping_PML}, the complexity of each iteration is superlinear~\cite{Borges_Gillman_Greengard:2017} with respect to the number of unknowns to recover. Another drawback is the cycle-skipping phenomenon, which refers to the convergence to spurious local minima. 
This is a byproduct of the lack of convexity of the problem, and the lack of low-frequency data, which is usually daunting (and expensive) to acquire.
Efforts to tackle this issue include adding regularization~\cite{Symes_Carazzone:1991,Leeuwen_Herrmann:2013}, extrapolating the data to lower frequency~\cite{li2016full}, modifying the problem to encourage robustness~\cite{ChenDingLiZepeda}, or using a data-driven estimate of the internal wave~\cite{data_driven_internal}.
The last issue is the limitation in the resolution~\cite{Fichtner2011,deBuhan:2017} to recover fine-grained details, due to the diffraction limit.

\subsection{Machine Learning Approaches}\label{sec:ml_approaches}

Most results produced by the classical approaches mentioned above are not yet desirable, thus, fueled by the development of modern ML tools, many ML approaches have been developed in recent years to bypass or attenuate the drawbacks of classical approaches. We divide them into two groups: deterministic and probabilistic approaches.

\textbf{Deterministic}

Generally speaking, most of ML-based methods used for inverse scattering employ a \textit{supervised} trained neural network to regress the scatterer that uses scattered fields as the input data~\cite{YL,dielectric_2019,Guo_2022}. In order to be successful, ML approaches in inverse problems tend to integrate physical and/or mathematical properties of the problem at hand in the architecture of the neural network. These approaches have been proven more successful than their classical counterparts, but they still have some limitations that prevent them {from being} fully applicable.

Some approaches, developed specifically for the inverse scattering problem, improve the performance of a classical approach by leveraging the available training data. In~\cite{ZHOU_warm_start} the authors train a neural network to give a better initial guess to a Gauss-Newton iteration algorithm and have faster and more accurate convergence.  On the other hand, in~\cite{MLZ}, wideband scattering data is deployed to approximate the inverse map.
Very recently, authors in~\cite{melia2024multi} also seek to approximate the inverse map by leveraging wideband scattering data with an iterative refinement approach akin to a Neumann series developed in~\cite{Ongie_2020,Gilton2021DeepEA}. Approaches that involve exploiting the physical structure of the problem, such as embedding rotational equivariant in the neural network construction for a homogeneous background, have also been examined~\cite{YL,equivariant,ZehuiZhou}. 

Although these methods sometimes produce satisfactory reconstructions, they also have significant drawbacks. {The m}ost important drawback is that deterministic ML models typically fail to provide any quantitative measurement of uncertainties, a task {of} paramount importance in reality. Additionally, these deterministic machine learning methods are highly sensitive to experimental configuration. Variations in frequency, and the number of transmitters and receivers, can all significantly impact the quality of the reconstruction.

\textbf{Probabilistic}

Probabilistic ML models automatically account for uncertainties, making them preferred for certain practical problems. Among probabilistic ML models, generative models are the most popular, offering multiple options to choose from: generative adversarial networks (GANs)~\cite{goodfellow2020generative,Radford2015UnsupervisedRL}, variational autoencoders (VAE)~\cite{Kingma2014a}, normalizing flows~\cite{kingmaNEURIPS2018} and diffusion models~\cite{Ho_DDPM2020}. Many of them have already been applied to solve inverse problems~\cite{implicit_neural_vlasic_2022,Kawar_2022}.

Different generative models offer varying performance, each with its own strengths and weaknesses. For instance, GANs have been deployed to learn the prior, and GAN priors have been shown to outperform sparsity priors in some compressive sensing tasks with reduced sample complexity~\cite{pmlr-v70-bora17a,pmlr-v139-kelkar21a,karras_gans}. However, they face challenges in generalization, particularly when the online data is out-of-distribution. Moreover, GANs are prone to catastrophic forgetting, i.e., forgetting previously learned tasks while learning new ones~\cite{ThanhTung2020CatastrophicFA}.

Normalizing flow, another generative model, is also employed to solve inverse problems, either by training the prior or the posterior distribution in Bayes formula~\cite{khorashadizadeh2023deep,Kothari2021TrumpetsIF}. The injectivity property of normalizing flows ensures zero representation error, enabling the recovery of any image, including those significantly out-of-distribution relative to the
training set~\cite{pmlr-v119-asim20a,whang2020compressed,pmlr-v202-liu23au}. Numerical strategies are also integrated to progressively increase dimension from a low-dimensional latent space~\cite{Kothari2021TrumpetsIF} for {enhanced} computational efficiency. In addition to direct sampling, the variational inference framework, serving as an alternative to Bayesian posterior formulation, has been explored in the context of inverse scattering~\cite{herrmann_WISE,herrmann_WISER,orozco2024aspireiterativeamortizedposterior} using normalizing flows. This approach has shown promising experimental results, although it lacks extensive analytical support.

Score-based sampling is another generative modeling approach for solving inverse problems. Usually, a score function is learned to denoise a Gaussian random variable to produce a sample from a desired probability measure. In the context of {the} inverse problem, this probability measure takes the form of a conditional distribution, directing the training to focus on the conditional score function~\cite{dasgupta2024conditional,jacobsen2023cocogen}. Numerical strategies to improve computational efficiency have also been investigated. Notably, the work by~\cite{bruna2024posteriorsamplingdenoisingoracles} introduces an elegant tilted transport technique that exploits the quadratic structure of the log-likelihood function to enhance the convexity of the target distribution. When combined with a learned denoiser for the prior, this method is shown to reach the computational threshold in certain cases. None of the aforementioned {works address the} equivariance structure inherent to the physical problem or {examine} its interaction with the training of the conditioning score function.

\section{Proof of the Proposition~\ref{prop:Fstar_invertible}}\label{sec:appendixA}

We should note that the injectivity of $F^\ast$ plays an important role that allows us to transit from using $\Lambda$ as the condition to using $\alpha_\Lambda$.

Indeed, since $F^\ast$ is injective, it is invertible within the range of $F^\ast$, making $\Lambda=(F^\ast)^{-1}\alpha_{\Lambda}$. In this appendix section, we prove this injectivity.

\begin{lem}[Jacobi–Anger Expansion~\cite{abramowitz1965handbook}]\label{lem:JAE}
   For $ z, \theta \in \mathbb{R} $, the following identity holds:
\begin{equation}
e^{iz \cos \theta} = \sum_{n=-\infty}^{\infty} i^n J_n(z) e^{in\theta}\,,
\end{equation}
where $ J_n(z) $ denotes the $ n $-th Bessel function of the first kind.
\end{lem} 

\begin{lem}\label{lem:besselintegrals}
    For any $ k, n \in \mathbb{N} $ and $ a \in \mathbb{R}^+ $, the following integrals hold:
    \begin{equation}
    \int_0^\infty J_{n}(x)J_{n-k}(x)J_{k}(ax)x\,dx = \frac{\cos(n(\pi-2\phi_a) - k(\pi-\phi_a))}{\pi a \sin(\phi_a)}\,,
    \end{equation}
    \begin{equation}
    \int_0^\infty J_{n}(x)J_{n-k}(x)Y_{k}(ax)x\,dx = \frac{\sin(n(\pi-2\phi_a) - k(\pi-\phi_a))}{\pi a \sin(\phi_a)}\,,
    \end{equation}
    where $ J_n(z) $ is the $ n $-th Bessel function of the first kind, $ Y_n(z) $ is the $ n $-th Bessel function of the second kind, and $ \phi_a $ is the base angle of the isosceles triangle with side lengths 1, 1, and $ a $.
    \begin{proof}
    See Sections B and C in~\cite{tripleintergralbessel} for proofs of the first and second formulas, respectively. 
    \end{proof}
\end{lem}

\begin{prop}[A rewriting of Proposition~\ref{prop:Fstar_invertible}]\label{proof:Fstar_injective}
    The back-scattering operator $(F^\omega)^\ast:L^2([0,2\pi]^2)\rightarrow L^2(\R^2)$ is injective.
    \begin{proof}
    We represent the intermediate field in the polar coordinates, given by~\eqref{eqn:polar_alpha}, we have
\begin{equation}\alpha^\omega(\theta,\rho)=((F^\omega)^* \Lambda^\omega)(\theta,\rho) = \iint_{[0,2\pi]^2} e^{i\omega\rho\cos(r)}e^{-i\omega\rho\cos(s)}\Lambda^\omega(r+\theta,s+\theta)\,dr\,ds\,.\end{equation}
To prove $(F^\omega)^\ast$ is injective, we set $(F^\omega)^* \Lambda^\omega=0$ and we aim to show $\Lambda^\omega = 0$ in $L^2([0,2\pi]^2)$. 

First, we represent $\Lambda^\omega$ by a complex Fourier series
\begin{equation}
\Lambda^\omega(r, s) = \sum_{p,q=-\infty}^\infty c_{p,q} e^{ipr} e^{iqs}\,.
\end{equation}
Thus, applying the phase shift
\begin{equation}
\Lambda^\omega(r+\theta, s+\theta) = \sum_{p,q=-\infty}^\infty c_{p,q} e^{i(p+q)\theta} e^{ipr} e^{iqs}\,.
\end{equation}

Then, using the Jacobi-Anger Expansion (Lemma~\ref{lem:JAE}) and the fact that $J_n(-x)=(-1)^n J_n(x)$, we have
\begin{equation}
e^{i\omega\rho\cos(r)} = \sum_{n=-\infty}^\infty i^n J_n(\omega\rho) e^{inr}\,,
\end{equation}
and
\begin{equation}
e^{-i\omega\rho\cos(s)} = e^{-i\omega\rho\cos(-s)} = \sum_{m=-\infty}^\infty (-i)^m J_m(\omega\rho) e^{-ims}\,.
\end{equation}

Substituting the expansions and the shifted $\Lambda^\omega$ into the integral, we have

\begin{equation}
    \begin{aligned}
        &\iint_{[0,2\pi]^2} \left(\sum_{n=-\infty}^\infty i^n J_n(\omega\rho) e^{inr}\right) \left(\sum_{m=-\infty}^\infty (-i)^m J_m(\omega\rho) e^{-ims}\right) \left(\sum_{p,q=-\infty}^\infty c_{p,q} e^{i(p+q)\theta} e^{ipr} e^{iqs}\right) \,ds\,dr \,,\\
&= \sum_{n,m,p,q=-\infty}^\infty i^n (-i)^m J_n(\omega\rho) J_m(\omega\rho) c_{p,q} e^{i(p+q)\theta} \iint_{[0,2\pi]^2} e^{i(n+p)r} e^{i(-m+q)s} \, dr \, ds\,,
    \end{aligned}
\end{equation}
where the equality follows from Fubini's theorem.

The exponential functions $e^{inr}$ and $e^{ims}$ are orthogonal over $[0, 2\pi]$
\begin{equation}
\int_0^{2\pi} e^{i(n+p)r} \, dr = 2\pi \delta_{n+p,0}\,, \quad \int_0^{2\pi} e^{i(-m+q)s} \, ds = 2\pi \delta_{-m+q,0}\,.
\end{equation}

Substituting back and simplifying, for almost every $\theta\in[0,2\pi]$ and $\rho\in[0,\infty)$, we have
\begin{equation}\sum_{n,m=-\infty}^\infty i^{n-m} J_n(\omega\rho) J_m(\omega\rho) c_{-n,m} e^{-i(n-m)\theta} = 0\,.\end{equation}

By using Fubini's theorem and setting $k = n - m$, our expression becomes
\begin{equation}
\sum_{k=-\infty}^\infty  i^{k}\left(\sum_{n=-\infty}^\infty J_n(\omega\rho) J_{n-k}(\omega\rho) c_{-n,n-k}\right) e^{-ik\theta}  = 0\,.
\end{equation}
For almost every $\rho\in[0,\infty)$, the series on the left can be viewed as a Fourier series in $\theta$ of a zero function. Consequently, all Fourier coefficients must be zero
\begin{equation}\sum_{n=-\infty}^\infty J_n(\omega\rho) J_{n-k}(\omega\rho) c_{-n,n-k} = 0 \quad \forall k\in\Z\,.\end{equation}

Now fix a $k\in\Z$, for all $a\in\R^+$, we multiply the series on the left by $J_k(a\omega\rho)\omega^2\rho$ and integrate over $\rho$ from $0$ to $\infty$. By Fubini's theorem and Lemma~\ref{lem:besselintegrals}, we have
\begin{align*}
    &\int_0^\infty\left(\sum_{n=-\infty}^\infty J_n(\omega\rho) J_{n-k}(\omega\rho) c_{-n,n-k} \right)J_k(a\omega\rho)\omega^2\rho\,d\rho \,,\\
    =& \sum_{n=-\infty}^\infty c_{-n,n-k} \int_0^\infty  J_n(\omega\rho) J_{n-k}(\omega\rho)J_k(a\omega\rho) \omega\rho\,d(\omega\rho)\,,\\
    =&\sum_{n=-\infty}^\infty c_{-n,n-k}\frac{\cos(n(\pi-2\phi_a) - k(\pi-\phi_a))}{\pi a\sin(\phi_a)}\,,
\end{align*}
where $\phi_a\in(0,\pi)$ for $a\in\R^+$.
Hence, our expression becomes
    \begin{equation}\sum_{n=-\infty}^\infty c_{-n,n-k}\cos(n(\pi-2\phi_a) - k(\pi-\phi_a))=0\,.\end{equation}
By Lemma~\ref{lem:besselintegrals}, and a similar computation with the series multiplied by $Y_k(a\omega\rho)\omega^2\rho$, we have
    \begin{equation}\sum_{n=-\infty}^\infty c_{-n,n-k}\sin(n(\pi-2\phi_a) - k(\pi-\phi_a))=0\,.\end{equation}
Therefore, by combining the two and simplifying, for all $a\in\R^+\implies \pi-2\phi_a\in (-\pi,\pi)$, we have
    \begin{equation}\sum_{n=-\infty}^\infty c_{-n,n-k}e^{in(\pi-2\phi_a)}=0\,.\end{equation}
We can view the sum as a Fourier series in $\phi_a\in(0,\pi)$ of a zero function. Hence,  coefficients $c_{-n,n-k}$ for $k,n\in\N$ must be $0$. 
Since $k$ is arbitrary, all Fourier coefficients $c_{k,l}$ in the original representation of $\Lambda^\omega$ are $0$. By Parseval's theorem, we have $\Lambda^\omega=0$. Therefore, we can conclude that $(F^\omega)^\ast$ is injective.
\end{proof}
\end{prop}

\section{Proof of the Theorem~\ref{thm:score_function_TE}}\label{sec:appendixB}
Throughout this section, we define the inner product on $\R^{n_\eta\times n_\eta}$ by the usual dot product
\begin{equation}\inner{v,w} = v\cdot w\,.\end{equation}

\begin{lem}\label{lem:inverse_translation}
    The inverse of the translation operator $ T_{\bm{a}} $ is $T_{-\bm{a}}$:
    \begin{equation}
   (T_{\bm{a}})^{-1} = T_{-\bm{a}}\,.
    \end{equation}

    \begin{proof}
        For all $v\in\R^{n_\eta\times n_\eta}$, $\bm{a}\in\N_\eta^2$ and $\bm{y}\in\N_\eta^2$, we have
        {\begin{equation}((T_{\bm{a}}\circ T_{-\bm{a}})v)_{\bm{y}} = v_{\tau_{-\bm{a}}\circ\tau_{\bm{a}}(\bm{y})}=v_{\bm{y}}\,.\end{equation}}
        Therefore, we have
        \begin{equation}T_{\bm{a}}\circ T_{-\bm{a}} = \bm{I}\,,\end{equation}
        where $\bm{I}$ is the identity operator on $\R^{n_\eta\times n_\eta}$.
    \end{proof}
\end{lem}

\begin{lem}\label{lem:unitarity_jacobian}
    The translation operator $ T_{\bm{a}} $ is a unitary linear transformation.  Consequently, the Jacobian $J_{T_{\bm{a}}}$ of the change of variables under the transformation $T_{\bm{a}}$ satisfies
    \begin{equation}|\det J_{T_{\bm{a}}}| = 1\,,\end{equation}
    and by Lemma~\ref{lem:inverse_translation}, we have
\begin{equation}\inner{T_{\bm{a}}v, w}=\inner{v, T_{-\bm{a}}w}\,.\end{equation}
    \begin{proof}
    for any $v,w\in\R^{n_\eta\times n_\eta}$, and any $\bm{a}\in\N_\eta^2$, 
    as $T_{\bm{a}}$ is invertible, it suffices to show
        {\begin{equation}\inner{T_{\bm{a}}v, T_{\bm{a}}w} = \sum_{\bm{y}\in\N_\eta^2} v_{\tau_{-\bm{a}}(\bm{y})}w_{\tau_{-\bm{a}}(\bm{y})} = \sum_{\bm{y}\in\N_\eta^2} v_{\bm{y}}w_{\bm{y}}=\inner{v,w}\,.\end{equation}}
    \end{proof}
\end{lem}

\begin{lem}\label{lem:discrete_filtering_TE}
Let $\sfF^\omega$ be the discretized forward map, and $(\sfF^\omega)^*$ be the discretized back-scattering operator, see Section~\ref{sec:discretization}. Then, for any $\bm{a}\in\N_\eta^2$, $(\sfF^\omega)^*\sfF^\omega$ commutes with the translation operator $T_{\bm{a}}$, i.e. \begin{equation}((\sfF^\omega)^*\sfF^\omega)\circ T_{\bm{a}} = T_{\bm{a}}\circ ((\sfF^\omega)^*\sfF^\omega)\,.\end{equation}
\begin{proof}
For the discretization of $\eta$, see Section~\ref{sec:discretization}, we used a Cartesian mesh of $n_\eta\times n_\eta$ grids for the physical domain $\Omega = [-0.5,0.5]^2$. We further denote the grid points by $\bm{x}_{(i,j)}$, so that
\begin{equation}\bm{\eta}_{(i,j)} = \eta(\bm{x}_{(i,j)})\,.\end{equation}
Similarly, we denote the grid points from the discretization of $\mathbb{S}^1$ by $\bm{s}_{(n)}$ and $\bm{r}_{(m)}$,  where 
\begin{equation}
\bm{s}_{(n)} = (\cos(s_n), \sin(s_n))\text{\quad and\quad}\bm{r}_{(m)} = (\cos(r_m), \sin(r_m))\,.
\end{equation}
For any $\bm{x}\in\R^2$, we define $\bm{x} \mod \Omega$ as the operation that maps $\bm{x}$ to the physical domain $\Omega$ by 
\begin{equation}
\bm{x} \mod \Omega = (( \bm{x} + \bm{0.5} ) \mod \bm{1}) - \bm{0.5}\,,
\end{equation}
where $\bm{c}=(c,c)\in\R^2$, and the modulo operation is applied element-wise. 

Then, we have the following identity
\begin{equation}\label{eqn:translation_identity}(T_{\bm{a}}\bm{\eta})(\bm{x})=\eta(\bm{x}-\bm{a}/n_\eta\mod \Omega)\,.\end{equation}

It is straightforward to find 
\begin{equation}\label{eq:90}
\begin{aligned}
    ((\sfF^\omega)^*\sfF^\omega\bm{\eta})(\bm{y}) &= \sum_{n,m=1}^{n_\sca} e^{i\omega(\bm{r}_{(m)}-\bm{s}_{(n)})\cdot \bm{y}}\sum_{i,j=1}^{n_\eta}e^{-i\omega(\bm{r}_{(m)}-\bm{s}_{(n)})\cdot\bm{x}_{(i,j)}}\eta(\bm{x}_{(i,j)})\,,\\
    & = \sum_{n,m=1}^{n_\sca} \sum_{i,j=1}^{n_\eta}e^{i\omega(\bm{r}_{(m)}-\bm{s}_{(n)})\cdot(\bm{y}-\bm{x}_{(i,j)})}\eta(\bm{x}_{(i,j)})\,.\\
\end{aligned}
\end{equation}
By applying a translation operator and {using} the identity in~\eqref{eqn:translation_identity}, similar to~\eqref{eq:90}, we have
\begin{equation}
\begin{aligned}
    ((\sfF^\omega)^*\sfF^\omega T_{\bm{a}}\bm{\eta})(\bm{y})&= \sum_{n,m=1}^{n_\sca} \sum_{i,j=1}^{n_\eta}e^{i\omega(\bm{r}_{(m)}-\bm{s}_{(n)})\cdot(\bm{y}-\bm{x}_{(i,j)})}\eta(\bm{x}_{(i,j)}-\bm{a}/n_\eta\mod \Omega)\,,\\
    &= \sum_{n,m=1}^{n_\sca} \sum_{i,j=1}^{n_\eta}e^{i\omega(\bm{r}_{(m)}-\bm{s}_{(n)})\cdot(\bm{y}-(\bm{x}_{(i,j)}+\bm{a}/n_\eta\mod \Omega))}\eta(\bm{x}_{(i,j)})\,,\\
    &= \sum_{n,m=1}^{n_\sca} \sum_{i,j=1}^{n_\eta}e^{i\omega(\bm{r}_{(m)}-\bm{s}_{(n)})\cdot((\bm{y}-\bm{a}/n_\eta\mod \Omega)-\bm{x}_{(i,j)})}\eta(\bm{x}_{(i,j)})\,,\\
    &= (T_{\bm{a}}(\sfF^\omega)^*\sfF^\omega\bm{\eta})(\bm{y})\,.
\end{aligned}
\end{equation}
Since $\eta$ is arbitrary, we can conclude that 
\begin{equation}(\sfF^\omega)^*\sfF^\omega\circ T_{\bm{a}} = T_{\bm{a}}\circ(\sfF^\omega)^*\sfF^\omega\,.\end{equation}
\end{proof}
\end{lem}

\begin{lem}\label{lem:invariance_equivariance}
    Suppose that an operator $\mathcal{P}\in C^1(\R^{n_\eta\times n_\eta})^{\otimes 2}$ acting on $(v,w)\in(\R^{n_\eta\times n_\eta})^{\otimes 2}$ satisfies translational invariance 
    \begin{equation}
        \mathcal{P}(T_{\bm{a}} v, T_{\bm{a}} w) = \mathcal{P}(v,w) \quad \forall \bm{a} \in \N_\eta^2\,,
    \end{equation}
    then the gradient of $\mathcal{P}$ with respect to $v$, denoted as $\nabla_v\mathcal{P}$, is translationally equivariant.
\begin{proof}
For any $\bm{a}\in\N_\eta^2$, by the definition of the gradient, for a small perturbation $ \epsilon h$ such that $\epsilon>0$ and $h\in\R^{n_\eta\times n_\eta}$:
\begin{equation}\label{eq:94}
\mathcal{P}(T_{\bm{a}}v + \epsilon h, T_{\bm{a}}w) = \mathcal{P}(T_{\bm{a}}v, T_{\bm{a}}w) + \epsilon \inner{\nabla_v \mathcal{P} (T_{\bm{a}}v, T_{\bm{a}}w), h}  + o(\epsilon)\,.
\end{equation}

By the assumption in this lemma, the fact that the inverse operator of $T_{\bm{a}}$ is $T_{-\bm{a}}$ as shown in Lemma~\ref{lem:inverse_translation}, and $T_{\bm{a}}$ is unitary as shown in Lemma~\ref{lem:unitarity_jacobian}, we have
\begin{equation}\label{eq:95}
    \begin{aligned}
        \mathcal{P}(T_{\bm{a}}v + \epsilon h, T_{\bm{a}}w) &= \mathcal{P}(v + \epsilon T_{-\bm{a}}h, w)\,,\\ 
        &=\mathcal{P}(v,w) + \epsilon \inner{\nabla_v \mathcal{P} (v,w), T_{-\bm{a}}h} + o(\epsilon)\,,\\
        &= \mathcal{P}(T_{\bm{a}}v,T_{\bm{a}}w) + \epsilon \inner{T_{\bm{a}}\nabla_v \mathcal{P} (v,w), h} + o(\epsilon)\,.
    \end{aligned}
\end{equation}

Comparing~\eqref{eq:94} and~\eqref{eq:95}, noting that $h\in\R^{n_\eta\times n_\eta}$ is arbitrary, we have:
\begin{equation}
    \nabla_v \mathcal{P} (T_{\bm{a}}v, T_{\bm{a}}w) = T_{\bm{a}}\nabla_v \mathcal{P} (v,w) \quad \forall a \in \N_\eta^2\,,
\end{equation}
concluding the lemma.
\end{proof}
\end{lem}

\begin{lem}\label{lem:transilation_invariance_blocks}
With the same assumptions as in Theorem~\ref{thm:score_function_TE}, 
denote $\bm{\eta}_0$ the ground truth perturbation that corresponds to the intermediate field $\bm{\alpha}$, we have
\begin{enumerate}
    \item The operator $\mathcal{U}: (\R^{n_\eta\times n_\eta})^{\otimes 2} \rightarrow \mathbb{R}$ defined as $\mathcal{U}(\bm{\eta},\bm{\eta}_0) = p_t(\bm{\eta}|\bm{\eta}_0)$ is translationally invariant.
    \item  The operator $\mathcal{W}: (\R^{n_\eta\times n_\eta})^{\otimes 2} \rightarrow \mathbb{R}$ defined as $\mathcal{W}(\bm{\alpha},\bm{\eta}_0) = p_0(\bm{\eta}_0|\bm{\alpha})$ is translationally invariant.
\end{enumerate}
\begin{proof}
\
\begin{enumerate}
        \item As discussed in Section~\ref{sec:diffusion_model_preliminary}, $p_{t}(\bm{\eta}|\bm{\eta}_0)$ is the perturbation kernel (see~\eqref{eqn:perturbation_kernel}): 
\begin{equation}
    p_{t}(\bm{\eta}|\bm{\eta}_0) = \mathcal{N}(\bm{\eta}; s(t)\bm{\eta}_0, s^2(t) \sigma^2(t)\bm{I})= \exp\left(-\frac{\|\bm{\eta} - s(t)\bm{\eta}_0\|^2}{2s^2(t) \sigma^2(t)}\right) \cdot \frac{1}{Z}\,,
\end{equation}
where $ Z $ is the normalization constant.

Since the translation operator is unitary, $\forall \bm{a} \in \N_\eta^2$, we have
\begin{align*}
    p_t(T_{\bm{a}}\bm{\eta} | T_{\bm{a}}\bm{\eta}_0) &= \exp\left(-\frac{\|T_{\bm{a}}(\bm{\eta} - s(t)\bm{\eta}_0)\|^2}{2s^2(t) \sigma^2(t)}\right) \cdot \frac{1}{Z}\,,\\
    &= \exp\left(-\frac{\|\bm{\eta} - s(t)\bm{\eta}_0\|^2}{2s^2(t) \sigma^2(t)}\right) \cdot \frac{1}{Z}\,,\\
    &= p_t(\bm{\eta} | \bm{\eta}_0).
\end{align*}
        
\item 
    Notice that the marginal distributions $p_0(\bm{\eta})=p_{\text{data}}(\bm{\eta})$. By the assumption that $\mathcal{P}(\bm{\alpha},\bm{\eta}_0) = p_0(\bm{\alpha}|\bm{\eta}_0)$ and $p_{\text{data}}(\bm{\eta})$ are translationally invariant, for all $\bm{a} \in \N_\eta^2$, we have
    \begin{equation}
    p_0(T_{\bm{a}}\bm{\alpha},T_{\bm{a}}\bm{\eta}_0) = p_0(T_{\bm{a}}\bm{\alpha}|T_{\bm{a}}\bm{\eta}_0)p_{\text{data}}(T_{\bm{a}}\bm{\eta}_0) = p_0(\bm{\alpha}|\bm{\eta}_0)p_{\text{data}}(\bm{\eta}_0)  = p_0(\bm{\alpha},\bm{\eta}_0)\,.
    \end{equation}
    
    Now it suffices to show the marginal distributions 
    \begin{equation}
        p_0(T_{\bm{a}}\bm{\alpha}) = p_0(\bm{\alpha})\quad \forall \bm{a} \in \N_\eta^2\,.
    \end{equation}
    According to the definition, this amounts to show
    \begin{equation}
        \int_{\bm{\eta}_0\in\R^{n_\eta\times n_\eta}}  p_0(T_{\bm{a}}\bm{\alpha}, \bm{\eta}_0) \,d\bm{\eta}_0= \int_{\bm{\eta}_0\in\R^{n_\eta\times n_\eta}} p_0(\bm{\alpha},\bm{\eta}_0)\,d\bm{\eta}_0\quad \forall \bm{a} \in \N_\eta^2\,.
    \end{equation}
    This is true because:
    \begin{equation}
        \begin{aligned}
         p_0(T_{\bm{a}}\bm{\alpha}) &= \int_{\bm{\eta}_0\in\R^{n_\eta\times n_\eta}}  p_0(T_{\bm{a}}\bm{\alpha}, \bm{\eta}_0) \,d\bm{\eta}_0\,,\\
            &=\int_{\bm{\zeta}_0\in\R^{n_\eta\times n_\eta}} p_0(T_{\bm{a}}\bm{\alpha}, T_{\bm{a}}\bm{\zeta}_0)|\det J_{T_{\bm{a}}}|\, d\bm{\zeta}_0\,,\\
            &=\int_{\bm{\zeta}_0\in\R^{n_\eta\times n_\eta}} p_0(\bm{\alpha}, \bm{\zeta}_0)\, d\bm{\zeta}_0\,,\\
            &= p_0(\bm{\alpha})\quad \forall \bm{a} \in \N_\eta^2\,.
        \end{aligned}
    \end{equation}
    where we used change of variable $\bm{\eta}_0 = T_{\bm{a}}\bm{\zeta}_0$ in the second equation, and applied Lemma~\ref{lem:unitarity_jacobian} in the third. Then using the definition of the conditional distribution,
\begin{equation}
p_0(T_{\bm{a}}\bm{\eta}_0|T_{\bm{a}}\bm{\alpha}) = \frac{p_0(T_{\bm{a}}\bm{\alpha},T_{\bm{a}}\bm{\eta}_0)}{p_0(T_{\bm{a}}\bm{\alpha})} = \frac{p_0(\bm{\alpha},\bm{\eta}_0)}{p_0(\bm{\alpha})} = p_0(\bm{\eta}_0|\bm{\alpha}) \quad \forall \bm{a} \in \N_\eta^2\,,
\end{equation}
concluding the proof.
    \end{enumerate}
\end{proof}
\end{lem}

\begin{thm}[A rewriting of Theorem~\ref{thm:score_function_TE}]\label{proof:score_function_TE}
With $\bm{\alpha}$, $\bm{\eta}$ and $p_t$ defined above, if $\mathcal{P}(\bm{\alpha},\bm{\eta})=p_0(\bm{\alpha} | \bm{\eta})$ and $p_\data(\bm{\eta})$ are both translationally invariant, then the physics-aware score function $\nabla_{\bm{\eta}}\log p_t(\bm{\eta}|\bm{\alpha})$ is translationally equivariant. More specifically,
assume: $p_0(T_{\bm{a}}\bm{\alpha} | T_{\bm{a}}\bm{\eta})=p_0(\bm{\alpha} | \bm{\eta})$
and $p_\data(T_{\bm{a}}\bm{\eta}) = p_\data(\bm{\eta})$ for all $\bm{a} \in \N_\eta^2$, then for $p_t \in C^1(\mathbb{R}^{n_\eta \times n_\eta})^{\otimes 2}$ and $p_t > 0$:
\begin{equation}
\nabla_{\bm{\eta}}\log p_t(T_{\bm{a}}\bm{\eta}| T_{\bm{a}}\bm{\alpha}) = T_{\bm{a}}\nabla_{\bm{\eta}}\log p_t(\bm{\eta}|\bm{\alpha}) \quad \forall \bm{a} \in \mathbb{N}_\eta^2\,,
\end{equation}
i.e. $\mathcal{Q}(\bm{\eta},\bm{\alpha})=\nabla_{\bm{\eta}}\log p_t(\bm{\eta}|\bm{\alpha})$ is {translationally} equivariant.

\begin{proof}
From the forward SDE~\eqref{eqn:forwardSDE}, it can be seen that conditioned on $\bm{\eta}_0$, $\bm{\eta}_t$ and $\bm{\alpha}$ are independent, so
    \begin{equation}\label{eq:independence_alpha}p_{t}(\bm{\eta},\bm{\alpha}|\bm{\eta}_0)= p_{t}(\bm{\eta}|\bm{\eta}_0)p_0(\bm{\alpha}|\bm{\eta}_0)\implies p_{t}(\bm{\eta}|\bm{\alpha},\bm{\eta}_0) = p_{t}(\bm{\eta}|\bm{\eta}_0)\,.\end{equation}
    As a consequence
    \begin{equation}
        \begin{aligned}
            p_t(\bm{\eta} | \bm{\alpha}) &= \int_{\bm{\eta}_0\in\R^{n_\eta\times n_\eta}} p_t(\bm{\eta},\bm{\eta}_0|\bm{\alpha})\, d\bm{\eta}_0\,,\\
            &= \int_{\bm{\eta}_0\in\R^{n_\eta\times n_\eta}} p_t(\bm{\eta}|\bm{\eta}_0,\bm{\alpha})p_0(\bm{\eta}_0|\bm{\alpha})\, d\bm{\eta}_0\,,\\
            &= \int_{\bm{\eta}_0\in\R^{n_\eta\times n_\eta}} p_t(\bm{\eta} | \bm{\eta}_0)p_0(\bm{\eta}_0| \bm{\alpha})\, d\bm{\eta}_0\,.\\
        \end{aligned}
    \end{equation}
   where we used~\eqref{eq:independence_alpha} in the third equation. Similarly,
\begin{equation}
        \begin{aligned}
            p_t(T_{\bm{a}}\bm{\eta} | T_{\bm{a}}\bm{\alpha}) 
            &= \int_{\bm{\eta}_0\in\R^{n_\eta\times n_\eta}} p_t(T_{\bm{a}}\bm{\eta} | \bm{\eta}_0)p_0(\bm{\eta}_0|T_{\bm{a}}\bm{\alpha})\, d\bm{\eta}_0\,,\\
            &=\int_{\bm{\zeta}_0\in\R^{n_\eta\times n_\eta}} p_t(T_{\bm{a}}\bm{\eta} | T_{\bm{a}}\bm{\zeta}_0)p_0(T_{\bm{a}}\bm{\zeta}_0|T_{\bm{a}}\bm{\alpha})|\det J_{T_{\bm{a}}}|\, d\bm{\zeta}_0\,,\\
            & = \int_{\bm{\zeta}_0\in\R^{n_\eta\times n_\eta}} p_t(\bm{\eta} | \bm{\zeta}_0)p_0(\bm{\zeta}_0|\bm{\alpha})\, d\bm{\zeta}_0\,,\\
            &= p_t(\bm{\eta} | \bm{\alpha})\,,
        \end{aligned}
    \end{equation}
    where we used a change of variable $\bm{\eta}_0 = T_{\bm{a}}\bm{\zeta}_0$ in the second equation, and called Lemma~\ref{lem:unitarity_jacobian}, and Lemma~\ref{lem:transilation_invariance_blocks} in the third equation.
    Consequently, the operator $\mathcal{R}(\bm{\eta},\bm{\alpha}) = \log p_t(\bm{\eta} | \bm{\alpha})$ satisfies translational invariance.
    By Lemma~\ref{lem:invariance_equivariance}, we can conclude that $\nabla_{\bm{\eta}}\mathcal{R}(\bm{\eta},\bm{\alpha}) = \nabla_{\bm{\eta}}\log p_t(\bm{\eta} | \bm{\alpha})$ satisfies translational equivariance.
\end{proof}
\end{thm}

\section{{CNN-Based Representation}}\label{sec:CNN}
Recall from Section~\ref{sec:DDPMformulation} that we have the analytical solution to the forward problem. Rewriting~\eqref{eq:solution_F-P_langevin} in the current context for the conditioning distribution, we have:
\begin{equation}
p_t(\bm{\eta}|\bm{\alpha}_{\sf\Lambda}) = p\left(\frac{\bm{\eta}}{s(t)}; \sigma(t)\Big|\bm{\alpha}_{\sf\Lambda}\right)\propto (p_\data\left(\,\cdot\,|\bm{\alpha}_{\sf\Lambda}\right)*\mathcal{N}(\,\cdot\,;\bm{0},\sigma^2(t)\bm{I}))\left(\frac{\bm{\eta}}{s(t)}\right)\,.
\end{equation}
Therefore, in the offline learning stage, the equivalent form of loss function~\eqref{eq:denoiser_score} is defined to find the denoiser for the conditional distribution for all $\sigma$.

To include the dependence of the noise level $\sigma$ in the training of the NN, we adopt the common approach through Fourier embedding~\cite{tancik2020fourier} and FiLM technique~\cite{perez2017filmvisualreasoninggeneral}. 

In a nutshell, Fourier embedding is an embedding technique that maps the noise level $\sigma$ into a higher-dimensional space using sinusoidal functions. To build the features, one creates a grid of logarithmically spaced frequencies $\omega_k$ which then are used to modulate $\sigma$ using sinusoidal functions, {i.e.}, $\sin(\pi \omega_k \, \sigma)$ and $\cos(\pi \omega_k \, \sigma)$. The output is then concatenated to form Fourier features, which are then fed through dense layers to create the Fourier embedding (see Algorithm~\ref{alg:FourierEmbedding}).
The Fourier embedding of the noise variable is then integrated into the model using FiLM, which adaptively modulates the neural network by applying an affine transformation to the hidden neurons, see Algorithm~\ref{alg:film}. 

\begin{figure}[h!]
  \centering
  \includegraphics[width=\textwidth]{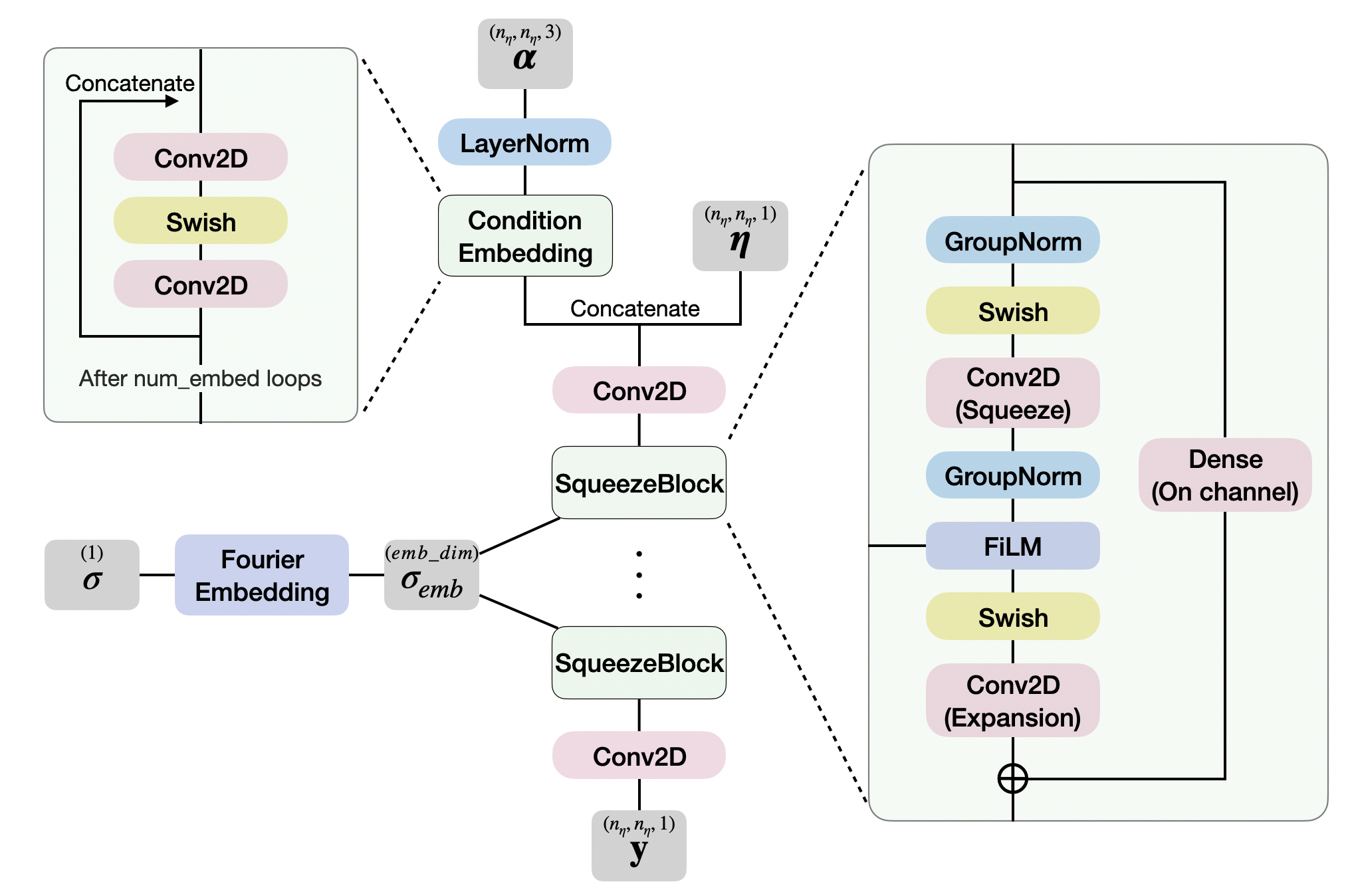}
  \caption{An overview of the architecture of the CNN-based representation.}\label{fig:CNN_architecture}
\end{figure}

For our CNN-based representation of the score function, we consider a network with three inputs: the noisy sample $\bm{\eta}_t$, the latent variable $\bm{\alpha}_{\sf\Lambda}$ that acts as conditioning, and the noise level $\sigma$ that modulates the rest of the network. 
We process the conditioning variable  $\bm{\alpha}_{\sf\Lambda}$ by a sequence of convolutional residual blocks with Swish functions as shown in Figure~\ref{fig:CNN_architecture}. The processed conditional input is merged with the noised $\bm{\eta}_t$ along the channel dimension. 

This merged conditional input and noised samples {are} then fed to a sequence of modified convolutional residual blocks~\cite{he2016deep} which we call \texttt{SqueezeBlocks}. These blocks are modulated with the Fourier embeddings stemming from the noise input $\sigma$.
The \texttt{SqueezeBlocks}, as specified in Algorithm~\ref{alg:SqueezeBlock}, are residual blocks that leverage a SqueezeNet~\cite{iandola2016squeezenetalexnetlevelaccuracy50x}, which reduces the number of features in the first convolution layer, as shown in Figure~\ref{fig:CNN_architecture}.

The overall architecture of our CNN-based representation is detailed in Algorithm~\ref{alg:CNN}, with {a} graphical overview shown in Figure~\ref{fig:CNN_architecture}.

\begin{remark}\label{rem:UViT}
Other choices are available too, and they will be used in {the} numerical section for comparison:
U-Net Vision Transformer (U-ViT): The U-ViT architecture~\cite{bao2023worthwordsvitbackbone} follows a U-Net structure with a downsampling path to encode the input image into feature maps, and an upsampling path to decode these feature maps back to the original spatial dimensions. The attention mechanisms enhance its ability to capture long-range dependencies.  We use the implementation in a public repository\footnote{\url{https://github.com/google-research/swirl-dynamics/blob/main/swirl_dynamics/lib/diffusion/unets.py}} and provide a summarized skeleton of the algorithm in~\ref{alg:UViT}.  We note that the \texttt{ConvBlock} in the algorithm is a special case of the \texttt{SqueezeBlock}, where the parameters \texttt{out\_channels} and \texttt{squeeze\_channels} are set to be equal, \texttt{PositionEmbedding} adds a trainable 2D position embedding, and \texttt{AttentionBlock} uses a multi-head dot product attention coupled with a residual connection.   
\end{remark}

\begin{algorithm}[h!]
\caption{Fourier Embedding}\label{alg:FourierEmbedding}
\begin{algorithmic}[1]
\Procedure{\texttt{FourierEmbedding}}{$\sigma$}
%
  \State \quad $\texttt{logfreqs} \gets \texttt{Linspace}(\texttt{0, log(max\_freq), emb\_dim // 2})$
  \State \quad $\sigma_{\text{freq}} \gets \pi \cdot\texttt{exp}(\texttt{logfreqs}) \cdot \sigma$
  \State \quad $\sigma_{\text{emb}} \gets \texttt{Concatenate}(\texttt{[sin}(\sigma_{\text{freq}})\texttt{,cos}(\sigma_{\text{freq}})\texttt{], axis=-1})$
  \State \quad $\sigma_{\text{emb}} \gets \texttt{Dense}(\texttt{features=2}\cdot\texttt{emb\_dim})(\sigma_{\text{emb}})$
  \State \quad $\sigma_{\text{emb}} \gets \texttt{Swish}(\sigma_{\text{emb}})$
  \State \quad $\sigma_{\text{emb}} \gets \texttt{Dense}(\texttt{features=emb\_dim})(\sigma_{\text{emb}})$ 
  \State \Return $\sigma_{\text{emb}}$
\EndProcedure
\end{algorithmic}
\end{algorithm}

\begin{algorithm}[h!]
\caption{FiLM}\label{alg:film}
\begin{algorithmic}[1]
    \Procedure{\texttt{FiLM}}{$\bm{x}\texttt{,}\sigma_{\text{emb}}$}
      \State \quad $\texttt{Affine} \gets \texttt{Dense}(\texttt{features=2}\cdot\bm{x}\texttt{.shape[-1]})$
  \State \quad $\texttt{params} \gets \texttt{Affine}(\texttt{Swish}(\sigma_{\text{emb}}))$

  \State \quad $\texttt{params} \gets \texttt{params.reshape}(\texttt{params.shape[:1] + (1,1) + params.shape[1:]})$

  \State \quad $\texttt{scale, bias} \gets \texttt{Split}(\texttt{params, 2, axis=-1})$

  \State \Return $\bm{x} \cdot (\texttt{scale }+\texttt{ }1)\texttt{ }+\texttt{ bias}$
    \EndProcedure
\end{algorithmic}
\end{algorithm}

\begin{algorithm}[h!]
\caption{SqueezeBlock}\label{alg:SqueezeBlock}
\begin{algorithmic}[1]
\Procedure{\texttt{SqueezeBlock}(\texttt{out\_channels,squeeze\_channels})}{$\bm{x}\texttt{,}\sigma_{\text{emb}}$} 
  \State $\bm{h} \gets \bm{x}$

  \State $\bm{h} \gets \texttt{GroupNorm}(\bm{h}\texttt{.shape[-1] // 4})(\bm{h})$
  \State $\bm{h} \gets \texttt{Swish}(\bm{h})$
  \State $\bm{h} \gets \texttt{Conv2D}(\texttt{features=squeeze\_channels, kernel\_size=(3,3)})(\bm{h})$
  \State $\bm{h} \gets \texttt{GroupNorm}(\bm{h}\texttt{.shape[-1] // 4})(\bm{h})$
  \State $\bm{h} \gets \texttt{FiLM}(\bm{h}\texttt{,}\sigma_{\text{emb}})$
  \State $\bm{h} \gets \texttt{Swish}(\bm{h})$
  \State $\bm{h} \gets \texttt{Conv2D}(\texttt{features=out\_channels, kernel\_size=(3,3)})(\bm{h})$
  \State $\bm{x} \gets \texttt{Dense}(\texttt{features=}\bm{h}\texttt{.shape[-1]})(\bm{x})$
  \State \Return $(\bm{x} + \bm{h}) / \sqrt{2}$
\EndProcedure
\end{algorithmic}
\end{algorithm}

\begin{algorithm}[h!]
\caption{CNN-based representation}\label{alg:CNN}
\begin{algorithmic}[1]
\Procedure{\texttt{CNN}(\texttt{num\_embed, num\_feature, num\_conv, squeeze\_ratio})}{$\bm{\eta}\texttt{,}\bm{\alpha}\texttt{,}\sigma$}
    \State $\bm{\alpha} \gets \texttt{LayerNorm}(\bm{\alpha})$

  \For{$i \gets 1$ \textbf{to} $\texttt{num\_embed}$}
    \State $\texttt{tmp} \gets \texttt{Conv2D}(\texttt{features=6, kernel\_size=(3,3)})(\bm{\alpha})$
    \State $\texttt{tmp} \gets \texttt{Swish}(\texttt{tmp})$
    \State $\texttt{tmp} \gets \texttt{Conv2D}(\texttt{features=6, kernel\_size=(3,3)})(\texttt{tmp})$
    \State $\bm{\alpha} \gets \texttt{Concatenate}(\texttt{[}\bm{\alpha}\texttt{,tmp], axis=-1})$
  \EndFor

  \State $\bm{y} \gets \texttt{Concatenate}(\texttt{[}\bm{\eta}\texttt{,}\bm{\alpha}\texttt{], axis=-1})$

  \State $\sigma_{\text{emb}} \gets \texttt{FourierEmbedding}(\sigma)$

  \State $\bm{y} \gets \texttt{Conv2D}(\texttt{features=num\_feature, kernel\_size=(3,3)})(\bm{y})$

  \For{$n \gets 1$ \textbf{to} $\texttt{num\_conv}$}
    \State $\bm{y} \gets \texttt{SqueezeBlock}(\texttt{out\_channels=num\_feature,}$
    \State \quad\quad  $\texttt{squeeze\_channels=num\_feature // squeeze\_ratio})(\bm{y},\sigma_{\text{emb}})$
  \EndFor

  \State $\bm{y} \gets \texttt{Conv2D}(\texttt{features=1, kernel\_size=(3,3)})(\bm{y})$

  \State \Return $\bm{y}$
\EndProcedure
\end{algorithmic}
\end{algorithm}

\begin{algorithm}[h!]
\caption{U-ViT}\label{alg:UViT}
\begin{algorithmic}[1]
\Procedure{\texttt{U-ViT}}{$\bm{\eta}\texttt{,} \bm{\alpha}\texttt{,} \sigma$}
  \State $\sigma_{\text{emb}} \gets \texttt{FourierEmbedding}(\sigma)$
  \State $\bm{\alpha}_{\text{emb}} \gets \texttt{Conv2D}(\texttt{features=emb\_dim})(\bm{\alpha})$
  \State $\bm{x} \gets \texttt{Concatenate}(\texttt{[}\bm{\eta}\texttt{,}\bm{\alpha}_{\text{emb}}\texttt{], axis=-1})$
  \State $\texttt{skips} \gets [\bm{x}]$
  \For{$\texttt{level}$ \textbf{in} $\texttt{levels}$}
    \State $\bm{x} \gets \texttt{Downsample}(\bm{x})$
    \State $\bm{x} \gets \texttt{ConvBlock}(\bm{x}\texttt{,} \sigma_{\text{emb}})$
    \If{$\texttt{level==levels[-1]}$}
         \State $\bm{x} \gets \texttt{PositionEmbedding}(\bm{x})$
        \State $\bm{x} \gets \texttt{AttentionBlock}(\bm{x})$
    \EndIf
    \State \text{Append} $\bm{x}$ \text{to }\texttt{skips}
  \EndFor
  \For{$\texttt{level}$ \textbf{in} $\texttt{levels}$}
    \State $\bm{x} \gets \texttt{Concatenate}(\texttt{[}\bm{x}\texttt{,skips.pop()], axis=-1})$
    \State $\bm{x} \gets \texttt{ConvBlock}(\bm{x}\texttt{,} \sigma_{\text{emb}})$
    \If{$\texttt{level==levels[0]}$}
        \State $\bm{x} \gets \texttt{AttentionBlock}(\bm{x})$
    \EndIf
    \State $\bm{x} \gets \texttt{Upsample}(\bm{x})$
  \EndFor
  \State $\bm{x} \gets \texttt{Concatenate}(\texttt{[}\bm{x}\texttt{,skips.pop()], axis=-1})$
  \State \Return $\texttt{Conv2D}(\texttt{features=1})(\bm{x})$
\EndProcedure
\end{algorithmic}
\end{algorithm}

\section{Metrics}\label{appendix:metrics}

\paragraph{Relative Root Mean Square Error (RRMSE)} This is a very well-known metric to primarily quantify the quality of our samples. We define it as follows,

\begin{equation}\label{eq:rrmse_eq}
    \text{RRMSE} = \frac{1}{N_{t}}\sum_{i=1}^{N_{t}}\frac{\|\bm{\eta}_{i} - \bm{\eta}_{0}\|_2}{\|\bm{\eta}_{0}\|_2}
\end{equation}
where $N_{t}$ is the size of the testing set, $\bm{\eta_{i}}$ is the sample generated for test $i$, $\bm{\eta}_{0}$ is the ground truth, and the norm used is the Frobenius norm. In the case of the probabilistic norms, we also take the average across all samples.

\paragraph{Continuous Ranked Probability Score (CRPS)} A good way to measure the efficacy of a probabilistic model is to compare the estimated probability distribution of $p$ to the ground truth value $\bm{\eta}_{0}$. For this task, we use the CRPS~\cite{crps_matheson} which is defined as follows,

\begin{equation}
    \text{CRPS}(P,\bm{\eta}_{0}) = \int (P(\bm{\eta}) - \bm{1}_{\{\bm{\eta}\geq \bm{\eta}_{0}\}})^{2}d\bm{\eta}
\end{equation}

where $P(\bm{\eta})$ is the CDF of $p$. However, in practice, we will use the following equivalent formulation of the CRPS~\cite{crps_swirl},
\begin{equation}
     \text{CRPS}(P,\bm{\eta}_{0}) = \mathbb{E}[\|\bm{\eta} - \bm{\eta}_{0}\|_2] - \frac{1}{2}\mathbb{E}[\|\bm{\eta} - \bm{\eta}'\|_2]
\end{equation}
where $\bm{\eta}$ and $\bm{\eta}'$ are i.i.d. samples of $p$.

\paragraph{Sinkhorn Divergence (SD)} One of the most popular metrics used to measure distance between distributions {is} the Optimal Transport (OT) based metrics, such as the Sinkhorn divergence, which we describe below. The field of OT is concerned with transforming (or transporting) one distribution into another, i.e., finding a map between them, in an optimal manner with respect to a pre-defined cost. 
The cost of the minimal (or optimal) transformation, often called the cost of the OT map, can then be used to define distances between distributions that `lifts' the underlying metric $\mathrm{d}$ defined on $\mathcal{U}$ to one over the space of probability measures defined on $\mathcal{P}(\mathcal{U})$~\citep{santambrogio2015optimal}.

In this context, we define the Kantorovich formulation of the OT cost~\citep{kantorovich1942translocation} as
\begin{align*}
    \mathcal{W}(\mu, \nu) = \min_{\gamma \in \Gamma(\mu, \nu)}\int_{\mathcal{U} \times \mathcal{U}} c(\bm{u}, \bm{v})d\gamma(\bm{u}, \bm{v}),
\end{align*}
where $c: \mathcal{U} \times \mathcal{U} \rightarrow \mathbb{R}^+$ is an arbitrary cost function for transporting a unit of mass from $\bm{u}$ to $\bm{v},$ and $\Gamma$ is the set of joint distributions defined on $\mathcal{U} \times \mathcal{U}$ with correct marginals, i.e.,
\begin{align*}
    \Gamma(\mu, \nu) = \{\gamma \in \mathcal{P}(\mathcal{U} \times \mathcal{U}) \mid P_{1\#}\gamma = \mu, P_{2\#}\gamma = \nu\},
\end{align*} 
with $P_1(\bm{u}, \bm{v}) = \bm{u}$ and $P_2(\bm{u}, \bm{v}) = \bm{v}$ being simple projection operators.
When $c(\bm{u}, \bm{v}) = \,d(\bm{u}, \bm{v})^p$ with $p \geq 1$, then $\mathcal{W}^{1/p}$ is known as a Wasserstein-$p$ distance.

Practically, finding OT maps is a computationally expensive procedure. We therefore use entropic regularized versions of OT costs, which are amenable to efficient implementation on computational accelerators, by means of the Sinkhorn algorithm~\citep{cuturi2013sinkhorn, peyre2019computational}:
\begin{align}\label{eq:ot_reg}
    \mathcal{W}_\varepsilon(\mu, \nu) = \min_{\gamma \in \Gamma(\mu, \nu)} \mathcal{W} + \mathrm{KL(\gamma || \mu \otimes \nu)},
\end{align}
where $\mathrm{KL}$ is the Kullback-Leibler divergence, and $\mu \otimes \nu$ is the product of the marginal distributions.
This gives rise to the Sinkhorn Divergence (SD):
\begin{align*}
    \text{SD}(\mu, \nu) = 2\mathcal{W}_\varepsilon(\mu, \nu) - \mathcal{W}_\varepsilon(\mu, \mu) - \mathcal{W}_\varepsilon(\nu, \nu),
\end{align*}
which alleviates the entropic bias present in \eqref{eq:ot_reg}, i.e. $\mathcal{W}_\varepsilon(\mu, \mu) \neq 0.$
Of note, the SD can be shown to interpolate between a pure OT cost $\mathcal{W}$ (as $\varepsilon \rightarrow 0$) and a MMD (as $\varepsilon \rightarrow 
\infty$)~\citep{ramachandran18, genevay2018learning, feydy2019interpolating}.

We use the Optimal Transport Tools library~\citep{cuturi2022optimal} with its default hyperparameters to perform this computation.


\paragraph{Mean Energy Log Ratio (MELR)} The energy spectrum is one of the main metrics to quantitatively assess the quality of the resulting snapshots~\citep{wan2023debias}. In a nutshell, the energy spectrum measures the energy in each Fourier mode and thereby provides insights into the similarity between the generated and reference samples.  

The energy spectrum is defined\footnote{This definition is applied to each sample and averaged to obtain the metric (same for MELR below).} as 
\begin{equation}
\label{eq:energy_spectrum}
    E(k) = \sum_{|\underline{k}| =  k} | \hat{\eta}(\underline{k}) | ^2 = \sum_{|\underline{k}| = k} \left | \sum_{i,j} \eta(x_{i,j}) \exp(-j 2\pi \underline{k} \cdot x_{i,j}/L) \right|^2
\end{equation}
where $k$ is the magnitude of the wave-number (wave-vector in 2D) $\underline{k}$, and $x_{i,j}$ is the underlying (possibly 2D) spatial grid. 
To assess the overall consistency of the spectrum between the generated and reference samples using a single scalar measure, we consider the mean energy log ratio (MELR):
\begin{equation}
\label{eq:MELR}
    \text{MELR} = \sum_k w_k\left |\log \left (E_{\text{pred}}(k) /E_{\text{ref}}(k) \right )\right |,
\end{equation}
where $w_k$ represents the weight assigned to each $k$. We further define $w_{k}^{\text{unweighted}} = 1/\text{card}(k)$ and $w_{k}^{\text{weighted}} = E_{\text{ref}}(k)/\sum_k E_\text{ref}(k)$. The latter skews more towards high-energy/low-frequency modes.

\section{{Software and Hardware Stack}} \label{sec:soft_hardware}

The wideband scattering data were generated using Matlab. Specifically, it was generated at frequencies of 2.5, 5, and 10 with a dimension of $n_\sca = 80$. It took approximately 8 hours to generate the data on a server equipped with two Xeon E5-2698 v3 processors (totaling 32 cores and 64 threads) and 256 GB of RAM.

The models presented in this paper were implemented using JAX~\cite{jax2018github} and Flax~\cite{flax2020github}, as well as the swirl-dynamics library\footnote{\url{https://github.com/google-research/swirl-dynamics/tree/main/swirl_dynamics/projects/probabilistic_diffusion}} for the ML-pipeline~\cite{wan2023debias}. The experiments were performed on two PNY NVIDIA Quadro RTX 6000 graphics cards.

\section{{Problem Formulation and Optimization}}\label{sec:formulation_optimization}
Following the notation introduced in Section~\ref{sec:architecture}, we denote the set of wideband scattering data by $\{{\sf\Lambda}^\omega\}_{\omega\in\bar{\Omega}}$ by ${\sf\Lambda}$, and the discrete inverse map by $\mathcal{F}_d^{-1}$.

Back-{P}rojection {D}iffusion models generate samples from $p(\bm{\eta} | {\sf\Lambda})$ by using the reverse-time SDE in~\eqref{eqn:alpha_condition_reverse_SDE}, whereas deterministic models approximate the discrete inverse map $\mathcal{F}_d^{-1}$ by a neural network $\Phi_{\bm{\Theta}}$ where $\bm{\Theta}$ denotes the trainable parameters of the deterministic network; namely, they reconstruct perturbations by $\bm{\eta} \approx \Phi_{\bm{\Theta}} ({\sf\Lambda})$. 

The training dataset is identical for both deterministic and diffusion models and it consists of 21,000 data pairs of perturbation and scattering data $(\bm{\eta}^{[s]}, {\sf\Lambda}^{[s]})$ following different distributions, where $[s]$ is the sample index. The evaluation is performed using testing datasets with 500 data points each, which have not been seen by the models during the training stage.  

The following paragraphs cover {the} training and sampling specifics of the denoiser $D_{\bm{\Theta}}$ introduced in Section~\ref{sec:architecture}.

\paragraph{Preconditioning}
We train a conditional denoiser following the form:
\begin{equation}D_{\bm{\Theta}}(\bm{\eta}, {\sf\Lambda}, \sigma) = c_{\text{skip}}(\sigma)\bm{\eta} + c_{\text{out}}(\sigma) S_{\bm{\Theta}_2}\left(c_{\inc}(\sigma)\bm{\eta},F_{\bm{\Theta}_1}(\sf\Lambda),c_{\text{noise}}(\sigma)\sigma\right) \,.
\end{equation}
For the choices of preconditioning, we employ the formulas used in~\cite{karras2022elucidating}.  Specifically, they are
\begin{equation}
\left\{
\begin{aligned}
\text{Skip scaling} &\quad c_{\text{skip}}(\sigma) = \frac{\sigma^2_{\text{data}}}{\left(\sigma^2 + \sigma^2_{\text{data}}\right)}\,, \\
\text{Output scaling} &\quad c_{\text{out}}(\sigma) = \frac{\sigma \cdot \sigma_{\text{data}}}{\sqrt{\sigma^2_{\text{data}} + \sigma^2}}\,, \\
\text{Input scaling} &\quad c_{\inc}(\sigma) = \frac{1}{\sqrt{\sigma^2 + \sigma^2_{\text{data}}}}\,, \\
\text{Noise cond.} &\quad c_{\text{noise}}(\sigma) = \frac{1}{4} \ln(\sigma)\,,
\end{aligned}
\right.
\end{equation}
where $\sigma_{\text{data}}$ is the standard deviation of the perturbations in the training dataset.

\paragraph{Training}
The denoiser is trained to minimize the expected $L_2$ denoising error at samples drawn from $p_{\text{data}}$ for each noise level $\sigma\sim p_{\text{train}}$
\begin{equation}\mathbb{E}_{\sigma \sim p_{\text{train}}}\mathbb{E}_{(\bm{\eta},{\sf\Lambda}) \sim p_{\text{data}}} \mathbb{E}_{\bm{n} \sim \mathcal{N}(\,\cdot\,;0, \sigma^2 I)} \left[\lambda(\sigma)\left\| D_{\bm{\Theta}}(\bm{\eta}+\bm{n} , {\sf\Lambda}, \sigma) - \bm{\eta}\right\|^2_2\right]\,,
\end{equation}
where noise levels have distribution $\sigma\sim p_{\text{train}}$ and {are} weighted by $\lambda(\sigma)$.

In our setting, we employ the loss weighting introduced in~\cite{karras2022elucidating}
\begin{equation}
    \lambda(\sigma) = \frac{\sigma^2 + \sigma_{\text{data}}^2}{(\sigma \cdot \sigma_{\text{data}})^2}\,.
\end{equation}

For the training noise sampling from $p_{\text{train}}$, we consider a function $\sigma_{\text{train}}(t)$ that is derived from a section of the tangent function $\tan(t)$. This section is linearly rescaled so that the input domain is $[0, 1]$ and the output range is $[0, \sigma_{\text{max}}]$. We then sample noise from a uniform distribution in $t \in [t_{\sigma_{\text{min}}}, 1]$ such that $\sigma_{\text{train}}(t_{\sigma_{\text{min}}}) = \sigma_{\text{min}}$.

\paragraph{Sampling}
We generate samples using the reverse-time SDE:
\begin{equation}
    d\bm{\eta} = \left[f(t)\bm{\eta} -g^2(t)\nabla_{\bm{\eta}}\log p_t(\bm{\eta}|\bm{\alpha}_{\sf\Lambda})\right] \, dt + g(t)\, dW_t\,.
\end{equation}
However, it is advised in~\cite{karras2022elucidating} that we formulate the SDE based on the scaling factor $s(t)$ and noise schedule $\sigma(t)$ defined in~\eqref{eqn:perturbation_kernel}, which can be rewritten as
\begin{equation}
f(t) = \frac{\dot{s}(t)}{s(t)}\quad \text{and} \quad g(t) = s(t) \sqrt{2 \dot{\sigma}(t) \sigma(t)}\,.
\end{equation}
Substituting the formulas into the reverse-time SDE, we have
\begin{equation}
 d\bm{\eta} = \left[\frac{\dot{s}(t)}{s(t)}\bm{\eta} -2s^2(t)\dot{\sigma}(t) \sigma(t)\nabla_{\bm{\eta}}\log p_t(\bm{\eta}|\bm{\alpha}_{\sf\Lambda})\right] \, dt + s(t) \sqrt{2 \dot{\sigma}(t) \sigma(t)}\, dW_t\,.
\end{equation}
In our experiments, we adopt the variance preserving formulation in~\cite{song2021scorebased}, where 
\begin{equation}
    s^2(t) = \frac{\sigma^2_{\text{data}}}{\sigma^2_{\text{data}}+\sigma^2(t)}\,.
\end{equation}
For solving the SDE, we consider a {discretization} of time $t$ by a total of $N$ steps, i.e. $t_n = \frac{n}{N-1}$ for $n = 0,1,\dots,N-1$, on which we employ an exponential decaying noise schedule:
\begin{equation}
    \sigma(t_n) = \sigma_{\text{max}}\left(\frac{\sigma_{\text{end}}}{\sigma_{\text{max}}}\right)^{t_n}\,.
\end{equation}

The SDE is then solved by the Euler-Maruyama method~\cite{Kloeden1992}.  

\paragraph{Hyperparameters}
In our experiments, we use normalized data, so we set $\sigma_{\text{data}}=1$.  For training noise sampling, we set $\sigma_{\text{min}}= 10^{-4}$ and $\sigma_{\text{max}} = 100$.  For solving the SDE, we use a time step $N=256$ and $\sigma_{\text{end}}=10^{-3}$. We {then} trained {Back-Projection Diffusion} models for 100 epochs using the Adam optimizer with Optax’s \texttt{warmup\_cosine\_decay}~\cite{deepmind2020jax} as our scheduler. The initial learning rate was set to $10^{-5}$, gradually increased to a peak of $10^{-3}$ over the first 5\% of the training steps, and then decayed to $10^{-8}$ by the end of training. We also employed an exponential moving average (EMA)~\cite{karras2018progressivegrowinggansimproved} of the model parameters with \texttt{ema\_decay}$=0.999$ to stabilize the training and improve performance.

The training specifics of the baseline deterministic models are detailed in Section~\ref{sec:baselines}. The performance of each model {is} assessed using different metrics described in Section~\ref{sec:metrics} that measure the error in terms of individual samples, and at the distributional level. 

\section{{Extension of Section~\ref{sec:posterior_distribution}}}\label{sec:extension_posterior_distribution}
\begin{figure}[h!]
    \centering
    \includegraphics[width=0.9\textwidth]{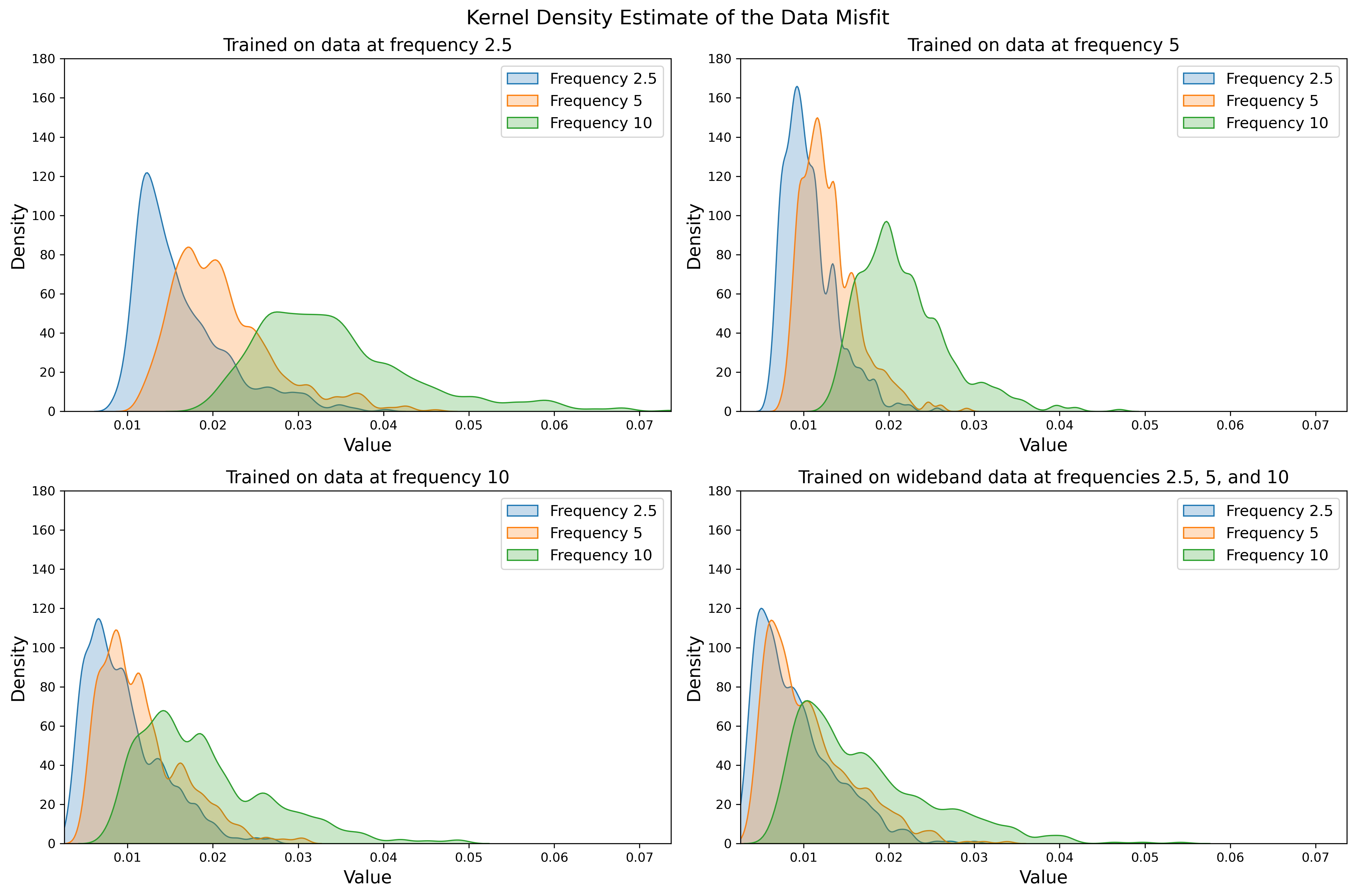}
    \caption{Estimated distributions of the data misfit for EquiNet-CNN for the Shepp-Logan dataset, trained on data at a single frequency of 2.5, 5, and 10, as well as at wideband frequencies including 2.5, 5, and 10.}
    \label{fig:data_misfit_shepp_logan}
\end{figure}

\begin{figure}[h!]
    \centering
    \includegraphics[width=0.9\textwidth]{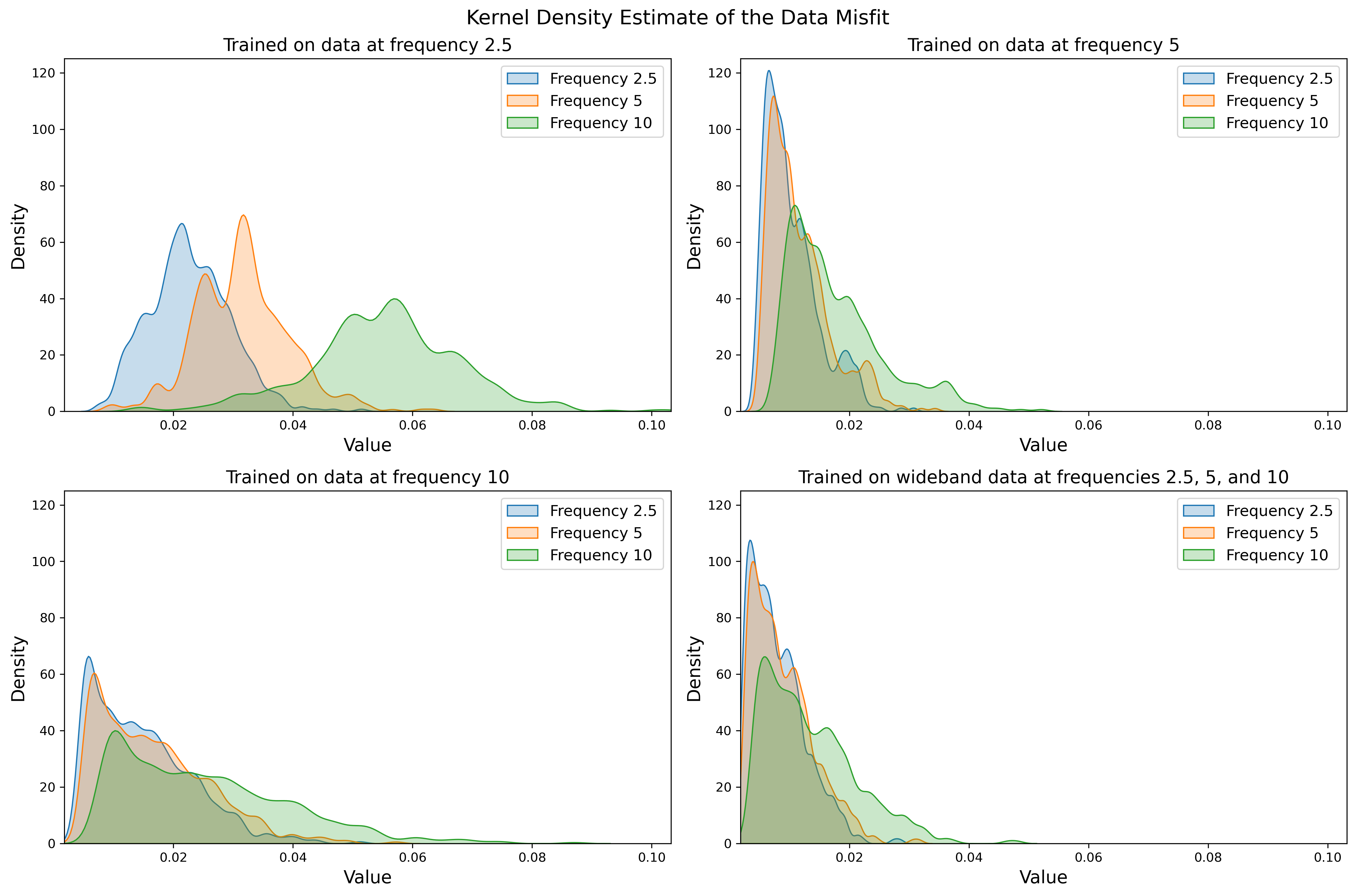}
    \caption{Estimated distributions of the data misfit for EquiNet-CNN for the 10h Overlapping Squares dataset, trained on data at a single frequency of 2.5, 5, and 10, as well as at wideband frequencies including 2.5, 5, and 10.}
    \label{fig:data_misfit_Squares}
\end{figure}

We further extend the experiments from Section~\ref{sec:posterior_distribution} to the Shepp-Logan and 3-5-10h Triangles datasets. In particular, Tables~\ref{tab:data_RRMSE_Shepp_Logan} and \ref{tab:data_RRMSE_Squares} present the statistics of the data misfit at frequencies of 2.5, 5, and 10. Figures~\ref{fig:data_misfit_shepp_logan} and \ref{fig:data_misfit_Squares} illustrate the estimated probability distributions of the data misfit for EquiNet-CNN trained on data at a single frequency of 2.5, 5, or 10, as well as on wideband frequencies including 2.5, 5, and 10.

\begin{table}[h!]
\centering
\begin{tabular}{cccccc}
\toprule
\multicolumn{6}{c}{Trained on Data at Frequency 2.5} \\
\midrule
Frequency & Mean (\%) & Median (\%) & Min (\%) & Max (\%)  & Std  (\%)\\
\midrule
2.5 & 1.625 & 1.453 & 0.844 & 4.022 & 0.546 \\
5  &  2.111 & 1.995 & 1.122 & 4.605 & 0.603 \\
10 &  3.434 & 3.243 & 1.852 & 7.368 & 0.942 \\
\end{tabular}

\begin{tabular}{cccccc}
\toprule
\multicolumn{6}{c}{Trained on Data at Frequency 5} \\
\midrule
Frequency & Mean (\%) & Median (\%) & Min (\%) & Max (\%) & Std (\%) \\
\midrule
2.5& 1.071 & 1.005 & 0.581 & 2.563 & 0.308\\
5  & 1.304 & 1.229 & 0.755 & 2.910 & 0.337\\
10 & 2.167 & 2.061 & 1.262 & 4.698 & 0.530\\
\end{tabular}

\begin{tabular}{cccccc}
\toprule
\multicolumn{6}{c}{Trained on Data at Frequency 10} \\
\midrule
Frequency & Mean (\%) & Median (\%) & Min (\%) & Max (\%) & Std (\%) \\
\midrule
2.5 & 0.959 & 0.860 & 0.291 & 2.692 & 0.451 \\
5   & 1.167 & 1.072 & 0.466 & 3.063 & 0.488 \\
10  & 1.896 & 1.759 & 0.756 & 4.906 & 0.772 \\
\end{tabular}

\begin{tabular}{cccccc}
\toprule
\multicolumn{6}{c}{Trained on Data at Wideband Frequencies including 2.5, 5, and 10} \\
\midrule
Frequency & Mean (\%) & Median (\%) & Min (\%) & Max (\%) & Std (\%) \\
\midrule
2.5 &  0.909 & 0.801 & 0.261 & 3.000 & 0.465 \\
5   &  1.064 & 0.936 & 0.354 & 3.384 & 0.513 \\
10  &  1.724 & 1.514 & 0.653 & 5.415 & 0.809 \\
\bottomrule
\end{tabular}
\caption{Statistics of data misfit at different frequencies for 500 samples generated using EquiNet-CNN on 1 data point from the Shepp-Logan dataset.}
\label{tab:data_RRMSE_Shepp_Logan}
\end{table}

\begin{table}[h!]
\centering
\begin{tabular}{cccccc}
\toprule
\multicolumn{6}{c}{Trained on Data at Frequency 2.5} \\
\midrule
Frequency & Mean (\%) & Median (\%) & Min (\%) & Max (\%)  & Std  (\%)\\
\midrule
2.5 & 2.275 & 2.223 & 0.744 & 5.142 & 0.671\\
5   & 3.159 & 3.134 & 0.874 & 6.364 & 0.779\\
10  & 5.527 & 5.563 & 1.300 & 10.324 & 1.267\\
\end{tabular}

\begin{tabular}{cccccc}
\toprule
\multicolumn{6}{c}{Trained on Data at Frequency 5} \\
\midrule
Frequency & Mean (\%) & Median (\%) & Min (\%) & Max (\%) & Std (\%) \\
\midrule
2.5 & 1.030 & 0.911 & 0.409 & 3.071 & 0.454\\
5   & 1.163 & 1.018 & 0.468 & 3.430 & 0.516\\
10  & 1.775 & 1.543 & 0.707 & 5.208 & 0.805 \\
\end{tabular}

\begin{tabular}{cccccc}
\toprule
\multicolumn{6}{c}{Trained on Data at Frequency 10} \\
\midrule
Frequency & Mean (\%) & Median (\%) & Min (\%) & Max (\%) & Std (\%) \\
\midrule
2.5 & 1.476 & 1.336 & 0.391 & 5.111 & 0.835\\
5   & 1.662 & 1.502 & 0.465 & 5.719 & 0.931\\
10  & 2.532 & 2.282 & 0.707 & 8.698 & 1.417\\
\end{tabular}

\begin{tabular}{cccccc}
\toprule
\multicolumn{6}{c}{Trained on Data at Wideband Frequencies including 2.5, 5, and 10} \\
\midrule
Frequency & Mean (\%) & Median (\%) & Min (\%) & Max (\%) & Std (\%) \\
\midrule
2.5 & 0.794 & 0.696 & 0.177 & 2.845 & 0.455 \\
5   & 0.886 & 0.770 & 0.216 & 3.167 & 0.504 \\
10  & 1.344 & 1.160 & 0.332 & 4.805 & 0.765 \\
\bottomrule
\end{tabular}
\caption{Statistics of data misfit at different frequencies for 500 samples generated using EquiNet-CNN on 1 data point from the 10h Overlapping Squares dataset.}
\label{tab:data_RRMSE_Squares}
\end{table}

\end{document}